\documentclass[letterpaper, 12pt, oneside]{Thesis}  % Use the "Thesis" style, based on the ECS Thesis style by Steve Gunn
\graphicspath{figures/}  % Location of the graphics files (set up for graphics to be in PDF format)

\usepackage[T1]{fontenc}
\usepackage{verbatim}  % Needed for the "comment" environment to make LaTeX comments
\usepackage{adjustbox}
\usepackage{multirow}
\usepackage{hhline}
\usepackage{url}
\usepackage{breakurl}
\usepackage{hyperref}
\usepackage[noadjust]{cite}
\hypersetup{urlcolor=blue, colorlinks=true}  % Colours hyperlinks in blue, but this can be distracting if there are many links.
   
\begin{document}
\frontmatter      % Begin Roman style (i, ii, iii, iv...) page numbering

% Set up the Title Page
\title  {A Genetic Programming System with an Epigenetic Mechanism for Traffic Signal Control}
\authors  {\texorpdfstring
            {\textit{Esteban Ricalde Gonz\'alez}}
            {Author Name}
            }
\addresses  {\groupname\\\deptname\\\univname}  % Do not change this here, instead these must be set in the "Thesis.cls" file, please look through it instead
\date       {\today}
\subject    {}

\maketitle
%% ----------------------------------------------------------------

\setstretch{1.8}  % It is better to have smaller font and larger line spacing than the other way round

% Define the page headers using the FancyHdr package and set up for one-sided printing
\fancyhead{}  % Clears all page headers and footers
\rhead{\thepage}  % Sets the right side header to show the page number
\lhead{}  % Clears the left side page header

\pagestyle{fancy}  % Finally, use the "fancy" page style to implement the FancyHdr headers

\pagestyle{empty}  % No headers or footers for the following pages
\null\vfill

% Now comes the "Funny Quote", written in italics
\textit{``Adding highway lanes to deal with traffic congestion is like loosening your belt to cure obesity.''}

\begin{flushright}
Lewis Mumford
\end{flushright}

\vfill\null

\clearpage  % Funny Quote page ended, start a new page
%% ----------------------------------------------------------------

% The Abstract Page
\addtotoc{Abstract}  % Add the "Abstract" page entry to the Contents
\abstract{
\addtocontents{toc}{\vspace{1em}}  % Add a gap in the Contents, for aesthetics

\setstretch{1.5}  % Reset the line-spacing to 1.8 for body text (if it has changed)

Traffic congestion is an increasing problem in most cities around the world. It impacts businesses as well as commuters, small cities and
large ones in developing as well as developed economies. One approach to decrease urban traffic congestion is to optimize the traffic 
signal behaviour in order to be adaptive to changes in the traffic conditions. From the perspective of intelligent transportation systems, this optimization 
problem is called the traffic signal control problem and is considered a large combinatorial problem with high complexity and uncertainty.  

A novel approach to the traffic signal control problem is proposed in this thesis. The approach includes a new mechanism for Genetic Programming inspired by Epigenetics.
Epigenetic mechanisms play an important role in biological processes such as phenotype differentiation, memory consolidation within generations and environmentally induced 
epigenetic modification of behaviour. These properties lead us to consider the implementation of epigenetic mechanisms as a way to improve the performance of 
Evolutionary Algorithms in solution to real-world problems with dynamic environmental changes, such as the traffic control signal problem.

The epigenetic mechanism proposed was evaluated in four traffic scenarios with different properties and traffic conditions using two microscopic simulators. 
The results of these experiments indicate that Genetic Programming was able to generate competitive actuated traffic signal controllers for all the scenarios tested. 
Furthermore, the use of the epigenetic mechanism improved the performance of Genetic Programming in all the scenarios. The evolved controllers adapt to modifications in 
the traffic density and require less monitoring and less human interaction than other solutions because they dynamically adjust the signal behaviour depending on the local 
traffic conditions at each intersection. 

A microscopic traffic simulator and an open-source modular generic framework to evaluate traffic controllers were developed as part of the research project. 
The proposed framework is the first open-source configurable framework to test machine learning methods on the traffic signal control problem.
}

\vfill\null

\clearpage  % Abstract ended, start a new page
%% ----------------------------------------------------------------

% The Acknowledgements page, for thanking everyone
\acknowledgements{
\addtocontents{toc}{\vspace{1em}}  % Add a gap in the Contents, for aesthetics

This thesis could not be possible without help of a number of people. First and foremost, I
would like to thank my supervisor, Prof. Wolfgang Banzhaf for his continuous
academic support and advice during my studies and research. I really appreciate his insights,
comments, motivation, immense knowledge, and expertise that considerably contributed to this
thesis. Prof. Banzhaf has been a caring and supportive advisor who has
always believed in me, allowing me to pursue independent work. I also thank his
financial support which helped me pursue my graduate studies more easily.

I would like to thank my supervisory committee, Prof. Minglun Gong, and
Prof. Oscar Meruvia-Pastor, for their meticulous review of my thesis and all the comments and suggestions during formal and informal meetings.
Their valuable contributions enriched different aspects of this work.
 
Appreciation goes out to the staff and faculty of the Computer Science Department who
offered me assistance in many levels of my graduate studies. Special thanks go out to Prof.
Adrian Fiech and Prof. Antonina Kolokolova. Adrian's enthusiasm in holding social gatherings and activities helped
boost the sense of community among students and faculty members. Antonina's advice and comments helped me to keep going
when I felt I could not do it anymore.
 
I would like to thank Darlene Oliver, Sharon Deir, Brenda Hillier and Erin Manning for their help and support
during the completion of my program. I also acknowledge the financial contributions of the
School of Graduate Studies of Memorial University which supported my graduate studies. I want to express my eternal gratitude 
to Prof. Ivan Booth and Prof. Jolanta Lagowski for reading and correcting the last version of this thesis.

I want to thank Memorial University to give me the opportunity to study in Canada, to interact with bright people from all the world, 
and learn about the similarities that connect us all. My gratitude goes to friends I made in the department, Xu, Chengling, Anastasia, Wasiq, Bahar, Alejandro, Sahand,
Ali and Javad, for their support, comments, time and friendship. I also want to thank my extended group of friends, Jairo, Jessica, Nhu, Daigo, Bony, Natalia,
Josu\'e, Andrea, Oihane, Ivan, John, Jolanta, Matt and Ted, for helping to make Newfoundland feel like home after all these years.

I express my greatest gratitude to my parents, Luisa and Mario, and my
brothers, Pablo and Sebastian. They sent their unconditional support and love even though I have been away from home for so many years.

Finally, special recognition goes to my girlfriend, Paula, who inspired me and
provided constant encouragement during the final stage. There are no words to describe
how much her love and support have helped me to take the final steps to finalize this project.

}

\vfill\null

\clearpage  % End of the Acknowledgements
%% ----------------------------------------------------------------

\setstretch{1.7}  % Set the line spacing to 1.7, this makes the following tables easier to read

\pagestyle{fancy}  %The page style headers have been "empty" all this time, now use the "fancy" headers as defined before to bring them back

%% ----------------------------------------------------------------
\lhead{\emph{Contents}}  % Set the left side page header to "Contents"
\tableofcontents  % Write out the Table of Contents

%% ----------------------------------------------------------------
\lhead{\emph{List of Figures}}  % Set the left side page header to "List if Figures"
\listoffigures  % Write out the List of Figures

%% ----------------------------------------------------------------
\lhead{\emph{List of Tables}}  % Set the left side page header to "List of Tables"
\listoftables  % Write out the List of Tables

%% ----------------------------------------------------------------

\newpage  % Start a new page
\lhead{\emph{Abbreviations}}  % Set the left side page header to "Abbreviations"
\listofsymbols{ll}  % Include a list of Abbreviations (a table of two columns)
{
% \textbf{Acronym} & \textbf{W}hat (it) \textbf{S}tands \textbf{F}or \\
\textbf{AIMSUN} & \textbf{A}dvanced \textbf{I}nteractive \textbf{M}icro-\textbf{S}imulation for \textbf{U}rban \textbf{N}etworks \\
\textbf{BSE} & \textbf{B}ovine \textbf{S}pongiform \textbf{E}ncephalopathy \\
\textbf{CA} & \textbf{C}ellular \textbf{A}utomata \\
\textbf{CJD} & \textbf{C}reutzfeldt-\textbf{J}akob \textbf{D}isease \\
\textbf{CORSIM} & \textbf{COR}ridor \textbf{SIM}ulation\\
\textbf{DNA} & \textbf{D}eoxyribo\textbf{N}ucleic \textbf{A}cid \\
\textbf{EA} & \textbf{E}volutionary \textbf{A}lgorithm \\
\textbf{EC} & \textbf{E}volutionary \textbf{C}omputation \\
\textbf{EL} & \textbf{E}pigenetic \textbf{L}earning \\
\textbf{ELGP} & \textbf{E}pigenetic \textbf{L}inear \textbf{G}enetic \textbf{P}rogramming \\
\textbf{EpiGP} & \textbf{EpiG}enetic \textbf{P}rogramming \\
\textbf{FHA} & \textbf{F}ederal \textbf{H}ighway \textbf{A}dministration \\
\textbf{GA} & \textbf{G}enetic \textbf{A}lgorithm \\
\textbf{GP} & \textbf{G}enetic \textbf{P}rogramming \\
\textbf{LQF-MWM} & \textbf{L}ongest \textbf{Q}ueue \textbf{F}irst with \textbf{M}aximum \textbf{W}eight \textbf{M}atching \\
\textbf{MiniTraSim} & \textbf{Mini}mum \textbf{Tra}ffic \textbf{Sim}ulator \\
\textbf{miRNA} & \textbf{mi}cro \textbf{R}ibo\textbf{N}ucleic \textbf{A}cid \\
\textbf{RNA} & \textbf{R}ibo\textbf{N}ucleic \textbf{A}cid \\
\textbf{SCATS} & \textbf{S}ydney \textbf{C}oordinated \textbf{A}daptive \textbf{T}raffic \textbf{S}ystem \\
\textbf{SUMO} & \textbf{S}imulation of  \textbf{U}rban  \textbf{MO}bility \\
\textbf{TCA} & \textbf{T}raffic \textbf{C}ellular \textbf{A}utomata \\
\textbf{TraCI} & \textbf{T}raffic \textbf{C}ontrol \textbf{I}nterface \\
\textbf{TRANSYT} & \textbf{TRA}ffic \textbf{N}etwork \textbf{S}tud\textbf{Y} \textbf{T}ool \\
\textbf{UTOPIA} & \textbf{U}rban \textbf{T}raffic \textbf{OP}timization by \textbf{I}ntegrated \textbf{A}utomation \\
\textbf{XML} & e\textbf{X}tensible \textbf{M}arkup \textbf{L}anguage \\
}

%% ----------------------------------------------------------------
%\clearpage  % Start a new page
%\lhead{\emph{Physical Constants}}  % Set the left side page header to "Physical Constants"
%\listofconstants{lrcl}  % Include a list of Physical Constants (a four column table)
{
% Constant Name & Symbol & = & Constant Value (with units) \\
%Speed of Light & $c$ & $=$ & $2.997\ 924\ 58\times10^{8}\ \mbox{ms}^{-\mbox{s}}$ (exact)\\

}

%% ----------------------------------------------------------------
%\clearpage  %Start a new page
%\lhead{\emph{Symbols}}  % Set the left side page header to "Symbols"
%\listofnomenclature{lll}  % Include a list of Symbols (a three column table)
{
% symbol & name & unit \\
%$a$ & distance & m \\
%$P$ & power & W (Js$^{-1}$) \\
%& & \\ % Gap to separate the Roman symbols from the Greek
%$\omega$ & angular frequency & rads$^{-1}$ \\
}
%% ----------------------------------------------------------------
% End of the pre-able, contents and lists of things
% Begin the Dedication page

\setstretch{1.5}  % Return the line spacing back to 1.8

%\pagestyle{empty}  % Page style needs to be empty for this page
%\dedicatory{For/Dedicated to/To my\ldots}

\addtocontents{toc}{\vspace{2em}}  % Add a gap in the Contents, for aesthetics

%% ----------------------------------------------------------------
\mainmatter	  % Begin normal, numeric (1,2,3...) page numbering
\pagestyle{fancy}  % Return the page headers back to the "fancy" style

% Include the chapters of the thesis, as separate files
% Just uncomment the lines as you write the chapters

\chapter{Introduction}
\label{chap:intro}

  \graphicspath{{figures/PNG/}{figures/PDF/}{figures/EPS/}{figures/}}
  \lhead{\emph{Chpt 1: Introduction}}  % Set the left side page header to "Introduction"

The United Nations reported that by 2030, urban areas will house 60\% of people globally. One in every three individuals will live in 
cities with at least half million of inhabitants \cite{nationen2016world}. In 2015, there were more than 900 million passenger cars in the
world, number that increases roughly 40 million per year \cite{Oica2018cars}. A study among 1,360 cities identified that, in 2017, 
drivers spent an average of nine percent of their travel time dealing with traffic congestion \cite{cookson2017inrix}. The same study concludes that
traffic congestion is a global phenomenon. It impacts businesses as well as commuters, small cities and
large ones in developing as well as developed economies.  Traffic congestion main impacts are 
transportation delays resulting in late arrival for employment, education and deliveries; 
increase the amount of time on the roads; the costs associated with fuel consumption; and 
the carbon dioxide emissions and air pollution caused by increased idling, acceleration and braking \cite{cookson2017inrix}. 

Any attempt to address the increasing traffic congestion problem should consider multiple components: improving public transportation systems, promoting
alternative transportation methods and policies (car sharing, bicycles, etc.), increasing the transportation infrastructure (new roads, roundabouts, more lanes 
for existing roads, etc.), and improving the efficiency of the existing components. One approach to improve the current infrastructure is to optimize the 
traffic signal behaviour in order to be adaptive to changes of the traffic conditions. However, urban traffic signal control is a large combinatorial 
problem with high complexity and uncertainty \cite{yang1996model,zhao2012computational}. 

Because of its complexity, the solution to the traffic signal control problem is not feasible for more than one intersection \cite{papageorgiou2003review}.
Therefore, proposed control strategies for traffic signal control introduce several simplifications to partially address 
the optimization of traffic signals of simple networks only composed of a few intersections.
Because of this limitation, solutions to more complex real scenarios require the use of microscopic simulators and heuristic algorithms.

Microscopic traffic simulators model individual elements of transportation systems, such as the dynamics of individual vehicles and the behaviour 
of drivers. These simulators depend on random numbers to generate vehicles, select routes and determine the behaviour of each component of traffic systems. 
In a microscopic traffic simulator, the dynamic variables of the model represent microscopic properties such as the position and the velocity of 
individual vehicles.

Different machine learning algorithms have been used in combination with microscopic traffic simulators to improve the signal behaviour of 
specific traffic scenarios. Some examples of machine learning algorithms used in the solution to the traffic signal problem are 
self-organization \cite{zubillaga2014measuring}, neural networks \cite{li2016traffic}, genetic algorithms \cite{armas2016traffic}, 
fuzzy logic \cite{padmasiri2014genetic} and reinforcement learning \cite{balaji2010urban}. However, traffic signal control is an 
atypical problem from the computer science perspective. There are no standard benchmarks, and test suites did not exist before 2016. 
Furthermore, a comparative study of the state of the art methods has not been carried out yet. 

Several factors complicate the objective comparison of different solutions to the traffic signal control problem.
Different studies consider diverse variables and parameters along various
simulators through different representations of the problem. Some scenarios consider external components such as accidents, road-blocks, public transportation,
bicycles, pedestrians or weather conditions as an approach to represent the uncertainty of the real phenomenon \cite{lu2016modelling}. Sometimes vehicle-to-vehicle,
vehicle-to-signal or signal-to-signal communications are considered.
Multiple commercial solutions; such as TRANSYT \cite{robertson1969transyt}, SCATS \cite{sims1980sydney}, UTOPIA \cite{mauro1990utopia} and BALANCE \cite{friedrich1995balance}; 
are available with drastic changes in performance from one version to other. The commercial solutions require expensive
licenses in order to be used or tested, and their technical reports include minimal information about the algorithms used.
 
Furthermore, traffic simulation is a computationally expensive process that is difficult to parallelize. To achieve parallelization, the road 
network should be partitioned into sub-networks and the simulation of these sub-networks should be executed on different computers \cite{igbe2003open, dai2010parallel}.
Most of the modern traffic simulators do not support parallelization by default and custom modifications are requiered to allow it \cite{ahmed2016partitioning}. 
Because of this restriction, an exhaustive amount of computational time is required to improve the signals of a specific traffic scenario regardless of the method used.
Finally, some implementations have strong dependencies with the simulator and/or the topology of the related traffic scenario and cannot be tested under 
different circumstances. For example, Li et al. \cite{li2016traffic} used Reinforcement Learning to train a Deep Neural Network to optimize time schedules 
of a basic scenario with a single intersection of two-directional roads with two lanes. No turns (left-turns, right-turns or U-turns) are allowed. 
Each traffic signal has only two possible phases: red and green. The algorithm equation is defined used these fixed parameters. Therefore, a redefinition
of this equation is required in order to apply this approach to more realistic scenarios (including turns, yellow phases or multiple intersections).

Despite the limitations here described, research groups in transportation, engineering and computer science around the world have tried to 
propose novel solutions for traffic signal control because of the cost and environmental impact traffic congestion has on today's society.

This thesis focuses on solutions to the traffic signal control problem based on Evolutionary Algorithms. \textit{Evolutionary Algorithms} (EAs) are methods used to 
approximate solutions for complex problems inspired by biological evolution.
In EAs, candidate solutions to optimization or learning problems are encoded within data structures called chromosomes. The phenotypic expression of these chromosomes 
plays the role of individuals in a population. Selection and combination operators, such as mutation and crossover, are applied to a population to produce new 
generations of solutions. The performance of a population gradually improves by repeating a cycle of genetic operations through multiple generations. 
EAs are parallel, context-free problem solvers that often produce acceptable solutions to complex problems of different disciplines. 

Evolutionary Algorithms can be used to solve real-world problems where the goal can change during the evolutionary process.
In this context, an EA must adapt to continuous changes of the goal and fitness evaluation parameters for the exploration of such dynamic environments. 
However, traditional implementations of Evolutionary Algorithms require modifications in order to adapt to continuous changes in the fitness landscape \cite{yaochu2005evolutionry}.
One approach to face these changes is to generate variable locally adaptable solutions. 

Evolutionary Algorithms are inspired by Darwin's theory of evolution by natural selection \cite{darwin2004origin}. Most implementations use a traditional approach 
and are based on the widely accepted gene-centred modern synthetic theory of evolution. However, previous work indicates that
EAs can benefit from additional mechanisms inspired by new developments in different areas of biology. 
Some examples of EA modifications inspired by biological mechanisms are migration between sub-populations \cite{kommenda2014genetic}, 
neutrality \cite{keller1996genetic}, role specialization \cite{ferrauto2013different} and 
regulatory networks \cite{banzhaf2003artificial}.

Epigenetics is defined as the study of stably heritable phenotype resulting from changes in gene expression without alterations in the DNA sequence 
\cite{berger2009operational}.
Recent discoveries indicate epigenetic mechanisms have a more active role in adaptation and development as part of biological 
evolution compared to what it was previously believed.
These mechanisms are related to properties such as phenotype differentiation, memory consolidation within generations and 
environmentally induced epigenetic modification of behaviour. These properties lead us to consider whether the implementation of epigenetic mechanisms 
may be beneficial to improve the performance of EAs in the solution to real-world problems with dynamic environmental changes, such as the traffic control
signal problem.

\textit{Genetic Algorithms} (GAs) are EAs where chromosomes are represented as an array of numeric values. GAs have been used before in
solution to the traffic signal control problem \cite{sanchez2010traffic,braun2011evolutionary,zhao2012computational,li2017optimizing}. The typical approach
assumes the traffic has a periodic behaviour and optimizes hourly time schedules of traffic signals based on historical data. However, given that 
traffic congestion is a dynamic phenomenon, this approach has the downside of requiring constant updates of the schedules. 
Furthermore, it requires constant monitoring and the design of contingency plans to react to atypical congestion events.

\textit{Genetic Programming} (GP) is an EA where chromosomes represent programs or algorithms to be evolved for the purpose of
inductive learning. Distinct representations have been used in different GP implementations. Examples include, dynamic tree structures 
(Tree-based Genetic Programming) \cite{koza1990genetic}, sequence of imperative programming instructions (Linear Genetic Programming) \cite{banzhaf1998genetic}, 
computer program graphs (Cartesian Genetic Programming) \cite{miller2000cartesian} and grammatical maps of integer arrays (Grammatical Evolution) 
\cite{o2001grammatical}.

GP can be used in solution to the traffic signal control problem by the evolution of traffic signal controllers. These controllers are not 
pre-calculated signal schedules but adaptive rules evolved for online modification of the duration of signal phases. For the research presented in this thesis, 
an epigenetic mechanism was incorporated in controllers evolved through GP to facilitate the adaptation to constant changes on the traffic conditions. 
This approach adapts to modifications in the traffic density and may require less monitoring and less human interaction than other solutions.

\section{Contributions of the Thesis}

The primary goal of this thesis is to propose a novel approach for the traffic signal control problem. The approach is
based on the evolution of traffic signal controllers through GP with the incorporation of a novel mechanism inspired by Epigenetics.
This thesis presents contributions to two different fields: Evolutionary Algorithms and traffic signal control.

From the EA perspective, the thesis presents recent developments in biology around Epigenetics and reviews previously published work related
to epigenetic mechanisms in Evolutionary Algorithms. A novel epigenetic mechanism that considers environmental changes to trigger epigenetic mutations
is presented and discussed. The proposed mechanism improves the performance of Genetic Programming in the solution to problems with dynamic environments 
because it facilitates the adaptation of individuals to environmental changes. 

From the traffic signal control perspective, the thesis presents a microscopic traffic simulator and an open-source modular framework to evaluate 
traffic controllers. The simulator models dynamic traffic conditions of complex traffic networks with several roads of multiple lanes using an approach 
with low computational cost. The modular framework operates with SUMO and can be used to evaluate traffic controllers generated by machine learning 
methods in solution to different traffic scenarios. SUMO \cite{SUMO2012} is an open-source multi-platform microscopic traffic simulation suite widely used for 
traffic simulation, traffic management and traffic signal optimization.  

Furthermore, the thesis presents a novel approach for evolution of actuated traffic controllers using Genetic Programming. The evolved controllers are 
not pre-calculated signal schedules but adaptive rules evolved for online control of the duration of signal phases. Finally, the thesis compares the 
performance of this approach to other methods traditionally used to solve the traffic signal control problem through different traffic scenarios 
ranging from a single traffic intersection to a section of a real city with real traffic data.   

Partial content of this thesis and related research has been published in conference proceedings as follows:

\begin{itemize}
\item \textbf{Esteban Ricalde} and Wolfgang Banzhaf. A Genetic Programming Approach for the Traffic Signal Control Problem with Epigenetic Modifications. 
 \textit{European Conference on Genetic Programming}. Springer International Publishing, 2016, pp. 133-148.
\item \textbf{Esteban Ricalde} and Wolfgang Banzhaf. Evolving Adaptive Traffic Signal Controllers for a Real Scenario Using Genetic Programming with an Epigenetic Mechanism. 
 \textit{16th IEEE International Conference on Machine Learning and Applications (ICMLA)}. IEEE, 2017, pp. 897-902.
\end{itemize}

\section{Overview}

Chapter \ref{chap:epigenetics} reviews the developments around evolution and Epigenetics from the perspectives of biology and computer science. Historical changes to 
the definition of biological evolution and Epigenetics are described. A brief introduction of EAs and GP is presented. Recently published implementations of mechanisms
inspired by Epigenetics for EAs are described and discussed. 

Chapter \ref{chap:approach} introduces the proposed epigenetic mechanism. The mechanism is based on activation and deactivation of chromosomal sections 
controlled by environmental modifications. A heuristic, called the adaptive factor, is introduced to measure the local variability of the environment.  

Chapter \ref{chap:application} introduces a general description of the traffic signal control problem and 
explores different solutions based on adaptive controllers, with emphasis on Evolutionary Algorithms. The chapter also presents a brief introduction to 
traffic simulation.

Chapter \ref{chap:synthScenarios} describes the representation used for the evolution of actuated traffic controllers using Genetic Programming. 
The adaptive factor is redefined in the context of traffic signal optimization. Experiments with two synthetic traffic scenarios are presented and
the results obtained are discussed.

Chapter \ref{chap:sumo} extends the work presented in Chapter \ref{chap:synthScenarios} operating with a simulator widely used by the community.
Two additional traffic scenarios are evaluated. One of the scenarios is the Adrea Costa scenario of Bologna city in Italy with real traffic 
data \cite{bieker2015traffic}.

Chapter \ref{chap:conclusions} concludes by discussing future research directions of our work.

Appendix \ref{apdx:our_sim} describes the proposed microscopic traffic simulator. The simulator uses multiple layers, multi-agent concepts and object oriented 
programming components to allow the simulation of multiple roads with multiple lanes. A modification to a popular traffic cellular automaton is proposed for
the vehicular model.
  
Appendix \ref{apdx:sumo_diff} describes the open-source modular framework to evaluate traffic controllers. The framework requires multiple 
configuration files and uses SUMO, a simulation suite widely used by the community.  

\section{Disclosure Statement}

In agreement with the third point of the voluntary commitment to research transparency \cite{schonbrodt2015voluntary} and 
the 21-word solution \cite{simmons201221},  we report how we determined our sample size, 
all data exclusions (if any), all manipulations and all measures for all the experiments presented in this thesis.

 % Introduction

\chapter{Evolution and Epigenetics}
\label{chap:epigenetics}

  \graphicspath{{figures/PNG/}{figures/PDF/}{figures/EPS/}{figures/}}
  \lhead{\emph{Chpt 2: Evolution and Epigenetics}}  % Set the left side page header to "Evolution and Epigenetics"

This chapter presents a brief introduction to evolution and Epigenetics from the perspective of biology and computer science.
Evolution is one of the most important and broadly discussed concepts in biology. The goal of this chapter is to explain how 
the concepts of evolution and Epigenetics have changed over time as new evidence has been discovered. 

This chapter also introduces Evolutionary Algorithms as population-based metaheuristic methods inspired by
biological evolution. Furthermore, recently published modifications to Evolutionary Algorithms inspired by epigenetic mechanisms 
are presented and discussed.

\section{Biological Evolution}
\label{sec:bioevol}

Evolution is the change in heritable characteristics of biological populations over successive generations. 
Evolutionary processes give rise to biodiversity at every level of biological organization, 
including species, individuals, organs, cells, and molecules.

The theory of evolution has been developed and tested over nearly 160 years. 
It has been developed gradually as new evidence is discovered and incorporated over previous accomplishments.
Discoveries in evolutionary biology have made a significant impact not only in biology 
but also in other disciplines such as agriculture, medicine, anthropology, sociology, economy and computer sciences. 

Charles Darwin and Alfred Russel Wallace were the first authors to relate the term evolution with the variation of 
the species as the process to achieve adaptation to changes in the environment. 
Darwin's book, \textit{The Origin of Species by Means of Natural Selection} \cite{darwin2004origin}, provided a rational
and convincing explanation of the causes as well as evidence of the fact of evolution in plants and animals \cite{campbell2017human}.
One of the main contributions made by Darwin was the description of natural selection.
The natural selection of the fittest species and individuals is the principal macroscopic mechanism of evolution.

%Alternative paragraph
%Even though the work of Darwin and Wallace largely demonstrated the ``facts'' of evolution, the biological mechanisms through 
%which evolution operated were unknown at the time their discoveries were published. 
Although the evidence presented by Darwin and Wallace was appealing and clearly explained, the biological mechanisms through 
which evolution operates were not known at the time their discoveries were published. It took nearly 80 years to gradually incorporate 
the advances in morphology, paleontology, systematics, cytology, genetics, and molecular research into what is now known as the  
\textit{Modern Synthetic Theory} of evolution. The modern synthesis reconciles Darwinian evolution with Mendelian genetics
and describes in detail the genetic mechanisms through which natural selection operates, such as genetic mutation and recombination, DNA inheritance, 
population genetics, isolation and speciation \cite{winters2018introduction}. The genetic material stored in the DNA has an important role in the modern 
synthetic theory of evolution. In other words, the modern synthesis can be considered a gene-centred view of evolution \cite{noble2011neo}.

More recently, different researches \cite{pigliucci2010evolution,laland2015extended,futuyma2017evolutionary} have discussed the idea of building upon 
the modern synthesis to develop what is called the \textit{Extended Evolutionary Synthesis}. This new conceptual framework of evolution 
is called extended synthesis because it is an extension, rather than a contradiction, to the modern synthesis. 

In addition to the elements proceeding
from the modern synthesis, the extended synthesis incorporates components based on new evidence of the role that
non-genetic mechanisms have in inheritance. Some of the components included by the extended synthesis are gene regulatory networks, 
phenotypic plasticity, niche construction, developmental biology and epigenetic inheritance. Gene regulatory networks are composed by molecular regulators 
that interact between them and with different components of the cell to control the gene expression levels of RNA and proteins. Phenotypic plasticity is the ability 
organisms have to modify their behaviour, morphology and physiology in response to changes in the environment. Niche construction is the process in which
the metabolism, activities and decisions of an organism modify the environment, thus affecting its own development and the development of other species.
The evolutionary-development theory studies the capacity that developmental processes have to evolve over time. Epigenetic inheritance is described in
detail in Section \ref{sec:epigen}.

\section{Evolutionary Algorithms}
\label{sec:evoalg}

In computer science, \textit{Evolutionary Computation} (EC) is a family of population-based trial and error general problem solvers. 
EC includes different metaheuristic techniques that can be used to approximate solutions to a wide range of problems, 
including NP-hard problems. 

\textit{Evolutionary Algorithms} (EAs) are the subset of EC that considers the techniques inspired by biological evolution. 
There are other methods included in Evolutionary Computation which are not strictly based on biological evolution. 
Ant colony optimization \cite{dorigo1992optimization}, particle swarm optimization \cite{kennedy1995particle}, artificial immune systems \cite{kephart1994biologically} 
and estimation of distribution algorithms \cite{larranaga2001estimation} are examples of techniques included in EC but not considered Evolutionary Algorithms.   

In the same way that evolutionary theory has gradually developed as different researchers find new evidence and connect known ideas, 
the study of Evolutionary Algorithms have required the collective work of multiple researchers over 60 years. The
first written description of an evolutionary process for computer problem solving was included in one of the early machine learning papers published in 1958 
\cite{friedberg1958learning}. However, the works by Bremermann et al. \cite{bremermann1965search} and Spendley \cite{spendley1962sequential} are evidence that 
similar ideas were discussed within different disciplines around the same time. 

It was during the decade of 1960s when the bases for three of the main Evolutionary Algorithms were independently proposed.
Holland \cite{holland1962outline} founded the field known as Genetic Algorithms in an attempt to understand the underlying principles of adaptive systems.    
Bienert, Rechenberg and Schwefel \cite{rechenberg1965cybernetic} proposed an Evolutionary Strategy to optimize the shape of a three-dimensional object to 
move in a wind tunnel. Fogel \cite{fogel1966artificial} presented the key concepts for Evolutionary Programming as the evolution of finite-state machines to 
forecast non-stationary time series.
        
Over the years, multiple modifications have been proposed to these original methods and new EAs have been proposed and explored. Most of the Evolutionary Algorithms share the 
following overall steps: (1) some type of representation is used to encode the candidate solutions, often called individuals, into chromosomes;
(2) the chromosomes of the initial population of individuals are generated; (3) one or multiple fitness functions are evaluated to calculate the survival aptitude of the individuals; 
(4) a selection operator is used to define which individuals will remain in the population for each generation; (5) reproduction operators, divided into conservation
operators and innovation operators \cite{banzhaf1998genetic}, are applied to generate new chromosomes every generation; and (6) a termination criterion defined a priori 
must be fulfilled to stop the experiment after multiple generations.
         
Although these basic steps are shared between most of the EAs, there are many different models with subtle differences and additional steps.
One of the main differences between the techniques considered as Evolutionary Algorithms is the representation used for the chromosomes of the individuals 
(arrays of real values, arrays of binary numbers, finite-state machines, syntactic trees, et cetera).

\subsection{Genetic Programming}
\label{subsec:gp}

\textit{Genetic Programming} (GP) is an Evolutionary Algorithm that uses an executable program representation.
As other Evolutionary Algorithms, it is a population-based search method. 
GP starts from a high-level statement of the problem requirements and iteratively generates a computer program to solve the problem.  
Because of the flexibility of its representation and its context independent methodology, 
GP can be used to propose solutions to problems of different scientific and technological disciplines.

As mentioned in Section \ref{sec:evoalg}, the idea of evolving programs has been one of the goals of machine learning ever since Friedberg \cite{friedberg1958learning} presented
a method to solve simple problems through the automatic modification of computer programs. Fogel's Evolutionary Programming \cite{fogel1966artificial}
was an early attempt at exploring the concept of evolving simple programs represented by finite-state machines.

In the 1980s, Genetic Algorithms were widely used to evolve valid solutions to a large number of problems.
Due to this popularity, many researchers tried to modify GA's representation and/or its operators
to approach the goal of program evolution. Two researchers, Cramer \cite{cramer1985representation} and Koza \cite{koza1989hierarchical}, presented 
case-specific examples indicating that tree structures are feasible options to represent programs as part of Genetic Algorithm systems. 
However, Koza \cite{koza1990genetic} was the first author to demonstrate the importance of such representation and that the artificial evolutionary 
process is a viable option for automatic program generation. This variation of the Genetic Algorithm began to be called Genetic Programming around 
the same time.
  
These days, Genetic Programming refers to any Evolutionary Algorithm used to evolve executable programs regardless of the data structure used for its representation.
Binary trees, linear sequence of instructions and graph-based structures are the most commonly used representations in GP \cite{banzhaf1998genetic}. 
Each of these program representations requires modifications in the way chromosomes are encoded and decoded, different genetic operators, and specific initial parameters.  

Maximum depth, function set, terminal set, population size, population initialization method, selection method, crossover method, crossover rate, mutation method,
mutation rate and elitism are the standard initial parameters required by tree-based Genetic Programming. There is a wide range of different methods that can be used 
for selection, crossover and mutation steps in GP, and new methods are presented in each GP conference. The detailed definition of
the standard components of GP is outside the scope of this thesis because it is already covered in numerous reviews of the literature 
\cite{koza1994genetic,banzhaf1998genetic,langdon2002foundations,poli2008field}.

\section{Epigenetics in Biology}
\label{sec:epigen}

In the modern synthesis model of evolution, the heritable information is stored exclusively in the chromosomal \textit{deoxyribonucleic acid} (DNA) 
of each cell. 
At the cellular level, this information is transferred, without external modifications, during cellular division. For the reproduction 
of eukaryotic organisms, the unmodified DNA is processed by genetic recombination and meiosis to give life to a new being. 

However, evidence collected in the past half century has led researchers to recognize the existence of epigenetic inheritance systems 
through which information additional to the DNA sequence can be transmitted in the lineage of cells and organisms. 
This discovery broadens the concept of inheritance and challenges the widely accepted 
gene-centred modern synthesis theory of evolution.

\subsection{Definition of Epigenetics}
\label{subsec:epigenDef}

Epigenetics is defined as the study of stably heritable phenotype resulting from changes in gene expression without alterations in the DNA sequence 
\cite{berger2009operational}.

The term Epigenetics was created in the early 1940s, when Conrad Waddington first defined it as ``the branch of biology which studies the 
causal interactions between genes and their products which bring the phenotype into being'' \cite{waddington2008basic,waddington2012epigenotype}.
At that time, little was known about the role genetic networks have in gene expression and gene regulation. 
Waddington's work pioneered the study of the interaction between genetic networks and the embryonic development process \cite{holliday2006epigenetics}. 
 
Waddington was mainly interested on channelling and plasticity. Despite the genetic variation, development usually leads to the same  
well-defined end results. In Waddington words, ``the development process is canalized by natural selection towards   
specific valleys of an epigenetic landscape'' \cite{waddington1942canalization}. On the other hand, cells 
with the same genetic material can differ markedly in structure and function.
Development has the plasticity to generate skin cells, brain cells and kidney cells from the same genetic material. Similarly, worker bees and
queen bees differ in morphology and behaviour even though they share the same genetic material. 
All these phenotypic heritable differences are not fully genetic. Epigenetic factors have an important role in these processes.

Waddington's work aimed to solve questions that geneticists had not asked themselves in a clear and consistent way. 
However, he did not detail the biological mechanisms by which these processes were carried out.

In the following decades, different studies were performed to connect genes and development. 
In 1958, Nanney summarized several of them as a review of the first 20
years of work about Epigenetics \cite{nanney1958epigenetic}.
However, at that time, different authors had their own definition of Epigenetics. The term was used 
almost as a synonym of developmental biology, and no specific epigenetic mechanisms had been discovered.

It was not until 1969 when the first clear evidence of an epigenetic mechanism was presented.
Griffith and Mahler \cite{griffith1969dna} proposed that DNA methylation (or demethylation) is important for 
long term memory in the brain. 

The work by Riggs \cite{riggs1975x} and the work by Holliday and Pugh \cite{holliday1975dna}, both of them presented
in 1975, outlined a molecular model for the inactivation of genes, inheritance of gene inactivity, 
and the implication that DNA methylation can have strong effects on gene expression.

The change in the meaning of Epigenetics was related to the discovery of the molecular mechanisms
to control gene activity and the inheritance of cell phenotypes.
The work of Holliday in cell memory, mainly around methylation, was one of the most
important contributions to the definition of epigenetic mechanisms. The change in the way Epigenetics is defined can be
read in his writings. 

In 1990, Holliday defined Epigenetics as ``the study of the mechanisms of 
temporal and spatial control of gene activity during the development of complex organisms'' \cite[pp. 432]{holliday1990mechanisms}. 
Although this definition is very close to Waddington's initial
definition, he began to provide more detail about the way the process is performed by adding that ``Mechanisms of epigenetic control
must include the inheritance of a particular spectrum of gene activities in
each specialized cell. In addition to the classical DNA code, it is necessary to
envisage the superimposition of an additional layer of information which
comprises part of the hereditary material, and in many cases this is very stable. The 
term epigenetic inheritance has been introduced to describe this situation.''

A few years later, Holliday defined Epigenetics as ``the study
of the changes in gene expression, which occur in organisms with differentiated cells, and the mitotic inheritance 
of given patterns of gene expression'' \cite[pp. 453]{holliday1994epigenetics}. In this definition, he is not
only considering DNA-protein interactions, but the discussion of the field has broadened to include DNA rearrangements 
in the immune system, and mitochondrial inheritance. 
By that time, it had been already discovered that information additional to the DNA sequence can be transmitted from 
parents to offspring through genomic imprinting.

A more recent change to the Epigenetics definition is the constraint that the beginning of the new epigenetic state 
should involve a transitory mechanism separated from the one required to maintain it \cite{berger2009operational}.

\subsection{Epigenetic mechanisms}
\label{subsec:epigenMecha}

Before going into more detail about recent work in Epigenetics, it is important to describe the
most commonly studied epigenetic mechanisms: the post-translational modification of
histone proteins and the methylation of DNA cytosine residues \cite{breton2017small}.

\subsubsection{Histone modification}
\label{subsubsec:histone}

\begin{figure}
\centering
\includegraphics[scale=0.9]{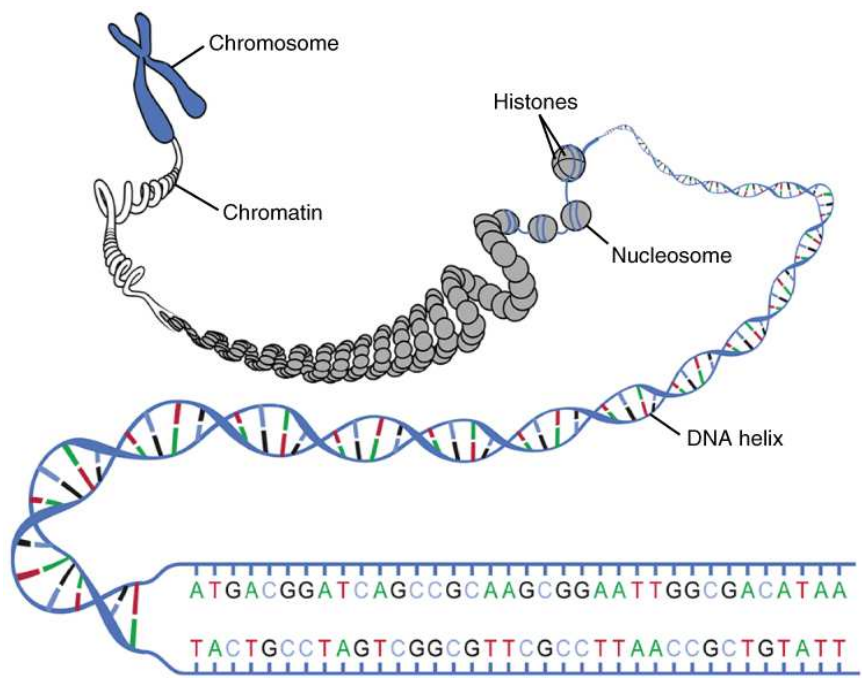}
\caption[Strands of DNA wrapped around histones to form chromatin]{Strands of DNA wrapped around histones to form chromatin \cite{OpenStax2016dnamacrostructure}.}
\label{DNA_macrostructure}
\end{figure}

Histone modification is related to the way DNA is wrapped. 
A human genome has over 3 billion base pairs. 
Each cell would require one meter of linear space to store the DNA in the shape of a double helix string. However, in eukaryotic cells, 
the DNA helix is curled up around nucleosomes of highly alkaline proteins, called histones, 
to form chromatin structures as shown in Figure \ref{DNA_macrostructure}. 

Two different types of chromatin are present: euchromatin and heterochromatin. The DNA surrounding the euchromatin is lightly packed 
and is often under active transcription. Meanwhile, the DNA around the heterochromatin is more densely packed 
than the DNA around the euchromatin.  Because it is tightly packed, it was assumed that it was inaccessible to polymerases and, 
therefore, not transcribed.
   
Histone modification is the process that causes a block of euchromatin to transform into heterochromatin and vice versa. 
At the beginning of this century, the idea of a ``histone code'' was being discussed as existing in parallel to the genetic code 
\cite{jenuwein2001translating}, 
because combinations of different histones create marks that affect the binding of regulatory factors to a stretch of DNA. 

Ten years ago, it was discovered that the process of histone modification changes depending the nucleosomes. 
In 2008, a study of 40 histone modifications across human gene 
promoters found over 4,000 different combinations used, over 3,000 occurring at a single promoter \cite{wang2008combinatorial}. Therefore, 
the histone code is massively complex and not well understood. Patterns of histone modifications are very intricate, and the biochemical 
paths of only a small number of the existing histone modifications had been fully studied.  

Furthermore, recent publications \cite{volpe2011rna} provide evidence that much of the DNA in the heterochromatin is in fact transcribed, but it
is continuously silenced via a different biological process known as RNA interference. These recent discoveries indicate that there is still
much to be studied and understood about chromatin, histone modification and Epigenetics. 

\subsubsection{DNA methylation}
\label{subsubsec:methylation}

\begin{figure}
\centering
\includegraphics[scale=1.0]{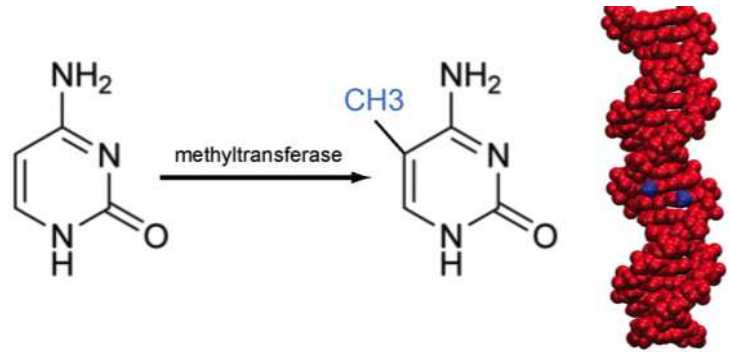}
\caption[DNA methylation]{DNA methylation \cite{grondinraphaelle2017dnamethylation}. The covalent addition of a methyl group to the fifth carbon of a
cytosine base affects the protein transcription, deactivating the transcription of a DNA section.}
\label{dna_mdna}
\end{figure}

DNA methylation is the most intensively studied epigenetic mechanism \cite{breton2017small}. It
involves the covalent addition of a methyl group (-CH$_3$) to the fifth carbon of a cytosine base as displayed in Figure \ref{dna_mdna}, 
generating 5-methylcytosine (5-mC) which occurs predominantly in the context of cytosines that precede guanines (CpGs).

DNA methylation is found in vertebrates, plants, many invertebrates, fungi and bacteria 
\cite{jablonka2014evolution}. 
Methylated cytosine does not change its role in the genetic code and is still paired with guanine. However, the methyl group affects the 
protein transcription either by binding with special proteins and preventing the enzyme RNA polymerase from working on it, or by interfering 
with the binding of regulatory factors of a gene control region.

In other words, cytosine methylation is a mechanism to silence DNA sections. During development, the methylation marks change and during cell division,
the modified (methylated) DNA sequence is transferred from cell to cell.

As it was mentioned before, the first suggestion that DNA methylation (or demethylation) might have an important biological role was made by Griffith and
Mahler, who proposed in 1969 that it can provide the basis for long term memory in the brain \cite{griffith1969dna}. 
 
\subsection{Importance of Epigenetics}
\label{subsec:epigenImp}

Epigenetics has grown to become a widely recognized sub-discipline of biology.
It includes studies of cellular regulatory networks that provide phenotypic stability \cite{alvarez2010from}, 
DNA changes regulated by development such as those identified in the immune system \cite{lim2013epigenetic}, 
cell memory mechanisms involving heritable changes in chromatin and DNA methylation \cite{dodd2007theoretical}, 
and self-propagating properties of some protein conformations and cellular structures \cite{morris2007centromere}. 

A lot of work has been dedicated to the study of cellular inheritance. Several studies around 
the trans-generational inheritance of some epigenetic variations have been published \cite{rakyan2003transgenerational,anway2005epigenetic,manikkam2013plastics,
nilsson2015environmentally}.
One of the productive areas at the end of the last century was the
study of the controlled responses of cells to genomic parasites and changes influenced by the environment \cite{matzke2000host}, 
which involves DNA methylation, RNA mediated gene silencing, 
and enzyme-mediated DNA rearrangements and repair. 

Although the importance of epigenetic inheritance in cell differentiation
and memory processes was recognized since 1970, its influence on macroscopic phenomena has been discussed mostly in 
the last ten years. Examples are environmentally induced epigenetic modification of behaviour \cite{herrera2012jack},
influence of Epigenetics in memory consolidation within generations \cite{day2011epigenetic}, 
inherited propensity for learning \cite{champagne2008maternal}, 
the role of Epigenetics in honeybee morphological differentiation mediated by royal jelly consumption \cite{miklos2011epigenomic}, 
and phenotypic changes in the modern human brain and behaviour compared to other hominids \cite{krubitzer2014evolutionary}.

From an evolutionary perspective, epigenetic strategies allow the adaptation and persistence of a population undergoing environmental
changes until more permanent genetic strategies can be found \cite{carja2017evolutionary}. At the cellular level, the epigenetic 
mechanisms help to build temporary islands of self-determined order to escape from the tendency of the organism components towards 
entropy \cite{rosslenbroich2016properties}. 

The study of the epigenetic mechanisms has helped scientists to understand more clearly the causes and effects 
of some diseases. For example, the human diseases \textit{Creutzfeldt-Jakob disease} (CJD) and kuru, the cattle disease
\textit{bovine spongiform encephalopathy} (BSE), and scrapie in sheep are caused by prion transmissible infectious protein
complexes \cite{prusiner1998prions,chernoff2001mutation}. Prions are unusual epigenetic mechanisms because their
stable inheritance and complex phenotypes are caused by protein folding rather than DNA-associated changes \cite{halfmann2010epigenetics}. 

DNA methylation, histone modification, nucleosome remodelling, and RNA mediated targeting regulate many biological processes that are 
fundamental for the development of cancer. Many tumor cells have aberrant, cell-heritable 
patterns of DNA methylation that are often associated with the silencing of tumor-suppressor genes \cite{baylin2000dna}.
Based on this discovery, different research groups are working on the development and testing of epigenetic drugs for 
cancer treatment \cite{sharma2010epigenetics,dawson2012cancer}.

Several studies showing the role of miRNA in different diseases have been recently published,
particularly allergic diseases such as asthma and atopic dermatitis \cite{chen2015protein,kan2015expression,perry2015role,omran2013micrornas}.
In the last few years, given that epigenetic changes are potentially reversible,
research centres and pharmaceutical laboratories are focusing on discovering epigenetic factors
of diseases and trying to find paths to control, reverse or reduce their effects \cite{oppermann2013epigenetics, moosavi2016role}.

Epigenetics is a dynamic discipline, driving new technological advances as well as challenging and revising traditional paradigms of evolution and biology. 
Thanks to new discoveries, the previous findings on genetics are now explored in different ways. The combination of Genetics and Epigenetics helps to 
understand better the roles and relationships that DNA, RNA, proteins, and environment have in inheritance and the development of diseases. 
The field of Epigenetics is expected to contribute to our understanding of the complex processes associated with gene regulation, cellular differentiation, 
embryology, ageing and disease, but also to systematically explore novel therapeutic methods.

\section{Epigenetics in EA} 
\label{sec:cs_epigen}

Evolutionary Algorithms are bio-inspired optimization methods based on the way biological evolution operates. 
Different modifications have been included in the EAs to incorporate new mechanisms inspired by biological processes.
Some examples of EA modifications inspired by biological mechanisms are migration between sub-populations \cite{kommenda2014genetic}, 
neutrality \cite{keller1996genetic}, role specialization \cite{ferrauto2013different} and 
regulatory networks \cite{banzhaf2003artificial}.  
Some of these modifications have proved to be beneficial in the solution to specific problems; however, the level of improvement 
normally depends on the selected property, the way the mechanism is implemented and the problem selected to test the modification.

Genetic Programming can be used to solve real-world problems where the goal changes during the evolutionary process.
In this context, GP must adapt to continuous changes of the goal and fitness evaluation parameters for the exploration of such dynamic environments. 
However, testing the performance of Genetic Programming in solution to dynamic problem environments is an open issue in the field \cite{oneill2010open} 
and may require modifications to GP standard implementation \cite{yaochu2005evolutionry}.
One approach to face these changes is to generate variable locally adaptable solutions.

As mentioned in Section \ref{sec:epigen}, recent discoveries indicate epigenetic mechanisms have a more active role in 
adaptation and development as part of biological  evolution compared to what it was previously thought.
It has been found that these mechanisms are related to properties such as phenotype differentiation, memory consolidation within generations and 
environmentally induced epigenetic modification of behaviour. These properties have lead different research groups to consider that the implementation of 
epigenetic mechanisms may be beneficial to improve the performance of Evolutionary Algorithms in solution to problems with dynamic environmental changes.  

The first mention of Epigenetics in the EA community was presented by Tanev and Yuta \cite{tanev2004implications}.
They worked with a modification of the predator-prey pursuit problem. GP was used to define a set of stimulus-response 
rules to model the reactive behaviour of predator agents. The implementation
included active and inactive histones in the representation and used age-based predators moving through different life stages (birth, development, survival
and death). An extra step called \textit{Epigenetic Learning} (EL) was included in the fitness evaluation. EL is basically a hill climber acting through epi-mutations
of the histone activation signals.

Although this approach did not include inheritance of epigenetic properties during evolution, 
it was found that the probability of success is larger when the Epigenetic Learning mechanism is included. The authors ascribed the difference to
the robustness gained with the representation by preserving the individuals from the destructive effects of crossover by silencing 
certain genetic combinations and explicitly activating them only when they are most likely to be expressed in corresponding beneficial phenotypic traits.
 
Fontana \cite{fontana2009epigenetic} used other multi-cellular morphogenic models for development with an integer number genetic representation controlled 
by a regulatory network with epigenetic activation and deactivation signals in different development phases. 
A two-dimensional cellular grid and a Genetic Algorithm running on the genome allowed the model to generate predefined 2-dimensional shapes.
 
In \cite{sousa2010epial}, Sousa and Costa presented an epigenetically controlled agent system for Artificial Life. In their experiment, the agents wandered around a
two-dimensional environment with walls and different attributes -temperature, light and food- that can vary over time. The goal of the agents was to survive and
to reproduce.

The behaviour of the agents was coded on binary strings. Activation of genes was controlled by methylation marks. An Evolutionary Algorithm
controlled the survival and reproduction of the different organisms. Several experiments were performed with different levels of epigenetic transference
between parents and offspring. The results showed a significant improvement: Non epigenetic populations found it hard to thrive in dynamic environments, 
while epigenetic populations were able to regulate themselves under dynamic conditions.

Chikumbo et al. \cite{chikumbo2012approximating} proposed a Multi-Objective Evolutionary Algorithm with epigenetic silencing for the land use management problem. 
The goal of the farm algorithm was to reduce the environmental footprint while maintaining a viable farming business through land use and/or 
management changes.

The chromosome encoded each paddock possible use and the system emulated gene regulation with epigenetic silencing based on 
histone modification and RNA editing mechanisms. A Pareto front visualization tool was developed composing the 14 
fitness criteria into 3 super-objectives. However, the approach was not compared with any other classical Multi-Objective Evolutionary Algorithm. 
Therefore, the improvement of the epigenetic variation could not be estimated.

In 2015, the same authors \cite{chikumbo2015triple} extended their previous work using a similar epigenetic based modification. The main modification was
the use of Hyper Radial Visualization, 3D Modelling and Virtual Reality to reduce the 14 fitness functions and display the solutions in a representation 
easier to understand for a group of experts. Again, the approach was not compared with a classical EA.    

Turner et al. \cite{turner2013incorporation} used an Artificial Gene Regulatory model with an epigenetic mechanism based on DNA methylation and chromatin 
modifications. 
The inclusion of epigenetic information gave the network the ability to allocate different genes to different tasks, effectively regulating gene expression 
according to the environment in which it was operating.

The goal of the model was to follow specific trajectories in a chaotic system (Chirikov's standard map \cite{chirikov1979universal}). 
The network was evolved using a Genetic Algorithm. The epigenetic mechanism improved the performance of the model in a dynamic system. With the ability to 
deactivate genes came the ability to increase the efficiency of the network. Hence, with each inactive gene for an objective, there was less computational 
effort required to complete a single iteration of the network simulation.

\textit{Epigenetic Linear Genetic Programming} (ELGP) was presented by La Cava et al. in \cite{la2014evolving}. The method incorporated a binary array 
equivalent in length to the genotype of each individual. This array, called epiline, allowed
the activation and deactivation of the genes. An Epigenetic Hill Climber was used to update the genotype configurations. In the original paper, 
ELGP was used to solve different symbolic regression problems. 

The same method was used to solve program synthesis problems \cite{la2015genetic,la2015inheritable} and as part of an evolutionary 
multi-objective optimization framework to generate models for wind turbines \cite{la2016automatic}. In all these papers, the authors 
reported that ELGP outperformed the GP baseline implementation in terms of fitness minimization, exact solutions, and program sizes.

The approach used by La Cava and Spector is simple and displays good performance when compared to standard GP. However, it is artificial 
and does not represent the way epigenetic mechanisms operate as biological processes. ELGP performs an additional hill climber evaluation
per each individual after the modification of the epiline. There is no equivalent to such event in the biological counterpart.   

Despite the way it is implemented, previous research has demonstrated that the inclusion of epigenetic components in Evolutionary Algorithms provides
robustness. This is a desirable characteristic in dynamic environments where adaptation is required along the evolutionary process.
 % Epigenetics

\chapter{The Proposed Approach}
\label{chap:approach}

  \graphicspath{{figures/PNG/}{figures/PDF/}{figures/EPS/}{figures/}}
  \lhead{\emph{Chpt 3: The Proposed Approach}}  % Set the left side page header to "Proposed Approach"

The previous chapter covered recent developments in evolution and Epigenetics, the biological mechanisms with which they operate, and
different approaches used to implement epigenetic mechanisms in Evolutionary Algorithms.

In biology, epigenetic mechanisms play an important role in memory consolidation within generations and adaptation to environmental changes.
Many of the previous epigenetic analogies in Evolutionary Computation use an artificial representation of these phenomena. 
For example, EL \cite{tanev2004implications} and ELGP \cite{la2016automatic} perform basic hill climbing steps after an epigenetic mutation.
Both implementations use random mutations to generate an alternative solution. The new solution only replaces the original if it
improves the fitness value. The biological equivalent to this hill climbing approach would be the generation of a new cell before every 
epigenetic change, followed by the apoptosis of the less adapted of both cells (original cell and cell affected by epigenetic change).  

Furthermore, these representations are missing an important point of the biological epigenetic mechanisms: their role in adaptation to environmental changes. 

The approach proposed in this chapter is to design a mechanism that considers environmental changes to trigger epigenetic mutations. 
As in the biological counterpart, these mutations do not change the chromosomal information. Instead, they modify additional markers, during the lifespan of
the individual, causing the activation and deactivation of the existing chromosome with the goal of improving the adaptation 
to changes within dynamic environments. 

Although the epigenetic mechanism proposed in this chapter is used in Chapter \ref{chap:synthScenarios}
and Chapter \ref{chap:sumo} for the solution to the traffic signal control problem, the procedures described in 
Section \ref{sec:CodeDeact}, Section \ref{sec:timeWindow} and Section \ref{sec:adptfactor} are general and are 
not associated with a specific problem. The epigenetic mechanism here presented is defined as a context-free modification for tree-based Genetic 
Programming and can be used in the solution of any dynamic optimization problem.

\section{Dynamic Environments}
\label{sec:dynamicEnv}

Before introducing the proposed approach, it is necessary to specify what is considered a dynamic environment in the context of this thesis.

Most optimization algorithms assume a static environment. The goal is to find solutions near to the global optimum with 
respect to a fixed measure or fixed set of measures. However, several real-world optimization problems operate in dynamic environments. Problems with dynamic environments are 
defined as problems where some elements of their domain vary with the progression of time \cite{uy2011semantic}.  
Therefore, the study of mechanisms to allow Evolutionary Algorithms to adapt to changes in the objective function, the problem instance, or the problem 
constraints without restarting the optimization process is required for the solution of dynamic real-world problems.

It is important to differentiate dynamic environments from environments affected by noise. Even if both of them are uncertain environments \cite{yaochu2005evolutionry},
they require totally opposite approaches to generate robust solutions from the perspective of metaheuristic optimization. 
The fitness evaluation of noisy optimization problems is affected by random interference (noise) originated by many different possible sources, such as errors in sensory measurement, 
effect of stochastic components or randomized simulations. In these environments, evaluations with exactly the same parameters may return different 
fitness values because of the effect of noise. 
The goal of an EA in the solution to noisy optimization problems is to be able to neutralize the noise in the fitness function \cite{arnold2001evolution,branke2003selection,fieldsend2015rolling}. 
In other words, heuristic methods should find a way to ignore the noise in the solution of noisy optimization problems. On the other hand, 
the fitness function of problems with dynamic environments is deterministic at any point of time but it does depend of time. The goal of an EA in the 
optimization of such environments is to be able to continuously track the changing optimum. In other words, optimization of problems with dynamic environments
requires the algorithm to identify and adapt to the variations rather than ignore them. 

Mori and Kita \cite{mori2000genetic} identified frequency, severity and predictability of the changes as criteria to be considered when optimization approaches are
defined to solve problems with dynamic environments. 
Frequency measures how often the environment changes. Severity measures how strong these changes are. 
Predictability measures the structure and recurrence of the probable states of the dynamic environment. Random changes are more difficult to optimize than changes
following a pattern.
For example, problems with drastic unpredictable changes occurring in every generation without any type of pattern can be approached in the same way that noisy 
optimization problems are approached. On the other hand, the simplest approach for predictable and clearly identified environmental changes occurring with low variability 
is to consider each change as an independent static problem and execute independent optimization processes on each of them \cite{raman1993jobshop}. 
For problems not at these extremes, the challenge for the Evolutionary Algorithms is to be able to continuously adapt the 
solution to a changing environment, while reusing the information accumulated during previous generations \cite{branke2012evolutionary}.
Hypermutation \cite{cobb1990investigation}, adaptive-mutation \cite{chigozirim2016adaptive}, random immigration \cite{grefenstette1992genetic}, 
memory-based immigration \cite{yang2008genetic}, explicit memory \cite{branke1999memory} and incremental learning with associative memory \cite{yang2008population} 
are examples of modifications implemented in Evolutionary Algorithms to address this challenge. 

The Genetic Programming modification presented in this chapter was designed to be used in the solution to optimization problems with dynamic environments 
composed by smooth environmental changes. Smooth changes are those where the environmental parameters are slightly modified on every time step in contrast with
random modifications with high variability. 

\section{Code Deactivation}
\label{sec:CodeDeact}

Our epigenetic mechanism is inspired by DNA methylation and uses a tree-based representation for Genetic Programming to evolve executable programs.
The mechanism only affects a specific type of node in a similar way that methylation only affects cytosine bases of the DNA. 
In all the examples of this thesis, the node affected is a \textit{conditional node}. However, in further implementations, it could be associated 
with any node representing a binary function such as addition, multiplication, logical conjunction or logical disjunction. 

\begin{figure}
\centering
\includegraphics[scale=1.0]{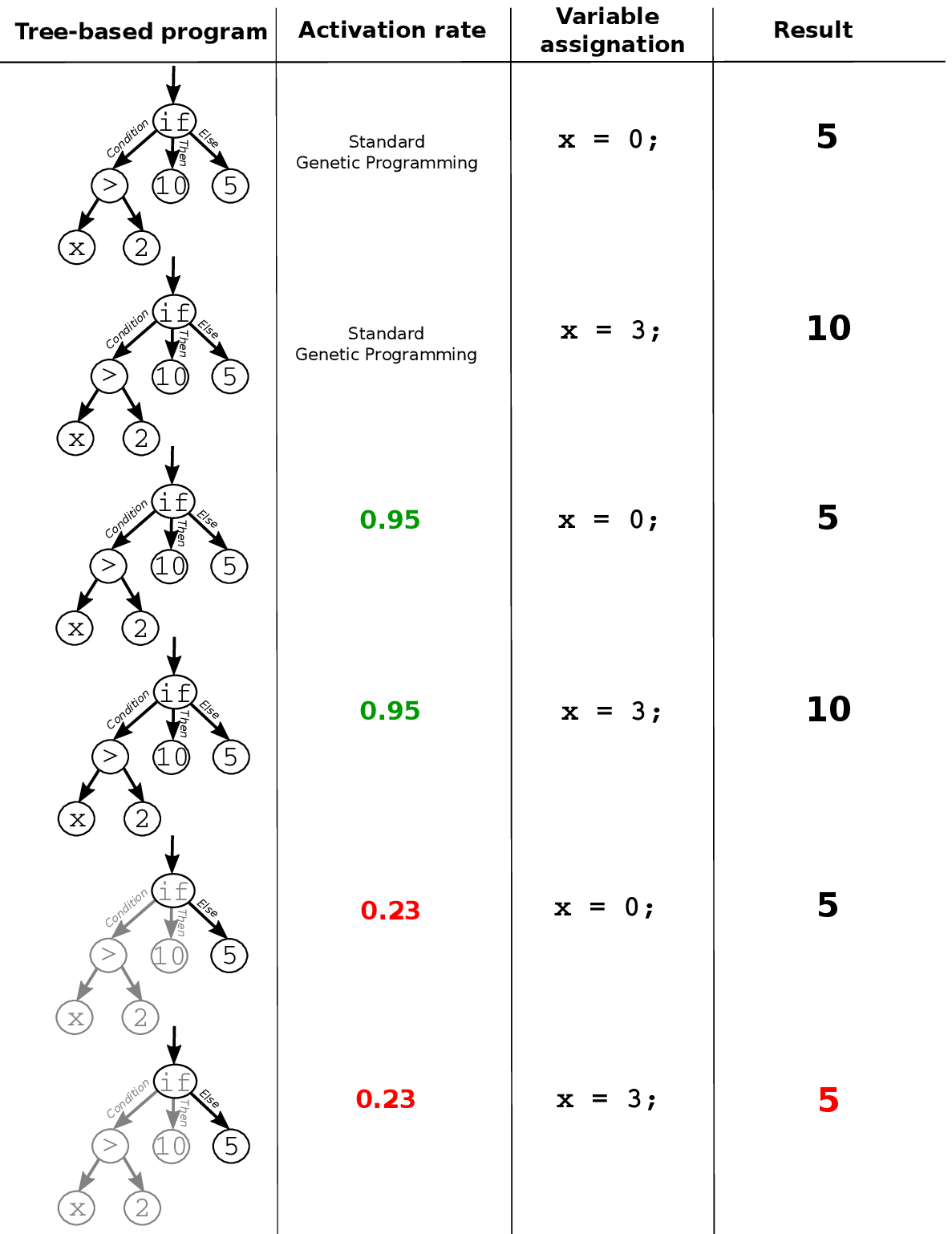}
\caption[Basic tree-based program affected by activation rate]{Basic tree-based program affected by activation rate. Activation rates in red 
are under the threshold. Activation rates in green are above the threshold. Nodes in grey are inactive.}
\label{basicChromosome}
\end{figure}

Each node of the selected type is associated with an activation rate in analogy
to the concentration of methyl groups attached to cytosine nucleotides along the DNA structure. As in the biological counterpart, the 
structure of the chromosome is not affected by the activation rate. 
However, during the evaluation step, if the activation rate is less than
a given threshold, defined by default as $50\%$, the \textit{conditional node} is ignored and only the 
\textit{else sub-tree} is executed, deactivating with that action the \textit{conditional sub-tree} and the \textit{then sub-tree} of the node.
Figure \ref{basicChromosome} presents a basic example of the change in the behaviour of a program using the activation rate.

All the activation rates of the individuals of the first generation are randomly initialized between $0\%$ and $100\%$. 
The activation rates can be affected by epigenetic mutations during the individual evaluation. These
mutations do not change the program structure, but they can change its behaviour as presented in Figure \ref{basicChromosome}. 
The process to trigger epigenetic mutations is described in Section \ref{sec:adptfactor}.
The activation rates are transferred to the offspring as part of the crossover operation in the same way methylated DNA is transferred between generations.

\section{Time Interval Window Configuration}
\label{sec:timeWindow}

So far, our approach is not really different from previous work in the field: a mechanism to activate and deactivate sections of the
chromosome is proposed, the mechanism is independent of the chromosome, the changes are transferred to offspring, and a 
procedure is used to change the chromosomal epigenetic markers.

\begin{figure}
\centering
\includegraphics[scale=0.6]{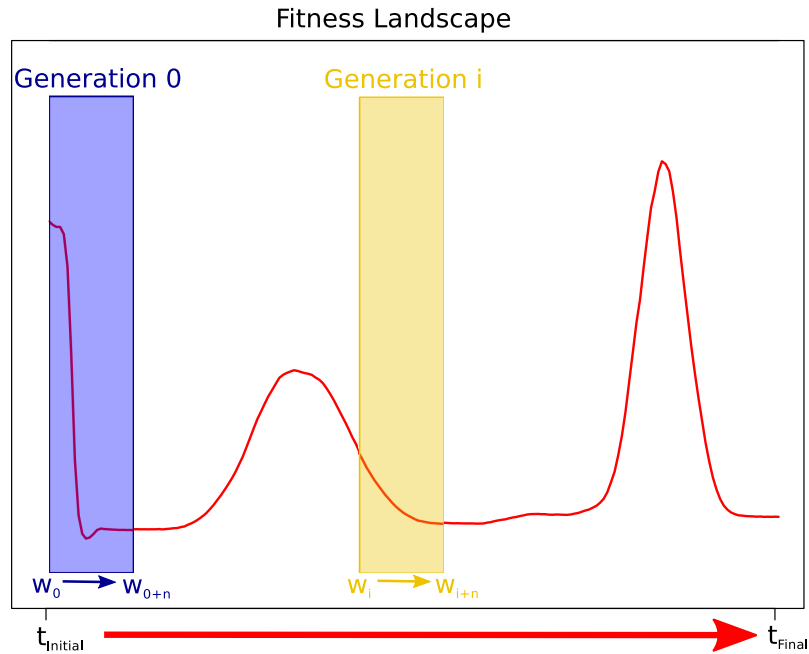}
\caption[Dynamic landscape with a time interval window defined by generation]{Dynamic landscape with a time interval window defined by generation. The $x$-axis 
of the plot corresponds to the time series used to evaluate the fitness landscape. The 
$y$-axis corresponds to the value of the independent variable $y$ defined in \ref{sec:adptfactor}. The blue rectangle represents the subsection of
the landscape used to evaluate the initial generation. The yellow rectangle represents the subsection of the landscape used to evaluate a generation $i>0$.}
\label{movWindow}
\end{figure}

Suppose a dynamic environment, such as those described in Section \ref{sec:dynamicEnv}, is used as fitness landscape. 
The evaluation of each individual can traverse multiple sequential steps in the landscape 
instead of a single step. In other words, all the individuals of each generation are evaluated using the same time interval of a movable time window in the 
dynamic fitness landscape.
Therefore, the environment changes throughout the lifespan of each individual evaluation and a time window moves along the
fitness landscape across the evolutionary process. Figure \ref{movWindow} presents a diagram of this configuration.

The time interval window defines the section of the fitness landscape being used to evaluate a specific generation of programs in the evolution
of solutions for a dynamic environment. It is important to remark that each generation will be executed using a different section 
of the fitness landscape. The use of a changing landscape hinders the adaptability of Genetic Programming, 
but it may be necessary in the solution to complex nonlinear problems where the evolutionary timescale is too slow to adapt
to the changes in the environment. Using this approach, Evolutionary Algorithms may be able to provide better solutions to problems 
with high variability like traffic control or weather forecasting.  

\section{The Adaptive Factor}
\label{sec:adptfactor}

Using the time interval window configuration described in Section \ref{sec:timeWindow}, an individual with epigenetic activation rates 
executes multiple evaluations of the evolved program. The environmental conditions can change for each evaluation.  
If a dramatic change happens in the environment during the lifespan of the individual, the program may not be able to provide a good 
solution for the updated environmental conditions even if it was a good solution for its initial configuration.
In that case, a modification to the program can help the system to adapt to the environmental changes in the same way that
inherited epigenetic changes allow biological populations to adapt to new environments. In the context of dynamic environments, 
while the problem is the same over long time scales (e.g. traffic control), the specific conditions change so fast that rules 
need to switch as well. 

A heuristic to measure environmental change is required to be able to implement this mechanism. We propose a simple approach called \textit{adaptive factor}.
It requires selecting an independent variable of the problem. For each program evaluation, the adaptive factor compares the value of the 
independent variable with the average of the same variable over the previous $n$ evaluations, where $n$ is a predefined interval size. 
The adaptive factor is calculated using Equation \ref{eq:adaptFactor}, where $y$ represents the independent variable, $i$ is the current evaluation 
step and $\overline{y}_{i}^n$ is the average over the interval 
\begin{equation} \label{eq:adaptFactor}
\lambda_i(y) = \frac{|{y}_{i}-\overline{y}_{i}^n|}{\overline{y}_{i}^n}.
\end{equation}
The average of the interval is calculated using Equation \ref{eq:meanWindowValue}, where $n$ is the predefined interval size and $i$ is the 
current evaluation step
\begin{equation} \label{eq:meanWindowValue}
\overline{y}_{i}^n = \frac{\sum\limits_{t=i-n}^{i} {y}_{t}}{n}.
\end{equation}
A normalization to the range $[0,1]$ is performed to the interval adaptive factor using Equation \ref{eq:normAdaptFactor}, where  
$\lambda_i(y)$ represents the interval adaptive factor calculated with Equation \ref{eq:adaptFactor} and $\lambda_{max}(y)$ represents the 
maximum value that $\lambda_i(y)$ can have during all the experiment 
\begin{equation} \label{eq:normAdaptFactor}
\lambda'_i(y) = \frac{\lambda_i(y)}{\lambda_{max}(y)}.
\end{equation}

The goal of the adaptive factor is to identify differences between the interval average value and the current value. 
A big difference indicates a change in the environment.

The normalized adaptive factor $\lambda'_i(y)$ is used as the probability in a mutation of the activation rates of  
the chromosome or \textit{epi-mutation}. At the end of the window interval, an epi-mutation based on the normalized adaptive factor is performed on the activation rates. 
This step is performed as an internal modification during the evaluation process of the individuals. 

The mechanism composed by the code activation rates described in Section \ref{sec:CodeDeact} and the epigenetic mutations triggered by
the adaptive factor here described works as an adaptation mechanism to the environmental changes without requiring an explicit hill-climber. 

At the end of the individual evaluation, the final activation rates are stored in the chromosome markers and transferred to the next generation. 
This process transfers the environmental information collected during the lifespan of the individual to the next generation in a similar way 
to the way epigenetic information is transferred at cellular level or inherited at population level in biological systems.

\section{Discussion}
\label{sec:epigenDisc}

This chapter presented a general modification to Genetic Programming inspired by recent discoveries in Epigenetics. 
The proposed mechanism is based on the DNA methylation process. It silences sections of the chromosome using
additional markers. Those markers are modified based on changes to environmental conditions during the lifespan of the
individual. The markers are transferred to the individual's offspring in a similar way to the way DNA methylation markers 
are transferred during cellular division.  

The proposed mechanism differs from previous work (see Chapter \ref{sec:cs_epigen}) 
in the incorporation of environmental components to the modification of the epigenetic markers. Our final objective is to use the epigenetic 
mechanism as a way to build temporary islands to allow adaptation to environments that are constantly changing. 
As mentioned in Chapter \ref{subsec:epigenImp}, this is one of the discovered advantages of the biological epigenetic mechanisms.

Our hypothesis is that the incorporation of the epigenetic mechanism presented in this chapter will allow Genetic Programming to provide better solutions 
to problems with dynamic environments composed of smooth environmental changes occurring with medium or high frequency, such as traffic control or weather 
forecasting. The approach presented in this chapter is general and can be used in the solution to any problem with a dynamic environment.
 % Proposed Approach

\chapter{Traffic Signal Control}
\label{chap:application}

  \graphicspath{{figures/PNG/}{figures/PDF/}{figures/EPS/}{figures/}}
  \lhead{\emph{Chpt 4: Traffic Signal Control}}  % Set the left side page header to "Real-World Application"

As mentioned in Chapter \ref{chap:intro}, traffic congestion is a global phenomenon. It impacts businesses as well as commuters in small cities and
 large ones in developing as well as developed economies. Traffic congestion cost has direct and indirect components. Direct costs include the amount of time 
commuters spend needlessly in congestion, fuel cost, social cost and the environmental cost of emissions released by the vehicles. The 
increase of prices of goods and services due to congestion is an example of its indirect costs. The traffic congestion cost for USA, UK and 
Germany during 2017 was almost \$461 billion dollars \cite{cookson2017inrix}.

Any attempt to address the increasing traffic congestion problem should consider multiple components: improving public transportation systems, promoting
alternative transportation methods and policies (car sharing, bicycles, etc.), increasing the transportation infrastructure (new roads, roundabouts, more lanes 
for exisiting roads, etc.) and improving the efficiency of the existing components. One approach to improve the current infrastructure is to optimize the traffic signal 
behaviour in order to be adaptive to changes of the traffic conditions. However, urban traffic signal control is a large combinatorial problem with high 
complexity and uncertainty \cite{yang1996model,zhao2012computational}.

Furthermore, experimentation with live traffic flows is not possible when we talk about traffic congestion because of safety concerns as well as the economic 
and social costs associated. The development of realistic traffic simulators originated with the goal of allowing the analysis and manipulation of traffic 
scenarios.

This chapter presents a general description of the traffic signal control problem and 
explores different solutions based on adaptive controllers, with an emphasis on Evolutionary Algorithms. 
In addition, this chapter presents a brief introduction to traffic simulation.

\section{The Traffic Signal Control Problem}
\label{sec:signal}

Traffic signals at intersections are the major control mechanism for urban traffic networks \cite{papageorgiou2003review}. They were originally 
installed to resolve conflicts between antagonistic streams of vehicles and pedestrians. The capacity of traffic signals to dynamically change
the priority for vehicles approaching intersections from different directions becomes crucial when traffic networks are saturated. 
Some signal configurations reduce the amount of time spent by vehicles in intersections compared to other signal configurations. 
Therefore, an optimal control strategy to minimize the total time spent by all vehicles in the network should exist.

The traffic signal control problem consists of finding the optimal signal control strategy to reduce stops, overall vehicle delay and maximize throughput
while maintaining driver safety in a traffic network composed by a single intersection, a street, multiple intersections in the same area, or an entire city.

Different approaches have been used to describe the traffic signal control problem using formal models. Haijema et al. \cite{haijema2017dynamic} represented 
the problem of minimizing vehicle delay at an isolated intersection as a Markov Decision Process \cite{bellman1957markovian}. Although only a single intersection 
was considered, the model was too complex to be solved numerically and an approximation policy was defined. The policy presented improvements over a fixed time schedule, 
but it was not compared with other adaptive solutions. 

In \cite{abu2003design}, Abu-Lebdeh and Benekohal defined the behaviour of a single street with multiple intersections as a multi-variable dynamic optimization 
problem with twelve constraints. They used Genetic Algorithms  to explore the problem using different strategies. The results presented were promising, but the 
methods were not compared with other approaches nor tested in a simulated environment. 

The solution to the traffic signal control problem of real traffic networks features several difficulties.
The traffic signals switch between green, yellow and red lights. This indicates the problem can be addressed using discrete variables, which renders it 
as combinatorial \cite{papageorgiou2003review}.
However, the size of the problem is very large when more than one intersection is considered.
Furthermore, the traffic network may be perturbed by multiple unpredictable and hardly measurable disturbances, such as accidents, illegal parking and road blocks.
Finally, the problem includes pressing real-time constraints. For example, changes to the system must be performed within a few seconds for real-time implementations 
and, given the speed of the vehicles, such decisions may imply mortal risk for passengers or pedestrians.

Urban traffic signal control is a large combinatorial problem with high complexity and uncertainty \cite{yang1996model,zhao2012computational}. 
Traffic congestion considerably affects daily life of many citizens. Furthermore, 
the fast growth of metropolitan populations makes control of traffic signals an increasingly challenging task.
Because of its complexity, the solution to the traffic signal control problem is not feasible for networks with more than one intersection \cite{papageorgiou2003review}.
Therefore, proposed control strategies for traffic signal control, such as the solution presented in \cite{abu2003design},
introduce several simplifications to partially address the optimization of traffic signals in simple networks only composed of a few intersections.
Because of this limitation, solutions to more complex real scenarios require the use of microscopic simulators (see Section \ref{sec:simulation}) 
and heuristic methods.

Two different approaches can be followed to address traffic signal control of real networks: pre-timed control and actuated control.
Pre-timed traffic control assumes the traffic has a periodic behaviour and optimizes fixed time schedules to operate traffic signals based on historical data.
On the other hand, actuated traffic control uses data from traffic sensors and control strategies to dynamically modify the traffic signals in real-time manner. 

\subsection{Pre-timed solutions}
\label{subsec:pretimed}

The optimization of time schedules is one of the traditional solutions to the traffic signal control problem. 
Pre-timed control or optimized traffic signal schedules consists of a sequence of traffic light phases that have fixed duration. 
Collectively, the preset green, yellow, and red intervals result in a deterministic sequence of phases with fixed cycle duration for the 
intersections of a traffic network assigned by time of the day or day of the week. The design of optimal signal schedules can be addressed by analysis 
of historical data, expected volume and turning movements, simulation of different conditions in traffic scenarios or using hybrid approaches. 
This operation can be performed through expert analysis or heuristic methods. 

However, given that traffic congestion is a dynamic phenomenon, this approach has the downside of requiring constant updates
of the schedules. Furthermore, it requires constant monitoring and design of contingency plans to react to atypical congestion events.

\textit{Traffic network study tool} (TRANSYT) \cite{robertson1969transyt} is one of the most popular programs for off-line optimization of pre-timed signal schedule plans. 
The latest version of the program features Genetic Algorithm optimization of cycle duration, phasing sequence, splits, and offsets. TRANSYT can compile
series of fixed-time signal schedules for different hours of the day or for special recurring traffic conditions. However, because traffic is a dynamic system, any
predefined traffic signal plan cannot adapt well to real traffic conditions \cite{zhao2012computational}.

In \cite{zhang2009multi}, Zhang et al. proposed a real-time online urban traffic signal control approach using a multi-objective discrete differential 
evolution method to optimize the light phase intervals of a single lane bidirectional network with two intersections including left-turn phases. 
The authors compared their algorithm with a pre-timed controller using a Poisson distribution to regulate the traffic flow. 
The proposed approach provided better results than the pre-timed controller for several single intersection scenarios.

Nie et al. \cite{nie2010based} used a two-dimensional Cellular Automaton and a $1+\lambda$ Evolutionary Strategy to update the time parameters of 
CA rules in a network composed by a grid of $20\times20$ intersections. The authors performed experiments with different traffic densities and the results demonstrated 
a better performance for the evolutionary approach compared to previous work done with the same Cellular Automaton. However, the simulated environment was too rigid and was not 
able to represent all the conditions of a real network.

S\'anchez-Medina et al. \cite{sanchez2010traffic} used a traffic simulator based on Cellular Automata and a Genetic Algorithm optimization technique to simulate 
and optimize the duration of traffic light phase intervals of ``La Almoraza'', a section of Saragossa city in Spain. The network has seven intersections, 
16 entry nodes, 18 exit nodes and 17 traffic signals.

Individual strategies were represented as an array containing the duration of the light phases for all the traffic signals of the network. The algorithm was tested with different 
traffic conditions and limited results were obtained. The methodology did not provide a significant improvement for the regular traffic conditions of 
the network. However, it increased the performance for scenarios with saturation over the congestion levels registered in the real world.

Li et al. \cite{li2016traffic} used Reinforcement Learning to train a Deep Neural Network to design signal timing schedules. They used a basic scenario 
consisting on a single intersection of two-directional roads with two lanes. Dynamic entry distributions were used to simulate two traffic waves 
during two different rush hours on a simulated day. The results presented an improvement of 14\% over the conventional reinforcement learning method. 
However, the scenario was too simple to provide meaningful conclusions.

\subsection{Actuated solutions}
\label{subsec:actuated}

Actuated traffic signal control or traffic-responsive signal control consists of multiple red-green configuration stages with variable duration that can be 
modified in response to real time measurements from traffic sensors installed at intersections. Detection is used to provide information about traffic demand to a 
traffic controller at each intersection. The duration of each phase is determined by data collected from the sensors, the controller parameters and the controller logic. 
Some authors \cite{wiering2004intelligent, zhao2012computational, younes2016intelligent} use the term \textit{intelligent} traffic signal control when an Artificial Intelligence 
model is used as the control algorithm in an actuated traffic signal control. 

Different approaches have been used to generate actuated traffic controllers. This section presents some of the recent publications
related to the topic.

\textit{Sydney Coordinated Adaptive Traffic System} (SCATS) \cite{sims1980sydney} is a plan-selection control system to manage the dynamic intervals of light phases
at traffic signals using an online approach. SCATS uses adaptive traffic control strategies to match the current traffic phases to the best signal schedule among 
the available predefined signal schedules based on the overall requirements of the system using online information measured by detectors connected to the on-street 
traffic controllers.

\textit{Urban Traffic Optimization by Integrated Automation} (UTOPIA) \cite{mauro1990utopia} is a distributed real-time traffic control system.
It optimizes traffic signal schedules and sorts the traffic light phases to satisfy traffic demand. In the fully adaptive mode, 
UTOPIA constantly monitors and forecasts the traffic status and optimizes the control strategy according to flow efficiency and/or 
environmental criteria.

\textit{BALancing Adaptive Network Control mEthod} (BALANCE) \cite{friedrich1995balance} is a decentralized macroscopic traffic model that estimates 
flows according to detector data, a control model and a mesoscopic traffic flow model to calculate the effect of a specific signal 
plan and different optimization algorithms. The following factors can be considered in the decision process: vehicle delay, number of vehicle stops
and queue length of the roads preceding intersections. Different weights can be associated to these factors depending on the goal to achieve.
BALANCE includes a centralized control module to perform event management tasks to handle atypical behaviours of the traffic network.  

In \cite{braun2011evolutionary}, Braum and Kemper modified BALANCE. 
They replaced the hill-climbing algorithm used at the tactical 
level of the solution with a Genetic Algorithm. Several experiments were performed with the traffic network of Ingolstadt, Germany. 
The results demonstrated a better performance over the original BALANCE. The system was implemented in the real-world and 
daily average delays were reduced by 21\%.

Padmasiri and Ranasinghe \cite{padmasiri2014genetic} used a Genetic Programming and fuzzy logic hybrid approach to define fuzzy rules for a 
scenario with a single intersection under the effect of different traffic densities. 
The set of evolved rules used traffic parameters as input and decided between extending or terminating green phases. 
Although the results presented an improvement over previous work, the study only considered a single intersection and the solutions lacked 
adaptability to changes when the traffic congestion was high.

Zubillaga et al. \cite{zubillaga2014measuring} proposed a set of self-organizing rules to coordinate urban traffic. They compared their approach 
with the green wave method \cite{torok1996green} in a network composed by a grid of $10\times10$ intersections under different traffic densities. 
The average velocity of the vehicles traversing the network was higher when the self-organized method was used.      

In \cite{covell2015micro}, an auction-based method was proposed to coordinate phase switching operations using local induction loop information.
The method was tested in a scenario generated with real-world traffic data from the Mountain View, California area. 
Auction-based control performed better than a fixed time schedule and a planning-based method.

The back-pressure routing algorithm  \cite{tassiulas1990stability} is a method for directing traffic around a queueing network designed to make decisions that 
(roughly) minimize the sum of squares of queue backlogs in the network from one time slot to the next. 
Yuan et al. \cite{yuan2016optimal} proposed a dynamic slot time mechanism to improve the performance of the original back-pressure routing algorithm 
in a traffic network. The mechanism was tested in an artificial network composed by a grid of $4\times4$ intersections with a specific traffic density. 
Partially successful results were reported over the original back-pressure algorithm. 

Yang and Ding \cite{yang2016actuated} proposed a breadth-first-search approach to the green wave method using gap-outs and extensions. 
The method was compared with a self-organized algorithm in a traffic network representing a section of Qidong, China. The green wave
method worked better than the self-organized method under highly saturated traffic conditions. 

One of the particularities of the traffic signal control problem from the computer science perspective is the lack of standard benchmark 
methods \cite{balaji2010urban, gokulan2010distributed, li2016traffic}. There are two different comparison approaches followed in the papers 
of the literature review here presented: compare the proposed method with fixed time schedules in a basic scenario consisting of a few intersections; or
compare the proposed method with a commercial solution in a network representing a real world scenario. However, commercial solutions 
require expensive licenses in order to be used or tested, and their performance varies greatly from one version to other.
Therefore, commercial solutions cannot be used as a reliable metric to compare the performance of different algorithms.   

Even if standard benchmark methods do not exist for traffic signal control, several studies used simple actuated 
controllers as a way to compare their solutions with methods more adaptive than fixed schedules. These controllers are easy to implement and 
independent of the network topology. Two of these controllers are introduced in Section \ref{subsubsec:longqueue} and Section \ref{subsubsec:timegap}. 

\subsubsection{Longest queue first scheduling algorithm}
\label{subsubsec:longqueue}

\textit{Longest Queue First with Maximal Weight Matching} (LQF-MWM) is an algorithm for scheduling traffic signals at an isolated intersection proposed by 
Wunderlich et al. \cite{wunderlich2007stable}. It was designed to maximize the traffic throughput while minimizing the average latency experienced by the traversing 
vehicles of a single intersection. The algorithm is based in a similar technique used for data packet switching in computer networks \cite{dimakis2006sufficient}. 

LQF-MWM requires the identification of non-conflicting traffic signal phase combinations and constant measurement of the size of queues associated with each inbound 
lane of the intersection. Instead of affecting the duration of the phases, the method alters the phase sequence based on the size of the queues associated with
each phase. In other words, the phases have no particular order in LQ-MWM, and are actuated based on the queue sizes.

The paper by Wonderlich, Elhanany and Urbanik \cite{wunderlich2007stable} presented a comparison of the longest queue algorithm with a fixed-time controller 
over simulations with different traffic densities as well as stability proof for a single intersection. LQF-MWM performed significantly better in terms of average
delay than the fixed-time controller for the scenario tested.  

LQF-MWM has been used to compare the performance of other solutions for signal optimization of traffic networks with different sizes because it is simple, stable,
and easy to implement.
The methods compared to LQF-MWM in the literature include ant colony systems \cite{wu2012cooperative}, dynamic traffic control \cite{chen2013dynamic}, 
case-based reasoning \cite{luati2016casebased}, green wave \cite{chen2017cooperative} and artificial immune system \cite{darmoul2017multiagent}. 

\subsubsection{Time-gap based approaches}
\label{subsubsec:timegap}

During the decade of 1980s, several actuated traffic control solutions based on the measurement of time gaps between vehicles approaching intersections were 
proposed. Two examples are the traffic regulation systems LHOVRA \cite{peterson1986lhovra} 
and MOVA \cite{vincent1988mova}, both still widely used, especially in Nordic countries such as Sweden \cite{lindorfer2017modeling}.

Actuated traffic controllers based on time gaps require the installation of multiple vehicle detectors on inbound roads of controlled intersections.
The number of detectors and distance between them depends on the method used and the traffic regulations of the country where it is implemented. 
The controllers work by prolonging traffic phases whenever a continuous stream of traffic is detected and affecting the cycle duration in response to 
measurements of dynamic traffic conditions.

Although, these approaches use real-time measurements to adapt the intersection behaviour to traffic flow changes, their strategies were predefined to control 
isolated intersections based on human expert analysis. As in expert system, there is a possibility these decision strategies face cases
not addressed in their original design. Furthermore, methods designed to operate on isolated intersections are not necessarily the optimal solution 
for interconnected and complex real traffic networks.

SUMO traffic simulation suite (see Section \ref{subsec:sumoRev}) includes a generic actuated control method based on time gaps. 
This method is commonly used to compare the performance of other solutions tested in SUMO \cite{benhamza2015adaptive,jin2015adaptive,liang2018realtime} because of the 
popularity that time-gap actuated control strategies still have and also because the method is already implemented within the simulator.

\section{Traffic Simulation}
\label{sec:simulation}

Analytical models and simulation models are two common approaches to perform traffic analysis.
Examples of analytical models used to represent the behaviour of traffic scenarios are queueing theory, 
optimization theory and differential equations \cite{askerud2017evaluation}.

Simulation is the process of building a computer model that suitably represents a real or proposed system 
which enables to extract valid inferences on the behaviour of the modelled system, 
from the outcomes of the computer experiments conducted on its model \cite{barcelo1998parallelization}.
Traffic simulation is a practical and efficient tool to analyze traffic scenarios. 
The simulation of traffic scenarios provides the opportunity to safely evaluate the effect of changes in the network, 
changes in traffic conditions or even environmental changes \cite{lu2016modelling}.
The analysis through computer simulation of traffic networks is a widely used approach because experimenting 
with traffic in the real world is neither safe nor practical.

Furthermore, the increasing power of computers has led to the emergence of a number of
different methods and software packages as practical traffic analysis tools.
Based on the level of detail used to describe the traffic state, traffic simulators can be classified into three groups: 
macroscopic models, mesoscopic models and microscopic models. 

Macroscopic models deal with aggregated characteristics of transportation elements, such as aggregated traffic flow dynamics and zonal-level 
travel demand analysis. Flow equations are normally used by macroscopic models, and they cannot provide information about individual components. 
Macroscopic models are mainly used for the simulation of highways. Examples of macroscopic traffic models are traffic flow models \cite{van2015genealogy} 
and the freeway traffic model \cite{kuhne1991macroscopic}.

Mesoscopic models analyze transportation elements grouped in small sets, where the elements contained in a set are considered homogeneous. 
A mesoscopic model does not model each individual vehicle, but it may still maintain information about the vehicles in the system.
Examples of mesoscopic traffic models are vehicle platoon dynamics \cite{wang2008high} and discrete-event time resolution \cite{burghout2006discrete}. 

Microscopic models study individual elements of transportation systems, such as dynamics of individual vehicles and behaviour of drivers. 
These models depend on random numbers to generate vehicles, select routes and determine the behaviour of each component of traffic systems. 
In a microscopic simulator, the dynamic variables of the model represent microscopic properties such as the position and the velocity of individual vehicles. 
Examples of methods used to represent microscopic traffic simulators are cellular automata \cite{nagel1992cellular} and multi-agent 
systems \cite{balmer2004towards}.

\subsection{Microscopic traffic simulators}
\label{subsec:microscopic}

Microscopic urban traffic simulators replicate the behaviour of real-world traffic networks by the simulation of the movement of individual vehicles  
in the network. Although microscopic models are usually associated with high computational costs, their representation of the systems is the closest 
to the real phenomena. 

In the last 20 years, a new type of microscopic models has been developed. These models are based on the cellular automata programming 
paradigm from statistical physics \cite{maerivoet2005cellular}. The advantage of these models over more realistic ones is their efficiency
and fast performance when used in computer simulations. 

\textit{Cellular automata} (CAs) are discrete mathematical models that can be used to represent the behaviour of self-organizing statistical systems. 
They became popular in the decade of 1980s through the work of Stephen Wolfram \cite{wolfram2002new}. 
Wolfram used CA to model different processes of several scientific disciplines. It was Wolfram who used cellular automata to model traffic 
flow for the first time. The CA rule 184 \cite{wolfram1984computation} is a one-dimensional CA with binary states that can be used to model the 
behaviour of freeway traffic flow of one-lane one-direction linear roads.

In 1992, Nagel and Schreckenberg \cite{nagel1992cellular} proposed a one-dimensional stochastic CA model to simulate different traffic conditions. 
Their model is composed by 4 rules (acceleration, break, randomization and car motion) and is able to reproduce several characteristics of
real-life traffic flow. For example, the spontaneous emergence of traffic jams. The CA proposed by Nagel and Schreckenberg has been widely
recognized as one of the simplest models to represent traffic congestion.

Traffic modelling through CA is called \textit{Traffic Cellular Automata} (TCAs). TCA is an active research field to this day. A good review
of the first 10 years of work in the field can be found in \cite{maerivoet2005cellular}. TCAs are dynamic systems discrete in time and space.
In order to achieve the discretization required by the Cellular Automaton, a road is divided into sections of a certain length, 
time is discretized to steps and each road section can either be occupied by a particle, called a vehicle, or empty. 

Although TCA models lack the accuracy of the time-continuous car-following models, they still have the ability to reproduce a wide range 
of traffic phenomena. Due to the simplicity of these models, they are very efficient in terms of computational time and can be employed to 
simulate large complex networks.

\subsection{Available simulators}
\label{subsec:sim_compare}

A rigorous comparison of all the different available simulation packages would be a time consuming task and is beyond the scope of this 
research. However, several partial reviews have been done in the last 10 years \cite{kotusevski2009review,ratrout2009comparative,pell2017trends}.
Based on these previous studies, we decided to focus our analysis on the following four simulators.

\subsubsection{CORSIM}
\label{subsec:corsimRev} 

\textit{CORridor SIMulation} (CORSIM) is a microscopic, stochastic traffic simulation program designed for the analysis of
freeways, urban roads and traffic corridors. Its first version was developed at the beginning of the 1970s decade under the
direction of the \textit{Federal Highway Administration} (FHA), part of the USA Department of Transportation. It is currently
maintained by McTrans Center of the University of Florida, but the latest version was released in 2010 \cite{mcTrans2010corsim}. 

CORSIM makes use of two different microscopic models: FRESIM represents traffic on highways and freeways and 
NETSIM represents traffic on urban streets. CORSIM uses car-following models to control the behaviour of the vehicles
and includes logic to allow lane-changing events. 

Although some of the surveys reported that CORSIM provides full functionality for a broad type of scenarios \cite{ratrout2009comparative,pell2017trends}, 
a trial version is not available in the official website. It does not even provide a link to a paid version without a license number
of the software. For these reasons, we were not able to test this simulator.

\subsubsection{AIMSUN}
\label{subsec:aimsunRev} 

\begin{figure}
	\centering
	\includegraphics[scale=0.4]{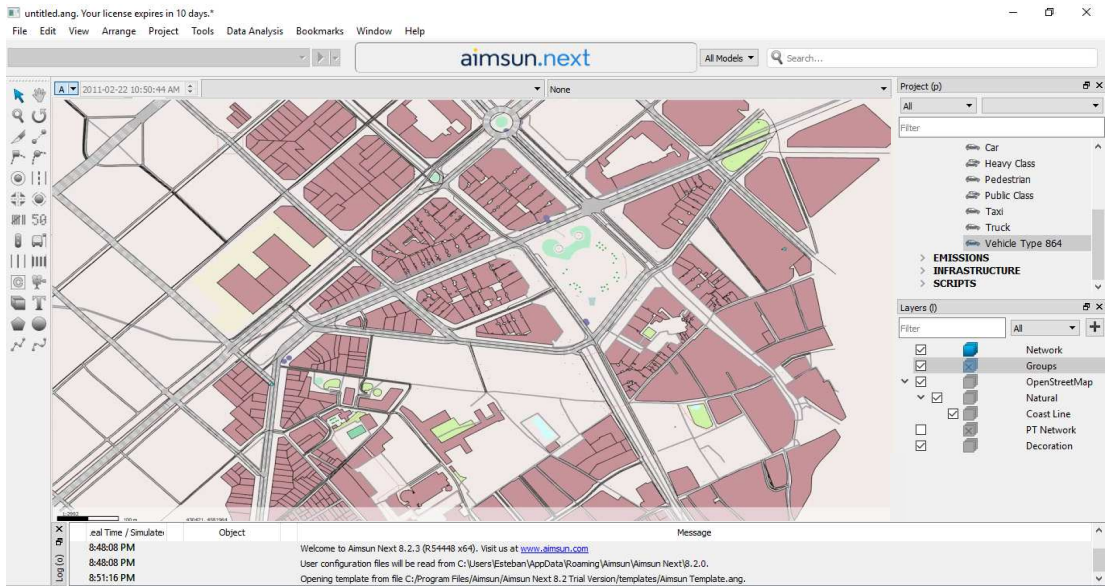}
	\caption[AIMSUN screen-shot]{AIMSUN version 8.2 screen-shot.}
	\label{aimsun}
\end{figure}

\textit{Advanced Interactive Micro-Simulation for Urban and Non-Urban Networks} (AIMSUN) is a commercial simulation package.
AIMSUN integrates different types of simulation models: travel demand modelling, macroscopic functionalities and a hybrid 
mesoscopic and microscopic simulator \cite{aimsun2018aimsun}.

The microscopic model was developed based on car-following, lane-changing, and gap acceptance algorithms. The mesoscopic component 
provides additional functionality to model dynamic aspects of very large networks and to reduce the calibration required
for individual vehicles. AIMSUN can simulate street networks, freeways, interchanges, weaving sections, 
roundabouts, pre-timed and actuated signals. 

The official website provides a trial version. The software does not include basic scenarios, but it allows 
to import traffic scenarios from other commercial simulators. Although it is only available for Windows, the simulation package is 
complete and includes multiple options to customize the properties of the vehicles and the behaviour of the drivers. 
However, it lacks components to include pedestrian models. The macroscopic information displayed while the 
simulator is executed is one of the highlights of AIMSUN. Figure \ref{aimsun} presents the user interface of this simulation package.

\subsubsection{VISSIM}
\label{subsec:vissimRev} 

\begin{figure}
	\centering
	\includegraphics[scale=0.4]{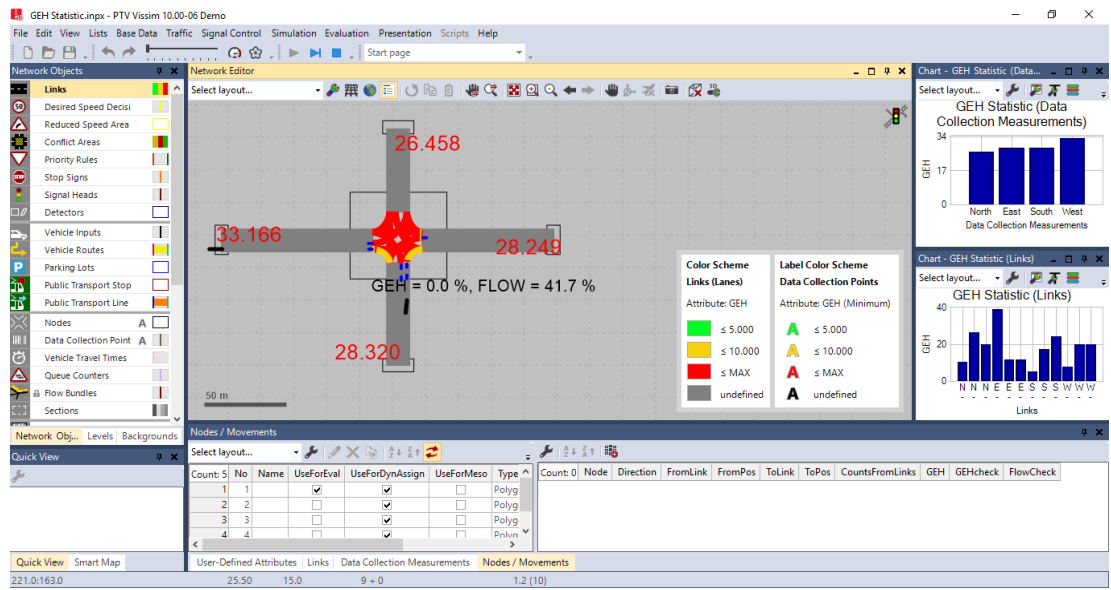}
	\caption[VISSIM screen-shot]{VISSIM version 10 screen-shot.}
	\label{vissim}
\end{figure}

VISSIM stands for \textit{Verkehr In St{\"a}dten - SIMulationsmodell} (Traffic in cities - simulation model). It is 
a discrete, stochastic, and time step microscopic simulator based on a traffic flow model. The model considers
drivers and vehicles as independent entities. VISSIM uses a car-following model for longitudinal vehicle movement and a
rule-based algorithm for lane-changing events. 

VISSIM controls the signals with an independent module called signal state generator. This module is an independent signal control 
software polling detector which retrieves information from the traffic simulator using discrete time steps. Traffic
signals are modelled independently of the main simulator, which facilitates their manipulation. 

The official website \cite{ptv2010vissim} provides a trial version. The simulator is only available for Windows.
VISSIM is the slowest program from all the simulators tested, even when the scenario executed was a basic network.
Figure \ref{vissim} presents the VISSIM user interface.

\subsubsection{SUMO}
\label{subsec:sumoRev}

\textit{Simulation of Urban MObility} (SUMO) \cite{SUMO2012} is an open-source
microscopic traffic simulation suite available since 2001. It was developed and is maintained by the
\textit{Institut f{\"u}r Verkehrssystemtechnik} (Institute of Transportation Systems), located in Braunschweig, Germany. 
SUMO is highly portable and it was designed to handle large road networks. 

\begin{figure}
	\centering
	\includegraphics[scale=0.37]{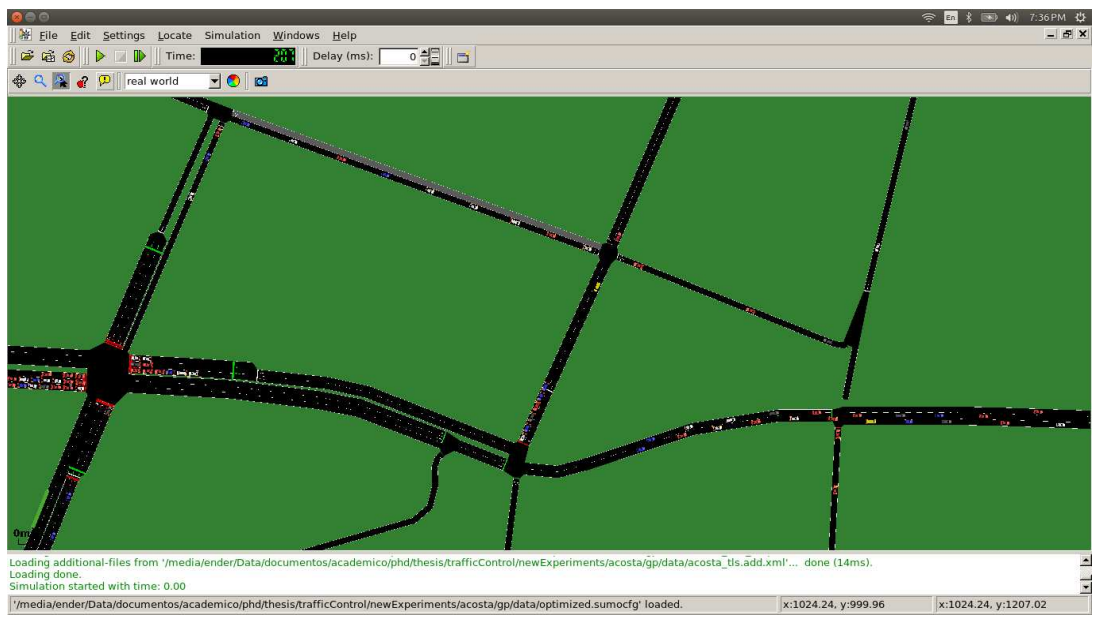}
	\caption[SUMO screen-shot]{SUMO version 0.25 screen-shot.}
	\label{sumo-gui}
\end{figure}

To model the vehicle dynamics and the drivers behaviour, 
SUMO uses an extension of the stochastic car-following model developed by Stefan Krau{\ss} \cite{krauss1998microscopic}.
To control the traffic signals, SUMO uses a time step of one second. Every second, SUMO determines 
the traffic signal settings for the next step.  SUMO has some built-in strategies for traffic signals, but
it also includes a traffic control interface called \textit{Traffic Control Interface} (TraCI).
TraCI allows the retrieval of values from simulated objects and online closed
loop feedback \cite{wegener2008traci} using a Python interface. 
With TraCI, we have almost unlimited freedom in programming adaptive traffic light controllers. Figure \ref{sumo-gui} presents the 
user interface of SUMO.
  
\subsection{Our traffic simulator: MiniTraSim}
\label{subsec:our_sim}

\begin{figure}
	\centering
	\includegraphics[scale=0.63]{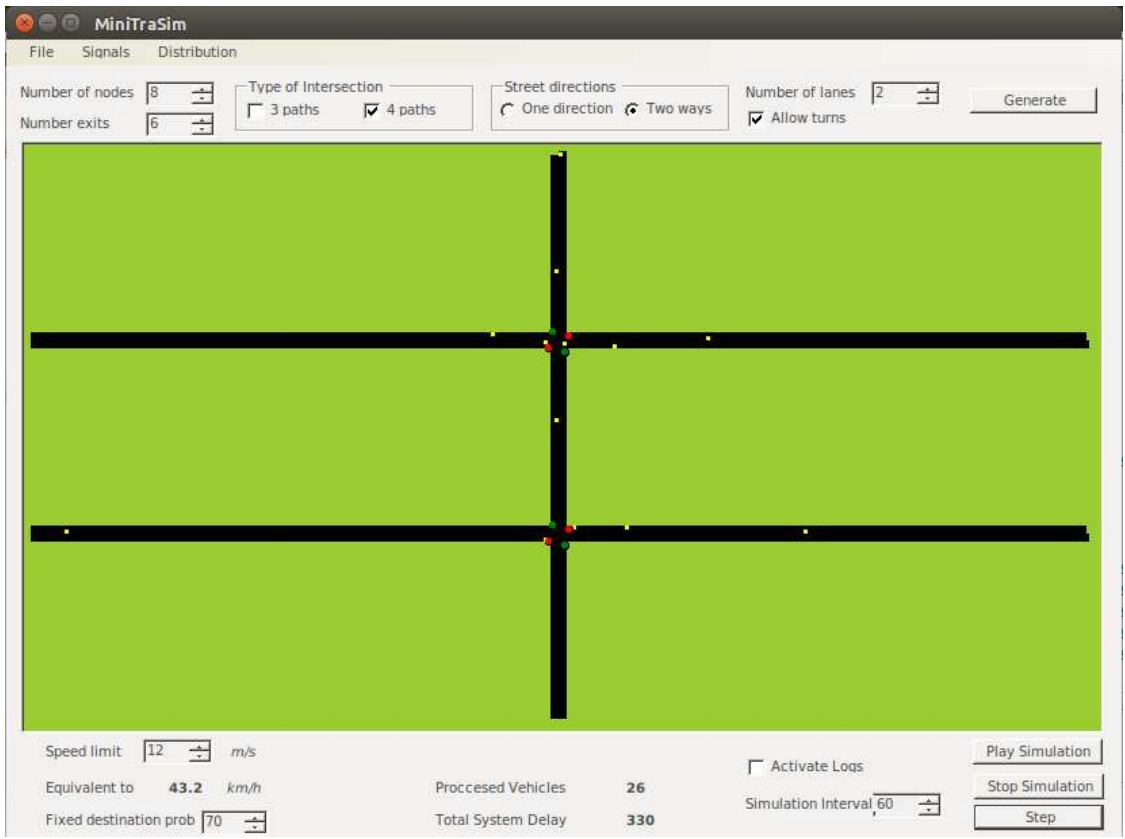}
	\caption[MiniTraSim screen-shot]{MiniTraSim graphic user interface screen-shot.}
	\label{mySim_gui}
\end{figure}

Although several commercial and open source simulators are available (Sections \ref{subsec:aimsunRev}, \ref{subsec:vissimRev} and \ref{subsec:sumoRev}), a 
microscopic model simulator was created from the ground as part of 
this research project. The decision was motivated by multiple reasons. The goals were to have full control of the environment and to simulate 
changing dynamic traffic conditions of complex traffic networks with several roads of multiple lanes using an approach with low computational 
cost. Our simulator is called \textit{Minimum Traffic Simulator} (MiniTraSim). Figure \ref{mySim_gui} presents the graphic interface of MiniTraSim.

For MiniTraSim, we decided to use the microscopic approach for traffic simulation because it better models the complex properties of the real 
phenomenon than the other two approaches (see Section \ref{subsec:microscopic}). The vehicular model of MiniTraSim is based on the TCA described 
in \cite{nagel1992cellular} and \cite{sanchez2010traffic}, but incorporates concepts borrowed from multi-agent systems and object oriented programming 
to allow the simulation of multiple roads with multiple lanes. A detailed description of MiniTraSim can be found in Appendix \ref{apdx:our_sim}.

\section{Discussion}
\label{sec:trafficDiscussion}

This chapter presented a brief introduction to traffic signal control, which is a large combinatorial problem with high 
complexity and uncertainty. Because of this complexity, finding the optimal signal configuration for networks composed of more than one 
intersection is not feasible using formal approaches. Therefore, solutions to complex real scenarios require the use of microscopic 
simulators and heuristic methods.  

Traffic signal control is an atypical problem from the computer science perspective. There are no standard benchmarks and 
test suites did not exist before 2016. Furthermore, a comparative study of the state of the art methods has not been carried out yet. 

Several factors complicate the objective comparison of different solutions to the traffic signal control problem.
Different studies consider diverse variables and parameters along various
simulators through different representations of the problem. Some scenarios consider external components such as accidents, road-blocks, public transportation,
bicycles, pedestrians or weather conditions as an approach to represent the uncertainty of the real phenomenon \cite{lu2016modelling}. Sometimes vehicle-to-vehicle,
vehicle-to-signal or signal-to-signal communications are considered.
Multiple commercial solutions; such as TRANSYT \cite{robertson1969transyt}, SCATS \cite{sims1980sydney}, UTOPIA \cite{mauro1990utopia} and BALANCE \cite{friedrich1995balance}; 
are available with drastic changes in performance from one version to other. The commercial solutions require expensive
licenses in order to be used or tested, and their technical reports include minimal information about the algorithms used.
 
Furthermore, traffic simulation is a computationally expensive process that is difficult to parallelize. To achieve parallelization, the road 
network should be partitioned into sub-networks and the simulation of these sub-networks should be executed on different computers \cite{igbe2003open, dai2010parallel}.
Most of the modern traffic simulators do not support parallelization by default and custom modifications are required to allow it \cite{ahmed2016partitioning}. 
Because of this restriction, an exhaustive amount of computational time is required to improve the signals of a specific traffic scenario regardless of the method used.
Finally, some implementations have strong dependencies with the simulator and/or the topology of the related traffic scenario and cannot be tested under 
different circumstances.

Despite the limitations here described, research groups in transportation, engineering and computer science around the world try to 
propose novel solutions for traffic signal control because of the economic cost, social and environmental impact traffic congestion 
has on today's society. A method to evolve actuated traffic controllers is presented in 
Chapter \ref{sec:probRepresentation}. The results of several experiments to test the performance of this method in the solution 
to traffic scenarios of different sizes are presented in Chapters \ref{chap:synthScenarios} and \ref{chap:sumo}.
 % Real-World Application

\chapter{Evolution of Actuated Traffic Controllers}
\label{chap:synthScenarios}

  \graphicspath{{figures/PNG/}{figures/PDF/}{figures/EPS/}{figures/}}
  \lhead{\emph{Chpt 5: Evolution of Actuated Traffic Controllers}}  % Set the left side page header to "Synthetic Scenarios"

Using the simulator described in Appendix \ref{apdx:our_sim} and the epigenetic mechanism described in Chapter \ref{chap:approach}, we propose a
method to test and evolve traffic controllers. The method uses Genetic Programming to \textit{automatically} evolve 
adaptive actuated controllers for traffic scenarios in the solution to the traffic signal control problem described in Chapter \ref{sec:signal}.   

A traditional approach to the problem using Evolutionary Algorithms is the optimization of signal schedules for traffic densities based
on historical data and fixed time intervals \cite{zhang2009multi, sanchez2010traffic,nie2010based}. However, given that traffic congestion
is a dynamic phenomenon, this approach has the downside of requiring constant updates of the schedules. Furthermore, it requires 
constant monitoring and design of contingency plans to react to atypical congestion events.

In our approach, the evolved controllers are not pre-calculated signal schedules but adaptive rules evolved for online modification
of the duration of signal phases in response to real time measurements of the local traffic conditions at intersections. 
This approach adapts to modifications in the traffic density and requires less monitoring and less human interaction 
because it dynamically adjusts the light behaviour depending on local traffic conditions of each optimized intersection of the traffic network.      

\section{Problem Representation}
\label{sec:probRepresentation}

With regard to Genetic Programming, traffic controllers to be evolved for a specific traffic network are represented as a 
forest of syntactic trees, where each tree represents an independent program for traffic signal control or \textit{controller}.
The mapping of controllers to intersections is selected depending on characteristics of the given traffic network. 
These mappings can be organized from general to specific: 
a single controller can be used to control all the intersections in the network; the intersections can be grouped by specific properties 
(e.g., number of entering lanes and exit lanes or number of lanes in each direction) and each group associated with a different controller; 
or an independent controller can be evolved for each intersection.

Intersections of simple monotonous networks, such as the networks used in \cite{zubillaga2014measuring,yuan2016optimal, kai2014real}, behave in a similar 
way. For this type of networks, it is reasonable to evolve a single controller for all the intersections in the network. However, networks simulating real 
world scenarios \cite{covell2015micro,bieker2015traffic,codeca2015luxembourg}, can contain complex intersections and include different traffic 
densities and priorities. For these cases, each intersection should be considered a different problem and mapped to an independent controller to be evolved.   

 \begin{table}
    \caption[Traffic variables on terminal set of Genetic Programming]{Traffic variables available to be included in terminal set of Genetic Programming.}
      \label{gpTermSetVariables} 
      \centering \small
      \begin{tabular}{|l|p{9cm}|} \hline
       \textbf{Variable} & \textbf{Description} \\ \hline
		{\tt verQueue} &  Sum of the number of vehicles stopped in the north-south direction and the number of vehicles stopped in the south-north direction of the current intersection  \\ \hline
		{\tt horQueue} &  Sum of the number of vehicles stopped in the west-east direction and the number of vehicles stopped in the east-west direction of the current intersection  \\ \hline
		{\tt 1stTopNeighbourQueue} &  Number of vehicles stopped in the north-south direction of the first intersection in the north direction of the current intersection  \\ \hline
		{\tt 1stBottomNeighbourQueue} &  Number of vehicles stopped in the south-north direction of the first intersection in the south direction of the current intersection  \\ \hline
		{\tt 1stLeftNeighbourQueue} &  Number of vehicles stopped in the west-east direction of the first intersection in the west direction of the current intersection  \\ \hline
		{\tt 1stRightNeighbourQueue} &  Number of vehicles stopped in the east-west direction of the first intersection in the east direction of the current intersection  \\ \hline
		{\tt 2ndTopNeighbourQueue} &  Number of vehicles stopped in the north-south direction of the second intersection in the north direction of the current intersection  \\ \hline
		{\tt 2ndBottomNeighbourQueue} &  Number of vehicles stopped in the south-north direction of the second intersection in the south direction of the current intersection  \\ \hline
		{\tt 2ndLeftNeighbourQueue} &  Number of vehicles stopped in the west-east direction of the second intersection in the west direction of the current intersection  \\ \hline
		{\tt 2ndRightNeighbourQueue} &  Number of vehicles stopped in the east-west direction of the second intersection in the east direction of the current intersection  \\ \hline
      \end{tabular}
 \end{table}

The function set contains mathematical operators (addition, subtraction, multiplication and protected division), 
logical operators (conjunction, disjunction and negation), comparison operators (equal to, greater than and less than), and a conditional operator
(ternary tree node with if, then and else branches). Different traffic variables local to each intersection are available to be included in the terminal set. 
The full list of variables is presented in Table \ref{gpTermSetVariables}. All the variables in the list are related to queue sizes of inbound roads at intersections.
As expressed by Little's theorem \cite{little2008little}, the queue size is directly proportional to the delay experienced by the vehicles in the network. Therefore, to 
minimize the vehicle delays one should seek to minimize the queue sizes \cite{wunderlich2007stable}.

For the experiments in this chapter, the queues associated with intersections only consider stopped vehicles on their inbound roads. These queues are not
limited to a specific area of the roads. Under highly congested conditions, queues can consider all the vehicles in a road as long as all of them 
are stopped.    

During the simulation, as part of the individual evaluation, the signal controller is executed at each intersection twice every light cycle with the current 
traffic parameters. The resulting integer number is added to the vertical green phase duration, subtracted from the vertical red phase duration, added to the 
horizontal red phase duration and subtracted from the horizontal green phase duration.

 \begin{table}
    \caption[Configuration parameters for GP]{Configuration parameters for the Genetic Programming model.}
      \label{gpParams} 
      \centering 
      \begin{tabular}{|c|c|} \hline
       \textbf{Configuration Parameters} & \textbf{Selected Values} \\ \hline
		Population size & 50 individuals  \\  \hline
		Number of generations & 200  \\  \hline
		Mutation probability rate per node & 5\%   \\ \hline
		Crossover probability & 80\%  \\ \hline
		Initial size limit & 5 levels  \\  \hline
		Maximum size limit & 7 levels  \\  \hline
		Selection operator & Tournament selection  \\  \hline
		Tournament size & 7 individuals \\ \hline
		Elitism & 1 individual \\
		\hline
       \end{tabular}
  \end{table}

Both scenarios in this chapter use the same parameters for Genetic Programming. These parameters are presented in Table \ref{gpParams}.
% I need to talk relate this selection to empirically well known parameters. This is missing. Do this meanwhile Banzhaf reads the thesis

\subsection{Longest queue actuated controller}
\label{subsect:mylongqueue}

An actuated controller was designed to increase mobility in the vertical direction or the horizontal direction when the size of the associated queue is greater than the 
size of the queue associated with the opposite direction for more than a pre-defined threshold. The threshold is set to 5 vehicles for the experiments in this chapter.  
This controller, called the longest queue actuated controller, is based on the longest queue first scheduling algorithm (See Chapter \ref{subsubsec:longqueue}). 

\begin{figure}
\centering
\includegraphics[scale=1.2]{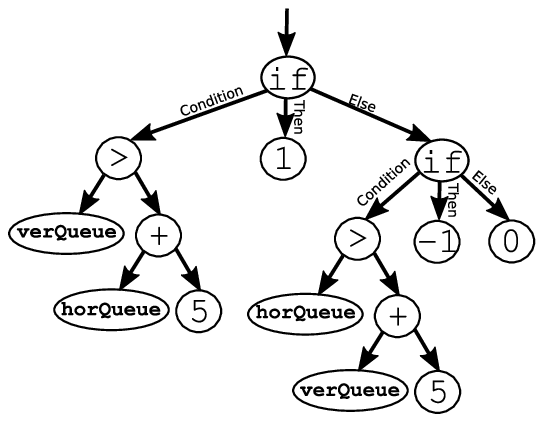}
\caption[Tree representing the longest queue actuated controller]{Tree representing the longest queue actuated controller. The variables {\tt verQueue} and {\tt horQueue}
return the number of stopped vehicles in the vertical direction and horizontal direction at the time the controller is evaluated for an intersection.}
\label{humanSolution}
\end{figure}

\begin{figure}
\centering
\includegraphics[scale=0.8]{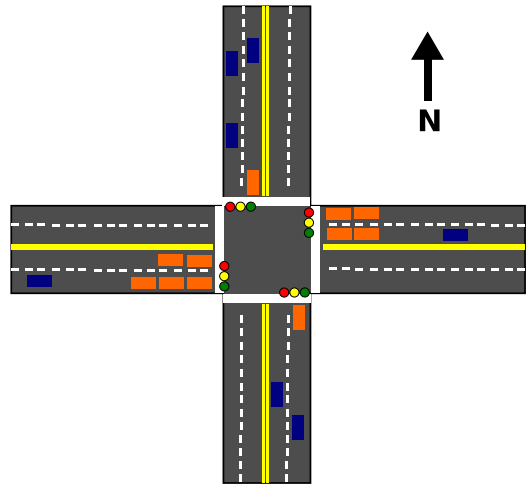}
\caption[Intersection timestamp example]{A single intersection with two lanes and traffic flow in both directions. Orange 
rectangles represent stopped vehicles considered in a queue of the intersection. Blue rectangles represent vehicles in motion and not considered in a queue.}
\label{singleIntersection}
\end{figure}

Figure \ref{humanSolution} displays a tree representation of the longest queue actuated controller.
Suppose the intersection presented in Figure \ref{singleIntersection} was evaluated using the longest queue actuated controller. 
With the traffic conditions presented in the figure, the value of {\tt verQueue} would be 2 and the value of {\tt horQueue} would be 9 because only the vehicles 
in orange colour are fully stopped and considered by the queues. 
Therefore, the output of the longest queue actuated controller for the traffic conditions presented in Figure \ref{singleIntersection} would be to increase by 
1 second the west-to-east and east-to-west green phase duration, decrease by 1 second the west-to-east and east-to-west stop phase duration, 
decrease by 1 second the north-to-south and south-to-north green phase duration, and increase by 1 second the north-to-south and south-to-north stop phase duration.

The particular moment in the signal cycle when the controller is executed clearly affects the controller behaviour. 
For this reason, every controller is executed twice in every light cycle. For the intersection presented in Figure \ref{singleIntersection}, the controller is
executed a first time when the west signal and east signal are switching from red to green and is executed a second time when the north signal and south 
signal are switching from red to green. The same execution approach was used for the all the evolved traffic controllers of the two scenarios presented 
in this chapter (see Sections \ref{chap:concept-proof} and \ref{chap:highway}).

\section{Diversity Operators}
\label{sec:diverOp}

Two different crossover operations are used by Genetic Programming methods in the scenarios presented in Sections \ref{chap:concept-proof} and \ref{chap:highway}: 
controller exchange and sub-tree exchange \cite[pp. 242]{banzhaf1998genetic}.
Controller exchange is triggered in $10\%$ of all the crossover operations. It exchanges one controller between two chromosomes. The operation 
swaps controllers in the same position on both chromosomes. Sub-tree exchange selects a random crossover point for a specific controller in 
both chromosomes and exchanges the two sub-trees selected only if both are of the same type; otherwise, a new crossover point is selected.  

Two different mutation operations are used by Genetic Programming methods in the scenarios presented in Sections \ref{chap:concept-proof} and \ref{chap:highway}: 
New tree mutation and point mutation. New tree mutation occurs with a probability of $0.1\%$. It selects
a random controller of the forest and replaces it with a newly generated controller. Point mutation replaces a single node with a node of the same type.

Strongly-Typed Genetic Programming \cite{montana1995strongly} requires the specification of a data type for every variable, constant, argument and 
returned value. This causes the initialization process and genetic operations to only construct syntactically correct trees.
In our solution, strong typing is enforced through the validation of the data types as part of the initialization process and 
the evaluation of the selected points before the application of the combination operators. 

\section{Epigenetic Mutation for Traffic Scenarios}
\label{sec:epimutation}

Trees representing traffic signal controllers use the activation markers defined in Chapter \ref{sec:CodeDeact}. Each \textit{conditional node} 
is associated with an activation rate. Activation rates are randomly initialized between $0\%$ and $100\%$ for the initial population. 
During the controller evaluation, if the activation rate is less than an activation threshold, with a default value of $50\%$, the conditional 
node is ignored and the \textit{else sub-tree} is executed, deactivating with that action the \textit{conditional sub-tree} and the \textit{then sub-tree}.

In the context of traffic optimization, the adaptive factor, previously defined with Equation \ref{eq:adaptFactor}, is redefined with Equation 
\ref{eq:traffAdaptFactor}, where $S_{c}(t)$ is the controller stability at the simulation step $t$, 
and $\overline{S}_{c}(t)$ is the controller average stability of a time interval 
\begin{equation} \label{eq:traffAdaptFactor}
\lambda_{c}(t) = \frac{|S_{c}(t)-\overline{S}_{c}(t)|}{\overline{S}_{c}(t)}.
\end{equation}
The controller average stability is calculated using Equation \ref{eq:meanTraffDiff}, where $T$ represents the duration of the 
interval, 5 light cycles by default,  and $S_{c}(i)$ is the stability of a controller $c$ at a specific simulation step $x$
\begin{equation} \label{eq:meanTraffDiff}
\overline{S}_{c}(t) = \frac{\sum\limits_{x=t-T}^t S_{c}(x)}{T}.
\end{equation}
The stability of a controller is measured with Equation \ref{eq:treeTraffDiff}, where $n$ is the total number of intersections assigned to a specific 
controller $c$, $t$ represents a specific time step of the simulation and $S_{c_i}(t)$ is the traffic stability of an intersection $i$ at the simulation
step $t$
\begin{equation} \label{eq:treeTraffDiff}
S_{c}(t) = \frac{\sum\limits_{i=1}^n S_{c_i}(t)}{n}.
\end{equation}
For each intersection evaluated by a traffic controller, the traffic stability, defined as the difference between the traffic congestion in 
vertical directions (from south to north and from north to south) and the traffic congestion in horizontal directions (from east to west and from west to east), 
is calculated using Equation \ref{eq:intersectTraffDiff}, where $i$ represents an intersection evaluated by the controller $c$
and $t$ represents the current time step 
\begin{equation} \label{eq:intersectTraffDiff}
S_{c_i}(t) = \mathrm{verQueue}_{i}(t) - \mathrm{horQueue}_{i}(t). 
\end{equation}
A normalization to the range $[0,1]$ is performed to the controller adaptive factor using Equation \ref{eq:traffNormAdaptFactor}, where  
$\lambda_c(t)$ represents the controller adaptive factor calculated with Equation \ref{eq:traffAdaptFactor}, $\lambda_{max}(t)$ represents the 
maximum possible value of $\lambda_c(t)$ for all the controllers through the entire experiment, and $\lambda_{min}(t)$ represents the minimum possible value 
of $\lambda_c(t)$ for all the controllers through the entire experiment. $\lambda_{max}(t)$ and $\lambda_{min}(t)$ are empirical values pre-calculated with an 
additional execution of the entire traffic scenario

\begin{equation} \label{eq:traffNormAdaptFactor}
\lambda'_{c}(t) = \begin{cases}
    0, & \text{if } \lambda_c(t)< \lambda_{min}(t)\\
    1, & \text{if } \lambda_c(t)> \lambda_{max}(t)\\
    \frac{\lambda_c(t)-\lambda_{min}(t)}{\lambda_{max}(t)-\lambda_{min}(t)},  & \text{otherwise}
\end{cases}.
\end{equation}

As discussed in Chapter \ref{sec:adptfactor}, the normalized adaptive factor for the controller $\lambda'_{c}(t)$ is used as the probability in an
epigenetic mutation of the activation rates associated with the traffic controller. During the simulation, an epigenetic mutation based 
on the normalized adaptive factor is performed on the activation rates of the signal controllers using equation \ref{eq:epimutation}, 
where $a_{c_{j}}(t)$ represents the activation rate of the $j$ conditional node of the controller $c$
at simulation time step $t$, $\lambda'_{c}(t)$ represents the adaptive factor (calculated with equation \ref{eq:intersectTraffDiff}), 
$R_{1}$ and $R_{2}$ represent two different random numbers from the standard uniform distribution on the interval $(0,1)$, 
$h$ represents the highest mutation modifier $0.1$ and $l$ represents the lowest mutation modifier $-0.1$ and $a_{c_{j}}(t+1)$ is constrained to the 
range $[0,1]$

\begin{equation} \label{eq:epimutation}
a{c_{j}}(t+1) = \begin{cases}
    a{c_{j}}(t) + (h - l) R_{2} + l, & \text{if } R_{1}> \lambda'_{c}(t)\\
    a{c_{j}}(t),  & \text{otherwise} 
\end{cases}.
\end{equation}

At the end of the simulation, the final activation rates are stored in the chromosome markers and transferred to the next generation. 
This process transfers the environmental information collected during the lifespan of the individual (simulation of the traffic scenario) to the 
next generation in the same way epigenetic information is transferred at cellular level. The conceptual idea behind the additional adaptive process 
is to keep the system behaviour stable over environmental perturbations, one of the roles of Epigenetics at cellular level in Nature, and to 
transfer the learned strategies to the following generation.

\section{Proof of Concept}
\label{chap:concept-proof}

An initial experiment was performed as proof of concept. The objective of the experiment is to establish if an epigenetic-based mechanism could 
be implemented for Genetic Programming to evolve traffic controllers using an online approach for a basic network affected by dynamic changes to the traffic 
conditions. This proof of concept was presented in \cite{ricalde2016genetic}.

\subsection{Traffic scenario}
\label{subsec:pocNetwork}

\begin{figure}
\centering
\includegraphics[scale=0.5]{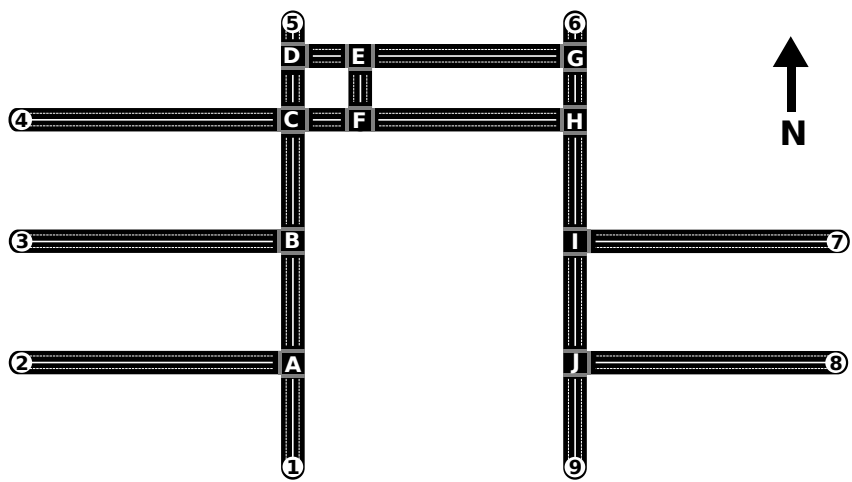}
\caption[Traffic network used for the proof of concept]{Traffic network used as proof of concept. Entry/exit nodes are marked with numbers. Intersections are marked with letters.
All the nodes are connected by two-lane bi-directional roads.}
\label{poc_trafficNetwork}
\end{figure}

An artificial network with 10 intersections, 9 entry/exit nodes and 31 traffic signals was designed for the proof of concept experiment. All the nodes are 
connected by two-lane bi-directional roads. The network is presented in Figure \ref{poc_trafficNetwork}. To reduce the complexity of the problem, instead of
evolving 10 independent controllers (each controller associated with one intersection), the intersections are grouped by number of intersecting roads and 
number of immediate connections to entry/exit nodes. Therefore, the problem is reduced to the design of four traffic controllers mapped to four 
sets of intersections: \{E, F, H\}, \{B, D, G, I\}, \{A, J\} and \{C\}.

Each simulation experiment models one hour of traffic. The traffic density of the different entry nodes in the network is modified every
21 seconds. This modification allows the simulator to represent real-world conditions where the network is only saturated at certain entry nodes for a specific
time of the day and the traffic peak slowly dissipates through a defined time interval.   

In order to replicate real-world similar conditions, the scenario simulates 16.5 hours of traffic. The traffic densities 
change during the simulated day and each entry node of the network follows a different distribution.

The first hour is considered a training step where all the entries follow a standard Poisson distribution going from zero density of 
traffic to the maximum saturation peak and declining again to zero traffic. The Poisson distribution was used because it correctly models
the arrival of vehicles, in one or multiple lanes \cite{mauro2013update} and because its flexibility allows the simulation of
changes in traffic densities (see Appendix \ref{sec:entry_prob} for more details).

\begin{figure}
\centering
\includegraphics[scale=0.7]{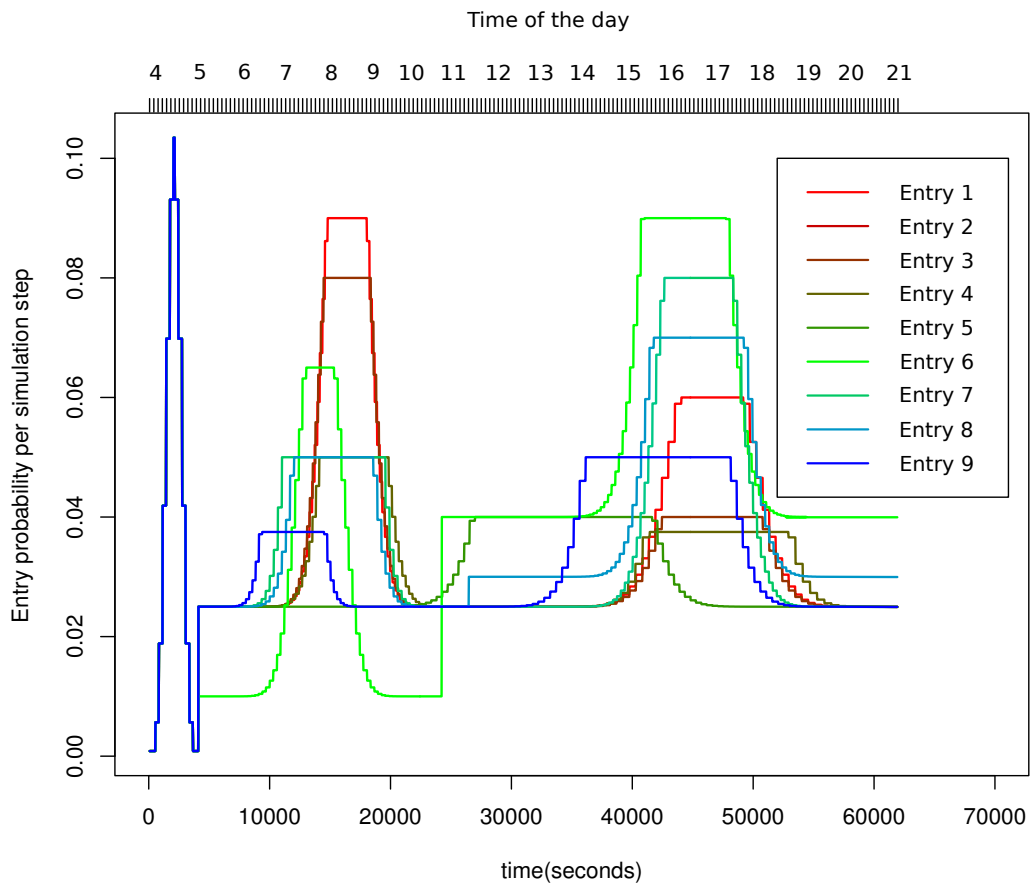}
\caption[Probability distributions for entry nodes of proof of concept scenario]{Probability distributions for traffic density on entry nodes of proof of concept scenario.}
\label{poc_inputDistrib3F}
\end{figure}

Two traffic waves are triggered through the remaining 15.5 hours of the scenario. The first one initiates from south-west entry nodes between 7 and 11 am. 
The second one from north-east entry nodes between 4 and 7 pm. 
Figure \ref{poc_inputDistrib3F} presents the probability distributions used to represent the defined behaviour for the network presented in Figure
\ref{poc_trafficNetwork}.

Although the complete scenario covers more than 16 hours of traffic, each simulation configuration only considers one hour of traffic.
A movable time window (see Chapter \ref{sec:timeWindow}) is used for the experiments. The window moves 5 minutes per configuration. With 
this approach, the full scenario is covered within 200 independent simulation configurations, and each GP generation is evaluated using a 
different time frame of the traffic scenario.

\subsection{Methods}
\label{subsec:pocMethods}

Five different algorithms were used to solve the traffic network presented in Figure \ref{poc_trafficNetwork}.
(1) default fixed time schedule, 
(2) longest queue actuated controller (see Section \ref{subsect:mylongqueue}), 
(3) an optimized time schedule evolved through a Genetic Algorithm,  
(4) actuated controllers evolved through Genetic Programming and (5) actuated controllers evolved through Genetic Programming with 
the epigenetic mechanism (see Section \ref{sec:epimutation}).

The baseline is a synchronized sequence of signal phases with fixed duration. For this method, all the signals are synchronized 
and the duration of each light phase is fixed to a specific value: 15 seconds for the green light, 5 seconds for yellow light, 10 seconds for red light and
10 seconds for the left turn. In the absence of a light controller, the traffic light system maintains the same traffic schedule for the 16 hours of 
the scenario.

Longest queue actuated controller is the second method tested. 
As described in Section \ref{subsect:mylongqueue}, this controller prioritizes the traffic direction with the highest demand at the
intersection level. As discussed in Chapter \ref{subsubsec:longqueue}, longest queue actuated strategies are commonly compared against the 
performance of other solutions for the traffic signal control problem because they are simple, stable, and easy to implement.
For each simulation, the signals start with the configuration used by the fixed time schedule. The controller is executed at 
each intersection twice for every light cycle. Therefore, the duration of the phases at each intersection can be modified depending on 
the local traffic conditions. Longest queue actuated controller is used for all the intersections. 

To compare our algorithm with other optimization methods commonly used in previous works, a signal schedule is evolved using a Genetic Algorithm 
similar to those presented in \cite{sanchez2010traffic}, \cite{matos2016traffic} and \cite{armas2016traffic}. The duration of 
each light phase of all the traffic signals in the system is stored as an integer chromosome. 
Standard crossover, mutation and tournament selection are used in an online optimization approach 
for 200 generations to approximate optimal pre-timed signal schedules for the movable time window along the 16.5 hours of traffic as 
described in Section \ref{subsec:pocNetwork}.  

Genetic Programming method and Genetic Programming method with the epigenetic mechanism evolve a set of actuated controllers
using an online approach with the representation described in Section \ref{sec:probRepresentation}. The second method includes the
the epigenetic mechanism described in Chapter \ref{chap:approach} and is affected by the epigenetic mutation 
defined in Section \ref{sec:epimutation}.

\subsection{Parameters}
\label{subsec:pocParams}

The parameters used for the Genetic Programming methods were presented in Table \ref{gpParams}. A similar configuration in terms of population
size, number of generations, selection method, mutation probability and crossover probability is used for the Genetic Algorithm. 

The terminal set contains integer constants in the range $[-10,10]$ and all the traffic variables listed in Table \ref{gpTermSetVariables}: 
{\tt verQueue}, {\tt horQueue}, {\tt 1stTopNeighbourQueue}, {\tt 1stBottomNeighbourQueue}, {\tt 1stLeftNeighbourQueue}, {\tt 1stRightNeighbourQueue}, \linebreak
{\tt 2ndTopNeighbourQueue}, {\tt 2ndBottomNeighbourQueue}, {\tt 2ndLeftNeighbourQueue} and {\tt 2ndRightNeighbourQueue}.

The function set contains mathematical operators (addition, subtraction and multiplication), 
logical operators (conjunction, disjunction and negation), comparison operators (equal to, greater than and less than), and a conditional operator
(ternary tree node with if, then and else branches).

Given the random components of the simulator (see Appendix \ref{sec:vehic_mod}), during the evolution process each individual is evaluated with 20 
independent runs of the associated simulation configuration. The final fitness of an individual is the average total system delay 
(see Appendix \ref{sec:performMea}) of 20 independent simulation runs with the same configuration.

\begin{figure}
\centering
\includegraphics[scale=0.7]{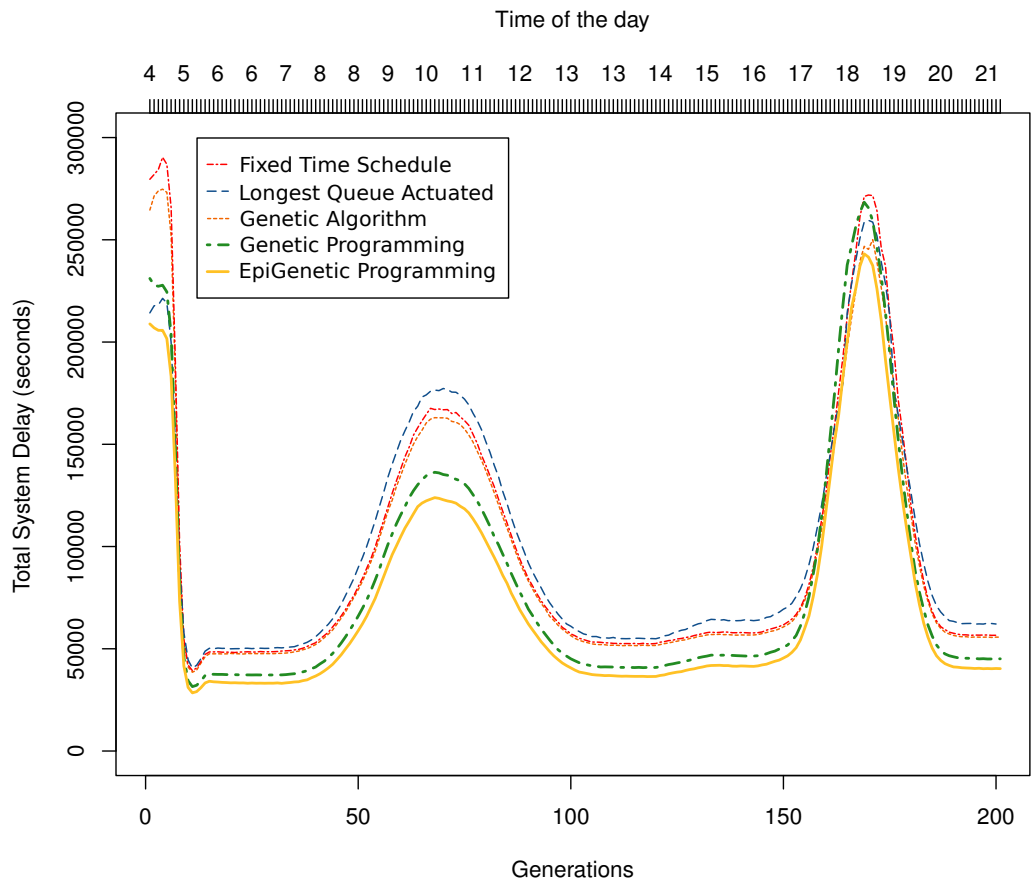}
\caption[Performance comparison for the proof of concept scenario]{Fitness curves of standard pre-timed, longest queue actuated control, GA pre-timed control, standard GP and GP with the epigenetic mechanism for the proof of concept scenario.}
\label{poc_fitness_plot}
\end{figure}

\subsection{Experiments}
\label{subsec:pocExperiments}

15 independent executions were performed for each method. Figure \ref{poc_fitness_plot} presents the comparison of the fitness obtained by the
five methods. For the fixed time schedule and the longest queue actuated controller, the average of the 15 independent executions is displayed for each generation.
For the Genetic Algorithm, standard GP and GP with the epigenetic mechanism the average fitness of the best individual per generation over the 15 independent executions  
is displayed for each generation.

Table \ref{poc_CumDiff} presents pairwise comparisons of the total system delay cumulative differences for combinations of the methods. 
The second column indicates the sum of the time differences the virtual vehicles spent during 16 hours of simulation for the different algorithms.   
The third column is that difference divided by the total delay of the fixed time schedule algorithm.

\subsection{Results}
\label{subsec:pocResults}

\begin{table}
    \renewcommand{\arraystretch}{1}
	\centering
	\caption[Total system delay cumulative differences for proof of concept scenario]{Total system delay cumulative differences of standard pre-timed, GA pre-timed control, standard GP and GP with the epigenetic mechanism in the proof of concept scenario.}
	\begin{tabular}{|p{4cm}|c|c|}
		\hline
		\textbf{Methods}
        & \multicolumn{1}{|p{3cm}|}{\textbf{Cumulative differences (seconds)}} 
        & \multicolumn{1}{|p{4cm}|}{\textbf{Relative difference with pre-timed schedule}}  \\ \hline
		Pre-timed vs GA  & 645294.6 & 3.33\% \\  \hline
		Pre-timed vs GP & 2881011 & 14.91\% \\  \hline
		Pre-timed vs EpiGP & \textbf{4580483} & \textbf{23.71\%} \\  \hline
		GP vs GA & 2235717 & 11.57\% \\  \hline
		GA vs EpiGP & 3935188 & 20.37\% \\  \hline
		GP vs EpiGP & 1699472 & 8.80\% \\ \hline
	\end{tabular}
	\label{poc_CumDiff}
\end{table}

The results displayed in Figure \ref{poc_fitness_plot} indicate that this proof of concept fulfilled the original objective.
Genetic Programming is able to evolve efficient traffic controllers for a traffic network affected by dynamic changes to the traffic conditions.  
Furthermore, adding the epigenetic mechanism to the Genetic Programming process allows GP to reduce by $10.06\%$ 
the total system delay for this specific scenario compared with standard GP and by $23.71\%$ compared to the default fixed 
time schedule.   

The learning curve of the evolutionary actuated control methods starts from the first generation being able to provide better 
solutions than the fixed time schedule for almost all the evaluation steps. This behaviour can be caused by the high variability of 
the traffic densities used for this experiment. Further experiments were performed with lower variability to analyze
the behaviour of the methods (see Section \ref{chap:highway}). 

GP with the epigenetic mechanism displays a lower fitness value than standard GP for almost all the evaluation points.
The difference between both methods is more drastic during the rush hours. A possible explanation is the adaptive
ability provided by the activation-deactivation of code by the epigenetic mechanism during simulation time.

It is worth noting that GP with the epigenetic mechanism outperformed the other four methods used in the experiments, providing an improvement of more than
$20$\% over the fixed time schedule. However, an evaluation of the methods using more scenarios with different levels of variability of the traffic 
conditions can lead to a better understanding of the differences presented in this proof of concept.

Although the objective of testing GP with an epigenetic mechanism in the solution to the traffic signal control problem for a network affected by dynamic changes of the traffic 
conditions was fulfilled, extensive experimentation is required to give statistical significance to the results presented in this section. To achieve that, 
independent experiments were performed for different scenarios, and statistical tools were used to perform analysis of the data produced (see Section \ref{subsec:hgwResults}). 
Scenarios of different sizes were evaluated to analyze the behaviour of the method under different circumstances (see Section \ref{chap:highway} and Chapter \ref{chap:sumo}). 
Finally, data from a real world network was acquired and tested in the scenario presented in Section \ref{sec:bologna}.

\section{Highway}
\label{chap:highway}

A new scenario was designed based on the results of the proof of concept. The network for the scenario contains a highway with two intersecting 
roads and two small secondary side roads. The goal for this experiment is to test the ability of the evolutionary methods to identify main roads from a traffic network,
generate controllers that benefit from the priorities identified, and efficiently use the information from connecting intersections to reduce the
total system delay of the scenario. 

\subsection{Traffic scenario}
\label{subsec:hgwNetwork}

\begin{figure}
\centering
\includegraphics[scale=0.5]{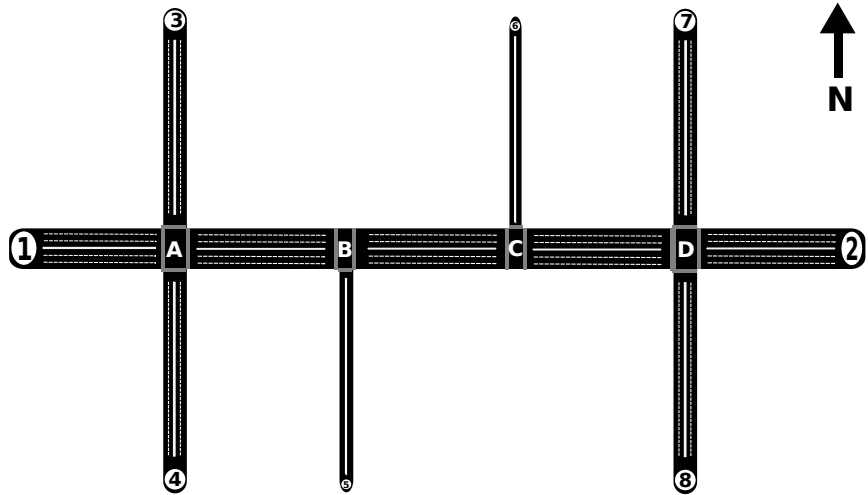}
\caption[Traffic network used for the highway scenario]{Traffic network used for the highway scenario. Entry/exit nodes are marked with numbers. Intersections are marked with letters}
\label{hwy_TrafficNetwork}
\end{figure}

An artificial network with 4 intersections, 8 entry/exit nodes and 14 traffic signals was designed to test Genetic 
Programming's ability to handle road priority. The main road is a bi-directional highway of 3 lanes in each direction. Two bi-directional roads, of 
two lanes in each direction, intersect the main road. Two additional bi-directional side roads of a single lane in each direction connect to the 
main road at different points and heading towards different exit nodes. The network is presented in Figure \ref{hwy_TrafficNetwork}. Each intersection 
of the network is associated with an independent controller to be evolved.
 
The traffic configuration for the scenario is similar to the configuration for the proof of concept scenario described in Section \ref{subsec:pocNetwork}.
The scenario simulates 17 hours of traffic. The entry nodes are configured to change the traffic densities every 22 seconds. 

\begin{figure}
\centering
\includegraphics[scale=0.8]{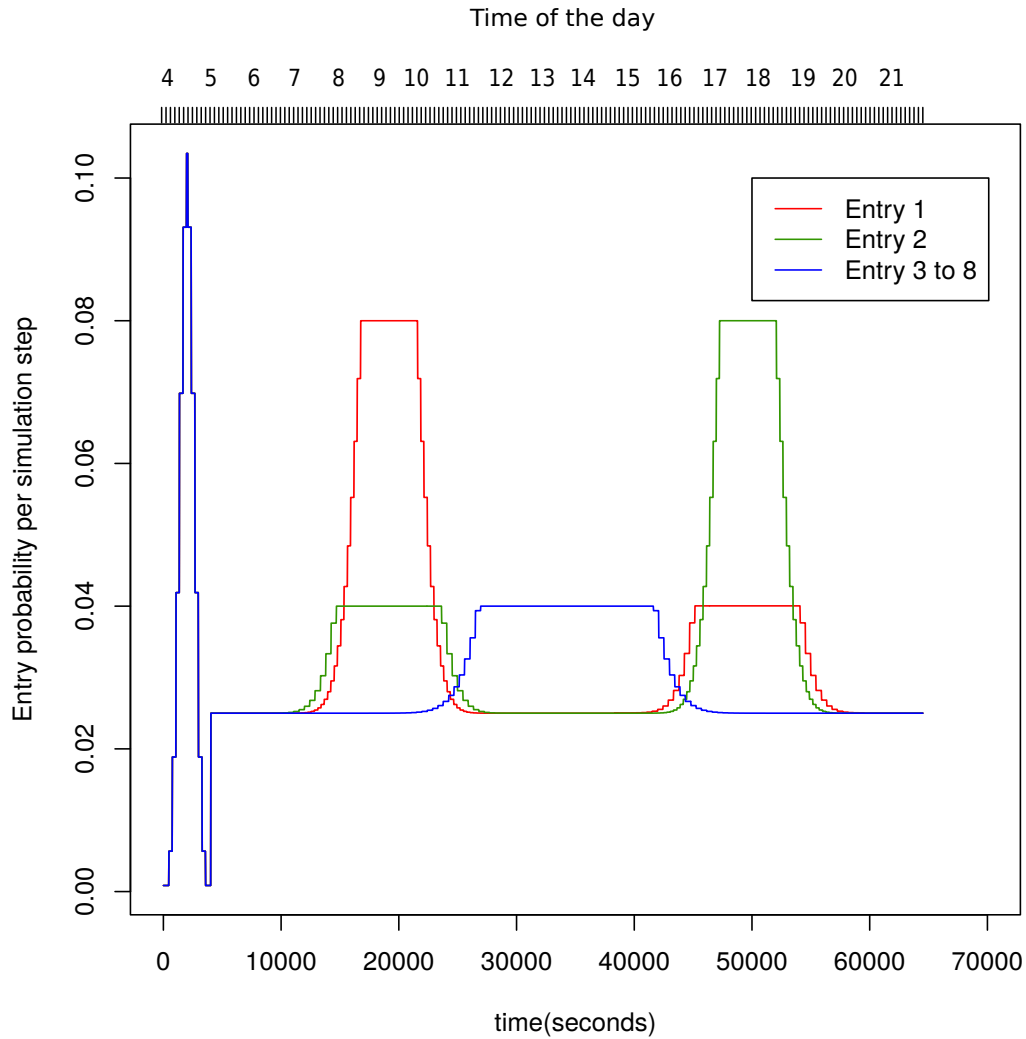}
\caption[Probability distributions for entry nodes of highway scenario]{Probability distributions for traffic density on entry nodes of highway scenario.}
\label{hwy_inputDistrib}
\end{figure}

The first hour is considered a training step. All the entry nodes follow a standard Poisson distribution going from zero density to the maximum 
saturation peak and declining again to zero traffic. Through the remaining 16 hours of the scenario, two traffic waves fill the main road. The first 
one initiates in the entry node 1 (west) between 7 and 11 am. The second wave goes from 4 to 7 pm and starts in the entry node 2 (east). 
Figure \ref{hwy_inputDistrib} presents the probability distributions used to represent the defined behaviour for the network presented in 
Figure \ref{hwy_TrafficNetwork}. 

The traffic densities for the entry nodes of the current scenario are simpler than those used in the proof of concept (see Section \ref{chap:concept-proof}). 
The traffic waves are symmetric for all the entry nodes. All the secondary roads follow the same entry distribution which do not interfere with 
the traffic waves in the main road. Even though these conditions make the highway scenario more artificial and separate it from real-world conditions, 
they may help to easily identify and study the adaptive effect of the evolutionary methods. 

Although the complete scenario covers 17 hours of traffic, each simulation configuration only considers one hour of the scenario.
A movable time window is used during the experiments in a similar way to the time window used in the proof of concept scenario (see Section \ref{subsec:pocNetwork}). 
The window moves 5.5 minutes per configuration. With this approach, the full scenario is covered within 200 independent simulation configurations, 
and each GP generation is evaluated using a different time frame of the traffic scenario.

\subsection{Parameters}
\label{subsec:hgwParams}

The five algorithms described in Section \ref{subsec:pocMethods} are executed independently to compare their performance in the highway scenario.

The scenario uses the same parameters for the Genetic Programming methods as the proof of concept. The parameters were presented in Table \ref{gpParams}. 
A similar configuration in terms of population size, number of generations, selection method, mutation probability and crossover probability is used for 
the Genetic Algorithm. 

The terminal set contains integer constants in the range $[-5,5]$ and some of the traffic variables listed in Table \ref{gpTermSetVariables}: 
{\tt verQueue}, {\tt horQueue},\linebreak {\tt 1stLeftNeighbourQueue}, {\tt 1stRightNeighbourQueue}, {\tt 2ndLeftNeighbourQueue} \newline and {\tt 2ndRightNeighbourQueue}. 
Only the east and west neighbour variables are considered because the network does not contain intersections connected in north direction or south direction. 
In case a controller tries to reach a nonexistent neighbouring position, the variable will always return zero. For example, the {\tt 1stLeftNeighbourQueue} 
variable of the intersection A will always return zero. 

The function set contains mathematical operators (addition and subtraction), 
logical operators (conjunction, disjunction and negation), comparison operators (equal to and greater than), and a conditional operator
(ternary tree node with if, then and else branches).

Given the random components of the simulator (see Appendix \ref{sec:vehic_mod}), during the evolution process each individual is evaluated with 
20 independent runs of the associated simulation configuration. The final fitness of an individual is the average total system delay  
(see Appendix \ref{sec:performMea}) of 20 runs of simulation runs with the same configuration.

\subsection{Experiments}
\label{subsec:hgwExperiments}

\begin{figure}
\centering
\includegraphics[scale=0.8]{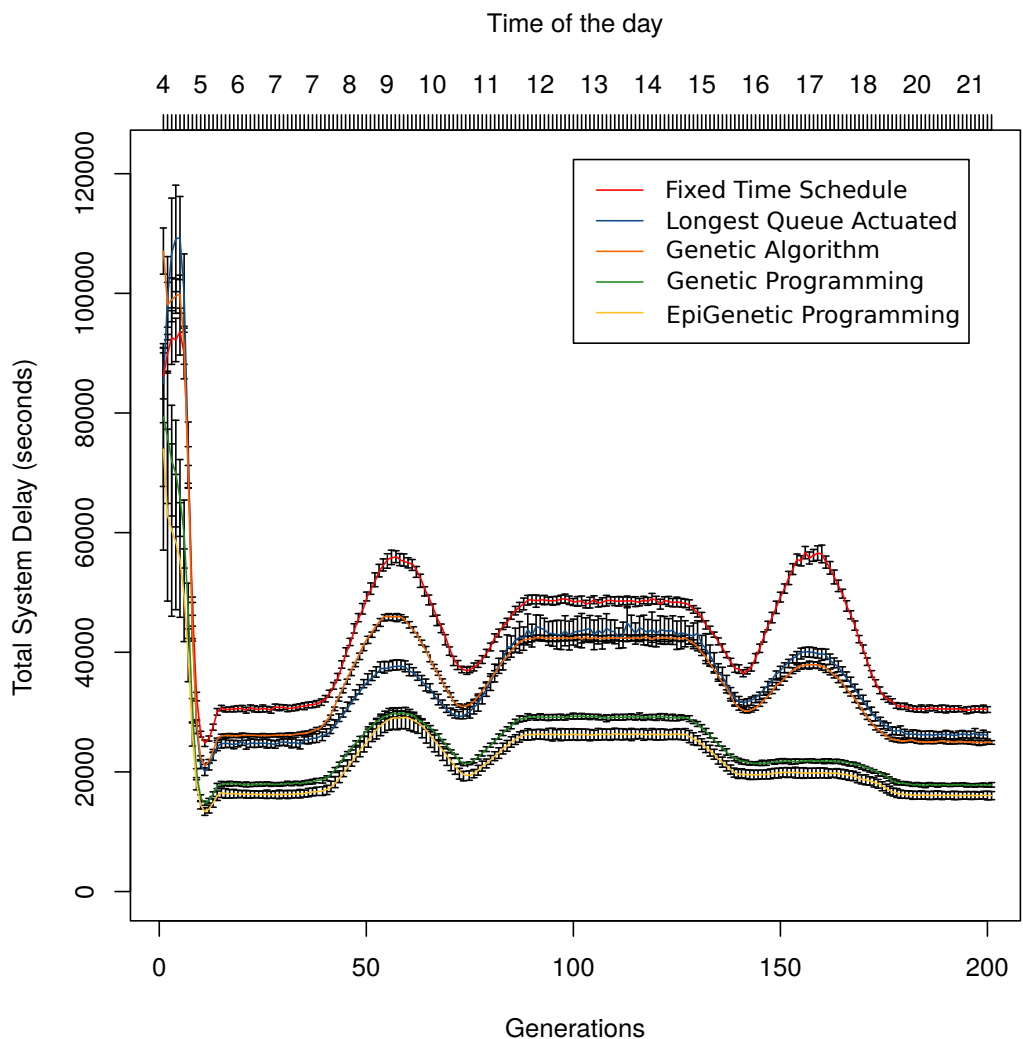}
\caption[Performance comparison for the highway scenario]{Fitness curves of standard pre-timed, longest queue actuated control, GA pre-timed control, standard GP and GP with the epigenetic mechanism for the highway 
scenario with standard deviation bars.}
\label{hwy_StdDeviationColor}
\end{figure}

In order to do a more extensive exploration of the sample space, 31 independent executions were performed for each method with the highway scenario. 
Figure \ref{hwy_StdDeviationColor} presents the comparison of the fitness obtained by the five methods. 
For the fixed time schedule and the longest queue actuated controller, the average and standard deviation of the 31 independent executions are displayed for each generation.
For the Genetic Algorithm, standard GP and GP with the epigenetic mechanism the average and standard deviation of the best individual of the 31 independent 
executions are displayed for each generation.

A single run of the highway scenario through one hour of traffic requires an average of 2.7 seconds on a hardware with a quad Intel Xenon x550 processor 
with speed of $2.67$ GHz and 8 GB of RAM running MiniTraSim (see Appendix \ref{apdx:our_sim}) in the version $3.2.8$ of Mono and the version $14.04$ of 
Ubuntu Linux operating system. Every candidate solution requires 20 simulation runs to be evaluated. Given the population size of 50 individuals and the 
total number of generations of 200, every independent run of GA, standard GP and GP with the epigenetic mechanism required an average of 150 hours only 
for the evaluation step. Meanwhile, every independent run of the fixed time schedule and longest queue actuated control required an average of 3 hours 
to be evaluated on the highway scenario. A cluster of 8 computers was used to execute in parallel the 31 independent experiments for the 5 methods tested 
on the highway scenario. Even with the use of the cluster, the total amount of computational time required for the execution of the experiments was 74 days.

The statistical analysis of the results obtained in the last generation does not necessarily represent the behaviour of the complete scenario because 
the environment changes dynamically. In order to consider these changes, eleven independent non-parametric statistical Friedman tests \cite{daniel1990applied}, spaced
by 20 generations, were performed over 31 independent executions of each of the five methods to test the statistical significance of the results presented.  

\begin{table}
\centering
\caption[Friedman test ranks for highway scenario]{Friedman test ranks every 20 generations for highway scenario.}
\begin{tabular}{|c|c|c|c|c|c|c|c|}\hline
    \textbf{Gen} & 
    ${\chi}^2$ & 
    \textbf{Fixed} &
    \textbf{LQ Actuated} & 
    \textbf{GA} &
    \textbf{GP} &
    \textbf{EpiGP} \\  \hline
1 & $72.877$ & 3.161 & 2.774 & 4.968 &  2.29 & 1.806 \\ \hline
20 & $119.355$ & 5 & 3.193 & 3.806 & 1.968 & 1.032 \\ \hline
40 & $123.226$ & 5 & 3 & 4 & 1.968 & 1.032 \\ \hline
60 & $121.213$ & 5 & 3 & 4 & 1.871 & 1.129 \\ \hline
80 & $117.806$ & 5 & 3.323 & 3.677 & 1.968 & 1.032  \\ \hline
100 & $118.581$ & 5 & 3.677 & 3.323 & 2 & 1 \\ \hline
120 & $117.187$ & 5 & 3.581 & 3.419 & 1.967 & 1.032 \\ \hline
140 & $121.729$ & 5 & 3.935 & 3.064 & 1.968 & 1.032 \\ \hline
160 & $122.451$ & 5 & 3.968 & 3.032 & 1.968 & 1.032 \\ \hline
180 & $119.871$ & 5 & 3.839 & 3.161 & 1.968 & 1.032 \\ \hline
200 & $121.058$ & 5 & 3.968 & 3.032 & 1.903 & 1.097 \\ \hline
\end{tabular}
\label{highay_friedman}
\end{table}

The eleven Friedman tests resulted in $p$-values $< 0.00001$. Therefore, we can conclude that there were statistically significant differences in the total system 
delay depending on the method used in solution to the highway scenario. Average ranks and ${\chi}^2$ values obtained by each method in the Friedman tests are presented in 
Table \ref{highay_friedman}. They indicate that Genetic Programming with the epigenetic mechanism has better performance than the other four methods tested.

Post hoc analysis of pairwise comparisons, through Dunn's tests \cite{dunn1964multiple}, were conducted for the eleven generations selected. 
The methods that did not display statistically significant differences
on the Dunn's tests by generation tested are presented in Table \ref{highway_posthoc}. There were no significant differences for several methods in
the test performed for the results obtained by the first generation. However, the tests performed on the results obtained for generations 20, 40, 60, 100, 
140, 160, 180 and 200 indicate statistically significant differences between all the methods compared.

\subsection{Results}
\label{subsec:hgwResults}

Figure \ref{hwy_StdDeviationColor} display more perceptible differences for the highway scenario than those presented for the proof of concept 
(see Section \ref{chap:concept-proof}). The influence of the entry traffic densities is more visible because simple and symmetric distributions were 
used for the traffic waves. The fixed time schedule and longest queue actuated controller clearly display no learning process. 
Both of the fitness lines are almost symmetric and follow a pattern similar to the entry probability distributions presented in Figure \ref{hwy_inputDistrib}.

\begin{table}
\centering 
\caption[Post hoc comparisons for highway scenario]{Post hoc Dunn's test pairwise comparisons every 20 generations resulting in non-statistically significant differences for highway scenario.}
\footnotesize
\begin{tabular}{|c|c|c|c|c|c|}\hline
    \textbf{Gen} & 
    \textbf{Fixed} &
    \textbf{LQ Actuated} & 
    \textbf{GA} &
    \textbf{GP} &
    \textbf{EpiGP} \\ \hline
1 & LQ Actuated & Fixed, GP & $-$ & LQ Actuated, EpiGP & GP  \\ \hline
20 & $-$ & $-$ & $-$ & $-$ & $-$ \\ \hline
40 & $-$ & $-$ & $-$ & $-$ & $-$ \\ \hline
60 & $-$ & $-$ & $-$ & $-$ & $-$ \\ \hline
80 & $-$ & $-$ & $-$ & $-$ & $-$  \\ \hline
100 & $-$ & $-$ & $-$ & $-$ & $-$ \\ \hline
120 & $-$ & GA & LQ Actuated & $-$ & $-$ \\ \hline
140 & $-$ & $-$ & $-$ & $-$ & $-$ \\ \hline
160 & $-$ & $-$ & $-$ & $-$ & $-$ \\ \hline
180 & $-$ & $-$ & $-$ & $-$ & $-$ \\ \hline
200 & $-$ & $-$ & $-$ & $-$ & $-$  \\ \hline
\end{tabular}
\label{highway_posthoc}
\end{table}

The Friedman tests presented in Table \ref{highay_friedman} and post hoc Dunn's tests presented in Table \ref{highway_posthoc} prove that the differences 
presented by the five methods compared are statistically significant. In ten of the eleven statistical analysis performed, the solutions generated by the 
Genetic Programming with the epigenetic mechanism were significantly better than the solutions generated by the other four methods tested.

It is worth noting that the longest queue actuated controller performed better in this scenario than in the proof of concept. A possible
cause is the difference in the number of lanes between the main road and the secondary roads. The accumulated queue sizes on the
main road are potentially higher than the queue sizes on the secondary roads. Because of this difference, the longest queue actuated 
controller is unintentionally assigning a higher priority to the main road. This provides an advantage to the longest queue actuated controller
 for this specific scenario. 

Because of the simple entry distributions, it is more clear that the three evolutionary methods display improvements during the second traffic wave. 
However, the improvements are more significant for the Genetic Programming methods. Both algorithms appear to use previous knowledge acquired 
during the first traffic wave to reduce the effect the second traffic wave has on the network.

GP with the epigenetic mechanism displays a lower fitness average for the best individual on 31 independent executions for the entire 200 generations over
the standard representation of Genetic Programming (relative differences between 3.8\% and 12.5\%). Although the difference between them is statistically significant, 
the performance of both methods remains close for the first 70 generations. 
However, after the end of the first traffic wave, the relative difference of the average on the fitness of the best individual between GP with the epigenetic mechanism 
and standard GP becomes wider and remains constant until the end of the scenario as can be 
seen in Figure \ref{hwy_StdDeviationColor}. A possible explanation for this behaviour is that, for some of the executions, GP requires some generations to create 
appropriate building blocks for the inactive sections of the chromosome controlled by the epigenetic mechanism.  

\section{Discussion}
\label{subsec:synthDiscussion}

The GP modification described in this chapter is a novel epigenetic approach specifically designed to work in 
the optimization of actuated controllers for traffic signals. Two artificial scenarios with different properties were analyzed and the results displayed 
an average improvement of 9.79\% compared to standard GP and an average improvement of 28.72\% compared to other methods traditionally used in solution 
to the traffic signal control problem. However, more scenarios of different sizes were evaluated to analyze the behaviour of 
the method under different circumstances (see Chapter \ref{chap:sumo}).

When the proof of concept was presented \cite{ricalde2016genetic}, one of the observations received from the community was that the simulator
used to execute the traffic scenario (see Appendix \ref{apdx:our_sim}) had not been used before. MiniTraSim should be compared 
to simulators widely used by the community (see Sections \ref{sec:simDiff} and \ref{sec:singleIntersec}). A similar performance over different 
simulators would provide more certainty to the experiments presented in this chapter.

Both traffic scenarios presented in this chapter have similar properties to real-world traffic networks. For example, the roads have multiple lanes; 
the speed limit used is similar to the urban speed limit of several countries including Brazil, Canada, Japan and UK \cite{wikimedia2018speedLimit}; and 
Poisson distributions are used to simulate two daily traffic waves in opposite directions, which is a typical behaviour of urban traffic 
networks \cite{geroliminis2011properties}.

However, the topology of both networks is rigid, in the sense that roads can only be oriented in four directions, and does not resemble most 
urban traffic networks in the real world. An additional scenario built using traffic data and signal data 
from a real world network was tested to provide more certainty to the improvements presented by the controllers evolved by GP with the epigenetic 
mechanism (see Chapter \ref{sec:bologna}).
 % Synthetic Scenarios

\chapter{Adaptable Controllers for Standard Simulator}
\label{chap:sumo}

  \graphicspath{{figures/PNG/}{figures/PDF/}{figures/EPS/}{figures/}}
  \lhead{\emph{Chpt 6: Adaptable Controllers for Standard Simulator}} 

To provide more certainty to the results presented in Chapter \ref{chap:synthScenarios}, we decided to 
replace MiniTraSim with a simulator widely used by the community. We selected SUMO from all the traffic simulation packages
presented in Chapter \ref{subsec:sim_compare} because it is open-source, free to use and multi-platform. Additional advantages 
over other simulators are: SUMO was designed to handle large road networks, allows online manipulation through TraCI with
standard Python and the recent release of traffic scenarios of real cities for SUMO suite to be used by the research 
community \cite{codeca2015luxembourg, bieker2015traffic}.

This chapter presents the differences between SUMO and MiniTraSim, the modifications
included in our framework to replace one for other and the evolution of traffic signal controllers for two additional 
traffic scenarios.  

\section{Simulator Differences}
\label{sec:simDiff}

SUMO (see Chapter \ref{subsec:sumoRev}) and MiniTraSim (see Appendix \ref{apdx:our_sim}) have several differences.
SUMO is a realistic general microscopic traffic simulator developed by a community for more than 15 years. MiniTraSim is a traffic
simulator designed and developed as part of this research project with the specific goal of using minimal resources to represent 
microscopic traffic simulation.  

SUMO uses a stochastic car-following model for vehicle and driver behaviour. Car-following models
are time-continuous models usually defined by differential equations describing the complete dynamics of the position and speed of each vehicle simulated.
In other words, SUMO uses real variables to store the position and speed of each vehicle. On the other hand, MiniTraSim uses a Traffic Cellular Automaton 
to represent the vehicle behaviour (see Appendix \ref{sec:vehic_mod}). \textit{Traffic Cellular Automata} (TCAs) are dynamic systems, discrete in time and space 
(see Chapter \ref{subsec:microscopic}). TCAs are less accurate than car-following models, but they are more efficient in terms of computational time. 

SUMO has a set of tools to handle traffic network generation and edition: NETGENERATE, NETEDIT and NETCONVERT. These tools can be used to create new 
traffic scenarios, to import traffic scenarios from other traffic simulators and to create traffic scenarios from traffic networks stored in 
OpenStreetMaps format \cite{OpenStreetMap2017}. MiniTraSim includes a basic tool to generate and edit traffic networks (see Appendix \ref{sec:raod_rep}).

The vehicle descriptions and routes need to be defined over independent files with specific XML formats before a traffic scenario is executed in SUMO. 
These files can be manually generated based on the behaviour observed in the real world. Additionally, SUMO provides a tool, called DUAROUTER, to 
generate routes with shortest paths from a file containing traffic demands of the different nodes in the network. In other words, SUMO does not 
calculate the routes during simulation time. On the other hand, MiniTraSim dynamically calculates the vehicle entries (see Appendix \ref{sec:entry_prob}) 
and vehicle routes (see Appendix \ref{sec:routing}) during simulation time.

\begin{figure}
\centering
\includegraphics[scale=0.5]{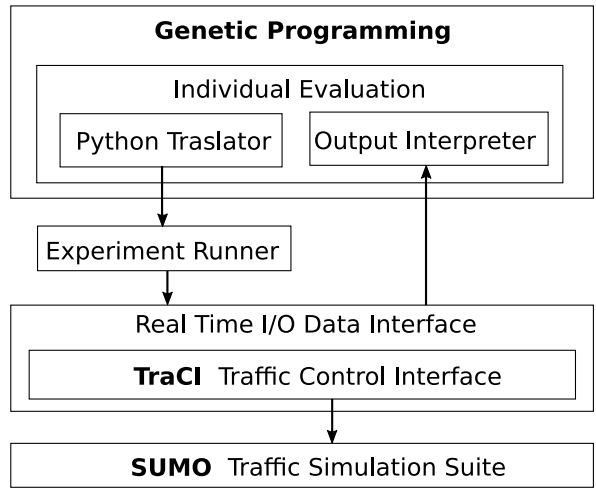}
\caption[Architecture diagram of framework for integration of SUMO and GP]{Architecture diagram of framework for integration of SUMO and Evolutionary 
Algorithm for optimization of actuated traffic signal controllers.}
\label{GPSumoInteraction}
\end{figure}

MiniTraSim was developed to incorporate the actuated control mechanism defined in Chapter \ref{chap:synthScenarios}. Every intersection is associated with a
traffic controller and every traffic signal has a queue counter for the number of vehicles stopped in the road associated with it. Intersections are allowed to 
have access to the queue counters of neighbouring traffic signals in order to use all the traffic variables defined in Table \ref{gpTermSetVariables}. 
On the other hand, SUMO allows the control of the traffic signals through TraCI. 

However, several modifications were required to implement the actuated control mechanism  defined in Chapter \ref{chap:synthScenarios} within SUMO. 
An architecture diagram of the framework implemented for this research project is displayed in Figure \ref{GPSumoInteraction}. 
SUMO does not have detectors to allow the identification of individual vehicles in full stop condition for an undefined section of a road. 
Instead, entry-exit detectors \cite{drebinger1983method} were used to identify the density of stopped vehicles on 
inbound roads at sections located before intersections. 
Appendix \ref{sec:sumo_detect} describes in detail the procedure required to add detectors to an existing SUMO traffic network. 
   
In order to have access and control over traffic signals, a communication layer was included in the framework to allow the interaction between GP and TraCI. 
The layer is called \textit{Experiment runner}. It controls the interaction between GP and SUMO through TraCI, including the execution of controllers 
and the epigenetic mutations. Appendix \ref{sec:sumo_runner} describes in detail the experiment runner layer. 

\begin{figure}
\centering
\includegraphics[scale=1.3]{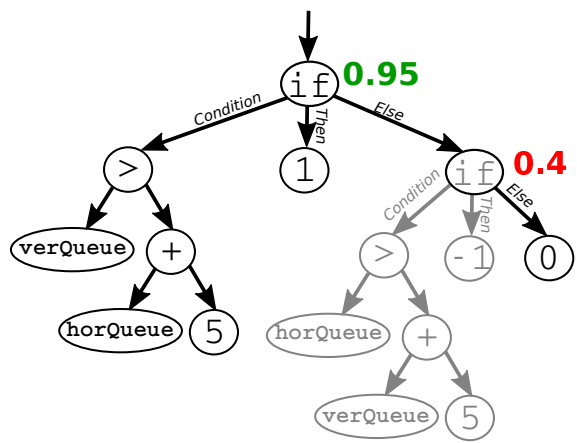}
\caption[Tree representing a traffic controller with epigenetic markers]{Tree representing a traffic controller with epigenetic markers. Activation rates in red 
are under the activation threshold. Activation rates in green are above the activation threshold. The nodes of the tree in grey are inactive because of the values of the associated
activation rates.}
\label{humanSolutionEpigen}
\end{figure}

The experiment runner requires the traffic controller to be in Python format in order to be executed. We implemented a basic translator to transform 
the controllers stored in dynamic memory into Python programs. For example, the Python class {\tt Controller0} is the output generated by the Python 
translator for the controller in Figure \ref{humanSolutionEpigen}, including the epigenetic vector. In this case, the second condition is ignored 
because the activation rate is less than the activation threshold. In other words, {\tt trafficRule} will return 0 even if {\tt hQueue > vQueue + 5} 
is true.

\begin{minipage}{\textwidth}
{\footnotesize \singlespacing
\begin{verbatim}
class Controller0(controller.Controller):
    def trafficRule(vQueue,hQueue):
        if (self.epiVect[0] > 0.5 and 
            vQueue > hQueue + 5) :
                return 1
        else:
            if (self.epiVect[1] > 0.5 and 
                hQueue > vQueue + 5):
                    return -1
        return 0
        
    def __init__(self):
        self.epiVect = [0.95, 0.4]    
\end{verbatim}
}\medskip
\end{minipage}

During the GP evaluation phase, the Python translator generates Python programs corresponding to the controllers to be evaluated. 
After these programs are generated, a synchronous call to the experiment runner is executed. The experiment runner handles the interaction with
SUMO through TraCI. After the experiment finishes, an additional process reads the SUMO standard output file to collect the information required to calculate
the performance measures of the scenario. 
   
\section{Single Intersection}
\label{sec:singleIntersec}

A simple scenario was designed to perform a basic test to compare the behaviour of MiniTraSim and SUMO. The scenario contains a single intersection and 
one hour of uniformly distributed vehicle entries. The objective of this experiment is not to simulate a network with real-world conditions, but to analyze the 
behaviour of both simulators in the evolution of actuated controllers for a stable basic traffic scenario.
 
\subsection{Traffic scenario}
\label{subsec:singleScenario}

\begin{figure}
\centering
\includegraphics[scale=0.5]{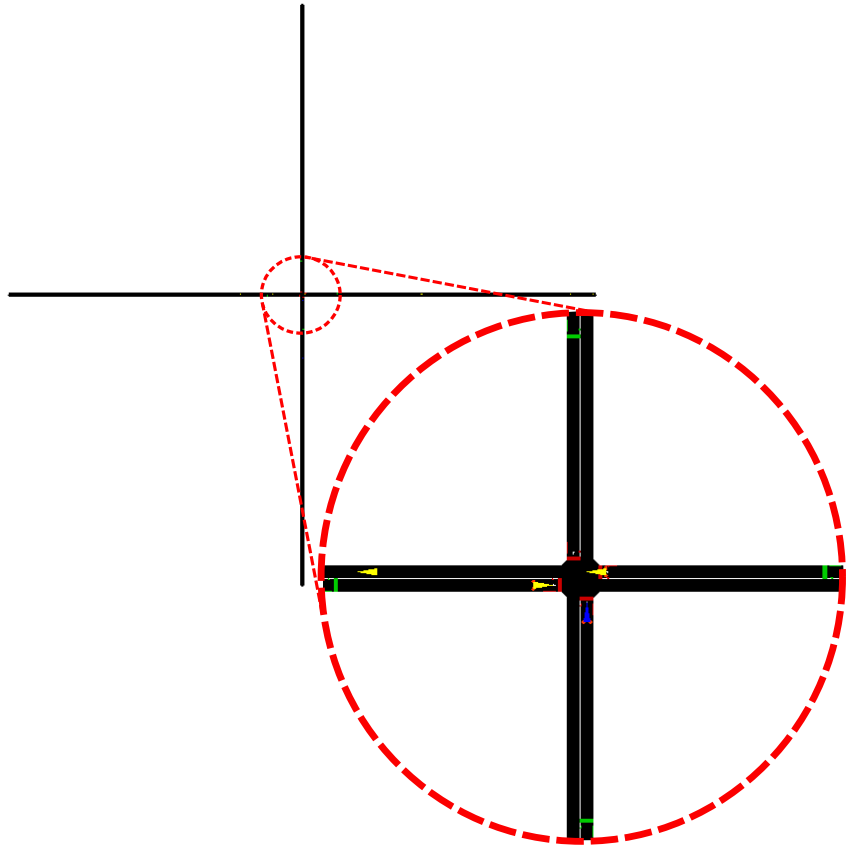}
\caption[Traffic network for single intersection scenario]{Traffic network for single intersection scenario.}
\label{singleIntersectionScenario}
\end{figure}

The scenario is a modification of the basic tutorial for the use of TraCI included in SUMO release \cite{dlr2017traciTutorial}.
The scenario contains one intersection of two roads with a single lane
in each direction. Entry-exit detectors were placed on the roads before reaching the traffic signals, as described in Appendix \ref{sec:sumo_detect}. 
The network with the additional detectors is presented in Figure \ref{singleIntersectionScenario}.  

The initial configuration of the traffic signals is presented in Table \ref{single_lightLapses}. For SUMO,
{\tt minDuration} of the green phases is defined as zero and {\tt maxDuration} is undefined. 
For MiniTranSim, a standard uniform distribution is used for entry demand instead of the Poisson distribution used for the experiments of 
Chapter \ref{chap:synthScenarios} and the probability $p$ for the randomization step of the vehicular model (see
Appendix \ref{sec:vehic_mod}) is set to $0$ to represent free flow conditions.  

Every second, the probability of arrival of a vehicle from the left entry and the right entry is $\frac{1}{10}$. Meanwhile, the probability of arrival of a
vehicle from the top entry and the bottom entry is $\frac{1}{30}$. Vehicles are not allowed to turn left or right and all vehicles have a fixed destination. 
In other words, vehicles starting from the top entry head towards the bottom exit and vice versa, while vehicles starting in the left entry head towards the 
right exit and vice versa.

\begin{table}
    \caption[Traffic signal schedule for single intersection scenario]{Traffic signal schedule for single intersection scenario.}
      \label{single_lightLapses} 
      \centering 
      \begin{tabular}{|c|c|} \hline
       \textbf{Light} & \textbf{Duration} \\ \hline
		Red &  33 seconds  \\ \hline
		Green &  27 seconds  \\ \hline
		Yellow &  6 seconds  \\ \hline
      \end{tabular}
\end{table}
 
The scenario covers one hour, 3,600 seconds, of simulation. Both simulators use the same simulation configuration and the same random seed every time the 
scenario is executed, thus ensuring that each execution is identical. This approach allows the evolution of controllers in a closed and consistent environment, 
from a different perspective to the approach used for experiments presented in Chapter \ref{chap:synthScenarios}. 
% Maybe describe a bit the approach used in the previous chapter?
  
\subsection{Parameters}
\label{subsec:singleParams}

Four of the algorithms described in Chapter \ref{subsec:pocMethods} are run independently in the single intersection scenario using both simulators. 
The methods compared are: (1) default fixed time schedule, (2) longest queue actuated controller (see Chapter \ref{subsect:mylongqueue}), 
(3) actuated controllers evolved through Genetic Programming, and (4) actuated controllers evolved through Genetic Programming with the epigenetic 
mechanism (see Chapter \ref{sec:epimutation}).

The scenario uses similar parameters for the Genetic Programming methods to those used in the experiments presented in Chapter \ref{chap:synthScenarios}. 
The parameters were presented in Table \ref{gpParams}. Sub-tree mutation \cite{banzhaf1998genetic} is used instead of point mutation and the mutation rate 
is modified to $20$\% per tree. 

The terminal set contains integer constants in the range $[-5,5]$ and two of the traffic variables listed in Table \ref{gpTermSetVariables}: 
{\tt verQueue} and {\tt horQueue}. The function set contains mathematical operators (addition and subtraction), 
logical operators (conjunction, disjunction and negation), comparison operators (equal to and greater than), and a conditional operator
(ternary tree node with if, then and else branches).

Given that the scenario does not change over time and random effects were removed through the use of the same random seed, each individual is evaluated with
a single simulation execution. It is important to remark that the properties of the scenario do not change over time. For the evolutionary methods, exactly the 
same configuration is evaluated for each generation. This approach is different from the approach used for the experiments presented in Chapter \ref{chap:synthScenarios}.
For the fixed time schedule and the longest queue actuated controller, a single simulation run is executed in the experiment. 

The fitness function is the total system delay over an hour of simulated traffic (see Appendix \ref{sec:performMea}). The calculation of averages is 
not required because a single simulation run is executed for each data point. 

\subsection{Experiments}
\label{subsec:singleExperiments}

\begin{figure}[!htp]
\centering
\includegraphics[scale=0.8]{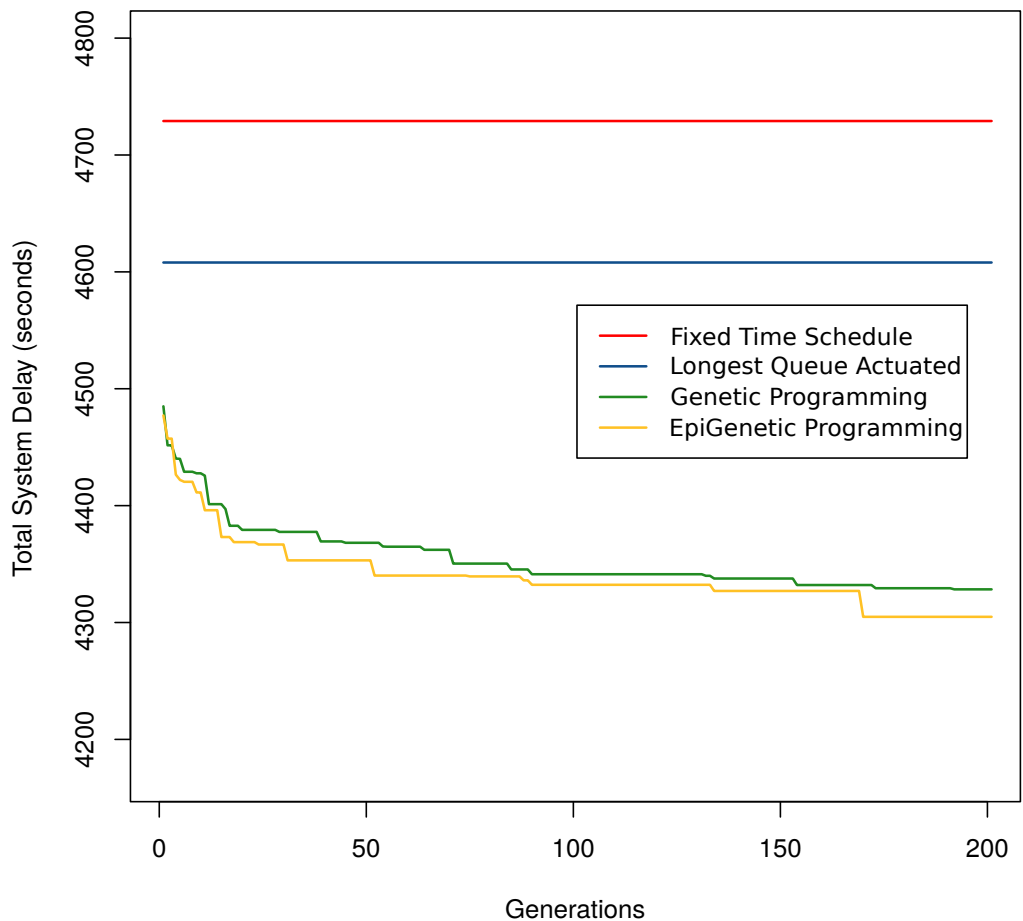}
\caption[Performance comparison for the single intersection scenario through MiniTraSim]{Fitness curves of standard pre-timed, longest queue actuated control, standard GP and GP with the epigenetic mechanism for the single intersection scenario through MiniTraSim. The $y$ axis representing Total System Delay does not start at 0 for this figure.}
\label{mySim_mySimVsSumo}
\end{figure}

\begin{figure}
\centering
\includegraphics[scale=0.8]{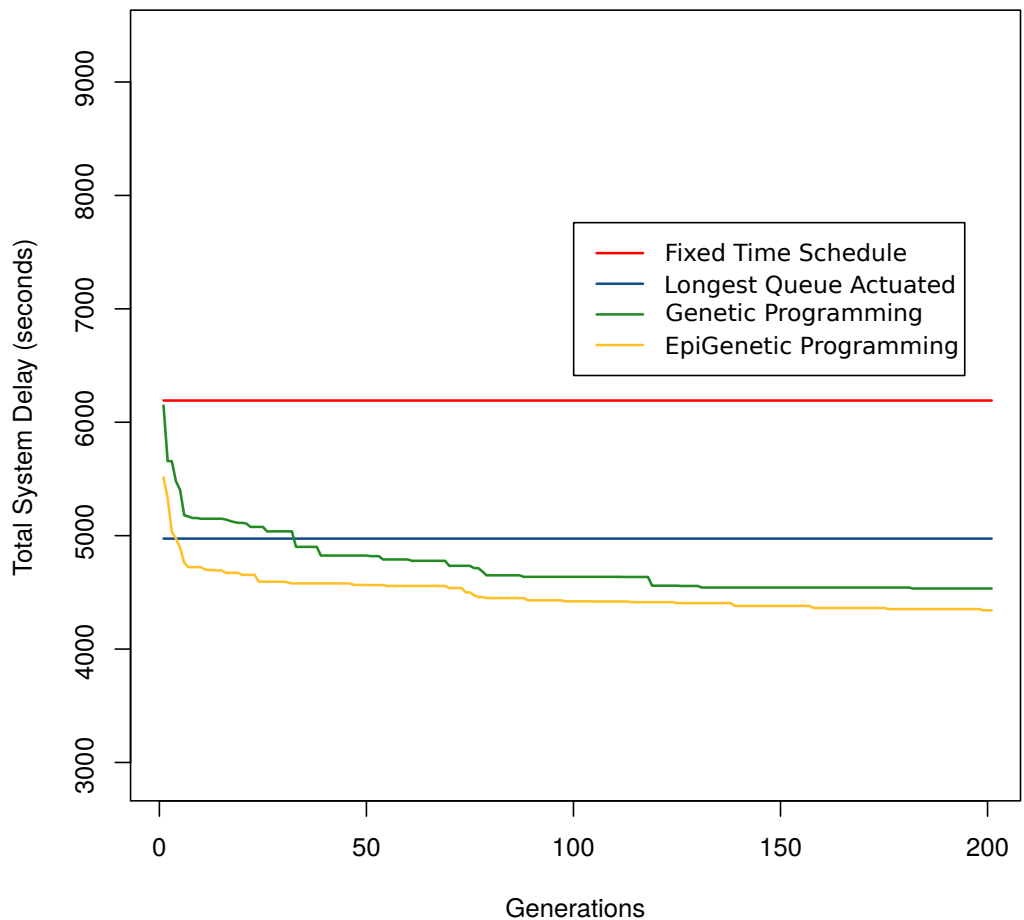}
\caption[Performance comparison for the single intersection scenario through SUMO]{Fitness curves of standard pre-timed, longest queue actuated control, standard GP and GP with the epigenetic mechanism for the single intersection scenario through SUMO. The $y$ axis representing Total System Delay does not start at 0 for this figure.}
\label{sumo_mySimVsSumo}
\end{figure}

Five independent executions were performed for each evolutionary method with both simulators. Figure \ref{mySim_mySimVsSumo} presents the comparison of 
the fitness obtained by the four methods with MiniTraSim. Figure \ref{sumo_mySimVsSumo} presents the comparison of the fitness obtained by the four 
methods with SUMO. 

For both figures, the total system delay of the single simulation run is displayed for all 
the generations of the fixed time schedule and the longest queue actuated controller. 
For standard GP and GP with the epigenetic mechanism, the average total system delay over five independent runs 
of the best individual in the population is displayed for each generation.

Table \ref{tab:mySimVsSumoDiff} presents the relative difference of the total system delay between the longest queue actuated method, 
the best individual for the last generation in GP, the best individual for the last generation in GP with the epigenetic mechanism and the  
standard pre-timed schedule divided by the total system delay of the pre-timed schedule with two different simulators 
(MiniTraSim and SUMO). GP with the epigenetic mechanism outperformed the other methods in both simulators.

\begin{table}
	\centering
	\caption[Relative performance differences for single intersection scenario]{Total system delay relative differences with standard pre-timed schedule for single intersection scenario.}
		\begin{tabular}{|p{5cm}|c|c|}
		\hline 
		\textbf{Method} & \textbf{MiniTraSim} & \textbf{SUMO} \\ \hline
		Longest queue actuated & 7.84 \% & 19.66 \%  \\  \hline
		Genetic Programming & 12.85 \% & 26.77 \% \\  \hline
		EpiGenetic Programming & \textbf{13.9 \%} & \textbf{29.87 \%}\\  \hline
	\end{tabular}
	\label{tab:mySimVsSumoDiff}
\end{table} 

\subsection{Results}
\label{subsec:singleResults}

From the results presented in Figures \ref{mySim_mySimVsSumo} and \ref{sumo_mySimVsSumo}, we conclude that Genetic Programming is able to evolve efficient traffic 
controllers for a single intersection without dependency on the representation used for the simulation of the traffic scenario. 
Furthermore, adding the epigenetic mechanism to Genetic Programming provides a consistent improvement to the solutions generated, even though the scenario 
does not change dynamically over the course of the generations.   

Although the traffic scenario does not change during the evolutionary process, the traffic conditions of the scenario change dynamically during 
the simulated hour. It appears that GP takes advantage of the epigenetic mechanism to evolve more adaptable controllers, even if the scenario 
is simpler and less variable than those used for the experiments presented in Chapter \ref{chap:synthScenarios}.  

For this basic scenario, the total system delay of the pre-timed schedule in SUMO (6250 seconds) is higher than the value reported for the same method in 
MiniTraSim (4762 seconds). Differences of this magnitude are normal given that both simulators use different vehicular models.
The relative differences between all the methods and the pre-timed schedule were greater for the experiments performed with SUMO than 
for the experiments performed with MiniTraSim as presented in Table \ref{tab:mySimVsSumoDiff}. Furthermore, the ranking of the methods tested is the
same using both simulators. In other words, the conclusions in terms of the performance of the approaches compared are the same independently of the
simulator used. However, it should be considered that the sample space compared is small (five independent executions).

The simulation of the single intersection traffic scenario with the fixed time schedule requires an average of $0.61979$ seconds in SUMO console mode. 
Executing the same scenario in MiniTraSim console mode requires an average of $0.46942$ seconds. These averages were calculated over 50 independent 
executions. Both simulators were executed on the same hardware with a dual Intel Core i5-7200U processor with a speed of 2.5 GHz and 4 GB of RAM 
running the version 16.04 of Ubuntu Linux operating system. This difference implies MiniTraSim is on average 24.26\% faster than SUMO for the single intersection 
scenario. This result was expected because MiniTraSim vehicular model is simpler than SUMO vehicular model.

\section{Real Network}
\label{sec:bologna}

Most research projects on heuristic methods in solution to the traffic signal control problem work with basic scenarios. 
Most of these scenarios are artificial and distant from real-world traffic conditions.
Recently, different scenarios based on real cities have been released for SUMO suite to be used by the research 
community \cite{codeca2015luxembourg, bieker2015traffic}. 
These scenarios contain an accurate representation of real networks and include 
real-world traffic data. We decided to test our algorithm in one of them. 

\subsection{Traffic scenario}
\label{subsec:bolognaScenario}

\begin{figure}
\centering
\includegraphics[scale=0.5]{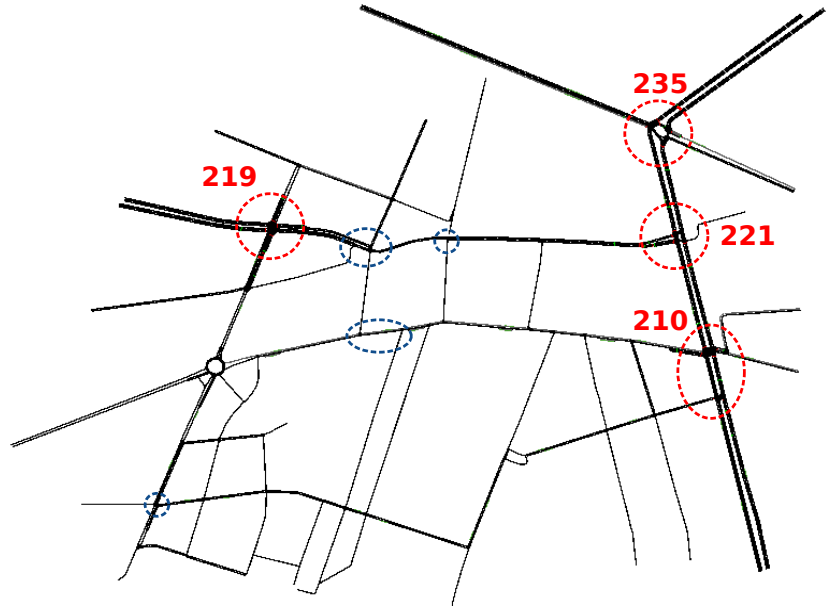}
\caption[Traffic network of the Andrea Costa scenario]{Traffic network of the Andrea Costa scenario. Only the intersections circled in red use 
actuated traffic controllers. The numbers in red correspond to identifiers of those intersections. Intersections circled in blue use fixed time schedules.}
\label{bologna_trafficNetwork}
\end{figure}

Due to the evaluation time required by our method, we chose the Andrea Costa scenario of the city of Bologna in Italy \cite{bieker2015traffic} 
presented in Figure \ref{bologna_trafficNetwork}. The model represents $2.45$ km\textsuperscript{2} of a real city with a 
total of 179 edges, 112 nodes, and eight intersections.  

The scenario includes one hour of real traffic for the morning peak between 8:00 am and 9:00 am.
More than 8,600 private vehicles and 160 buses are included in the scenario.
As described in \cite{bieker2015traffic}, the city of Bologna uses the UTOPIA system for traffic signal control. 
UTOPIA \cite{mauro1990utopia} optimizes traffic signal schedules and sorts the traffic light phases to satisfy traffic demand. 
All the experiments in this scenario are executed using the traffic schedule generated with UTOPIA.

A preliminary analysis of the scenario allowed us to identify the intersections 
most affected by traffic congestion. We focused our development on controllers for the four 
traffic intersections circled in red in Figure \ref{bologna_trafficNetwork} because they connect the main roads of the network and are the most 
congested intersections. The four intersections circled in blue in Figure \ref{bologna_trafficNetwork} connect secondary roads of single lanes. 
In our experiments, these four intersections use the fixed time schedule generated with UTOPIA. 
The experiments presented in this section were published in \cite{ricalde2017evolving}.

\subsection{Time-gap based actuated control}
\label{subsec:bolognaActuated}

Our approach is compared with an actuated controller based on time gaps already implemented in SUMO suite \cite{sumoWiki}.  
This controller works by prolonging the duration of traffic phases whenever a continuous stream of traffic is detected. 
The time-gap based actuated controller affects the cycle duration in response to dynamic traffic conditions. 
See Chapter \ref{subsubsec:timegap} for a more detailed description of time-gap actuated control strategies.

SUMO requires the usage of two additional phase attributes for intersections controlled by time-gap based actuated controllers.
The two additional attributes are minimum duration and maximum duration. These attributes are used to define the allowed range of durations 
for each phase. For the actuated control of the four optimized intersections in the Andrea Costa scenario, we used the maximum duration and minimum duration 
attributes generated by UTOPIA.

\subsection{Parameters}
\label{subsec:bolognaParams}

The scenario uses similar parameters for the Genetic Programming methods as those used in the experiments presented in Chapter \ref{chap:synthScenarios}. 
The parameters were presented in Table \ref{gpParams}. Sub-tree mutation \cite{banzhaf1998genetic} is used instead of point mutation and 
the mutation rate was changed to $20$\% per tree. 

The population size was reduced to 40 individuals to decrease the execution time of the algorithm. 
The terminal set includes the variables {\tt vQueue}, {\tt hQueue} (see Table \ref{gpTermSetVariables}) and integer numbers between
$-5$ and $5$. An independent controller is generated for each of the four intersections identified with indices 210, 219, 221 and 235. 
Therefore, GP evolves a forest of 4 programs. 

The mapping of the variables {\tt vQueue}, {\tt hQueue} and the connecting roads of an intersection in a real network is not a trivial task. 
Appendix \ref{sec:sumo_ligths} provides a description in detail of how it was performed for intersection 221.  However, the association of detectors and 
selection of affected phases for intersections 210 and 235 required a deeper analysis. 

\subsection{Experiments}
\label{subsec:bolognaExperiments}

Four different methods for traffic signal control were tested in the Andrea Costa scenario:
(1) the optimized pre-timed schedule generated with UTOPIA,
(2) the actuated controller based on time gaps described in Section \ref{subsec:bolognaActuated}, 
(3) evolution of actuated controllers through GP,
and (4) evolution of actuated controllers through GP with the epigenetic mechanism.
All these methods use the time schedules, minimum duration and maximum duration for each signal phase defined by UTOPIA and included in 
the Andrea Costa scenario.

For the time gap actuated method, GP and GP with the epigenetic mechanism, only the intersections identified with indices 210, 219, 221 and 235 in 
Figure \ref{bologna_trafficNetwork} are optimized.  

In this experiment, average delay is used as the fitness function instead of total system 
delay. Average delay is defined as the sum of the times
of all vehicles stopped in the system, divided by the number of vehicles considered. Average delay is commonly 
used to measure the performance of traffic signal control methods \cite{braun2011evolutionary, yang2016actuated}.
This measure was selected because it represents an average of the amount of time each vehicle is being delayed due to traffic congestion in the network
and because it is easy and fast to calculate from the data generated by SUMO suite.

The same route files are used for each simulation run. Therefore, the output for each execution of the 
optimized pre-timed schedule and time-gap actuated controller is the same.
Hence, only one run was required for the optimized pre-timed schedule and the time-gap based actuated controller. 
However, five independent executions were performed for GP and GP with the epigenetic mechanism. 

The execution of the Andrea Costa scenario requires an average of 34 seconds to run on a hardware with a dual Intel Core i5-7200U processor with speed of $2.5$ GHz 
and 4 GB of RAM running the version $0.25$ of SUMO and the version $16.04$ of Ubuntu Linux operating system. In GP, every individual requires a single 
simulation execution to be evaluated. Given the population size of 40 individuals and the total number of generations of 200, every independent run of standard 
GP and GP with the epigenetic mechanism required an average of $75.5$ hours only for the evaluation step. The large amount of time required by each 
execution constrained the number of independent executions to five for each method. The total amount of computational time required for the execution of the 
experiment was 31 days.

\begin{figure}
\centering
\includegraphics[scale=0.8]{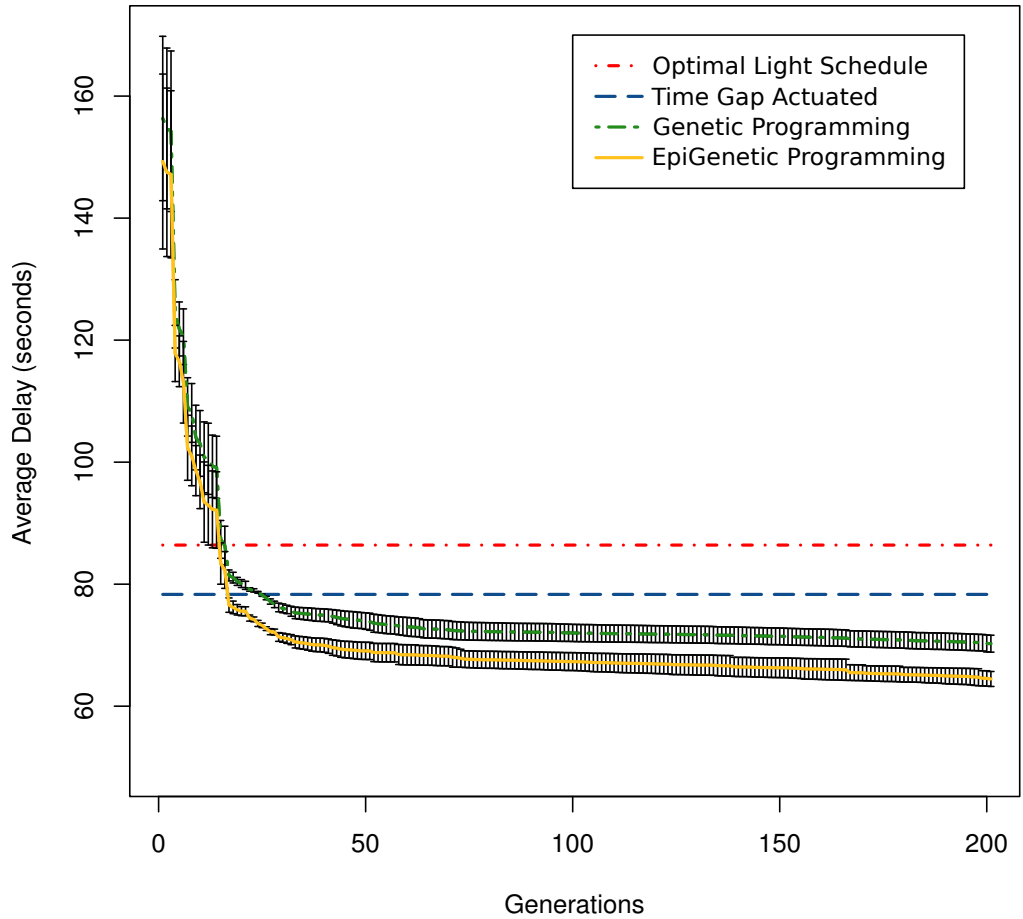}
\caption[Performance comparison for the Andrea Costa scenario]{Fitness curves of optimized pre-timed schedule, time-gap based actuated control, standard GP and GP with the epigenetic mechanism for 
the Andrea Costa scenario with standard error bars. The $y$ axis representing Average Delay does not start at 0 for this figure.}
\label{bologna_fitnessF}
\end{figure}

Figure \ref{bologna_fitnessF} presents a comparison of the results obtained by the
four methods. For GP and GP with the epigenetic mechanism the average and standard error of the best 
individual of the five independent executions are displayed for each generation. 

As expected with a randomly generated initial population, during the first 10 generations of GP and GP with the epigenetic mechanism, 
the fitness values are even worse than the value of the optimized pre-timed schedule. However, GP with the epigenetic mechanism only requires 
an average of 25 generations to evolve controllers that outperform the average delay produced by the time-gap based actuated controller. After that, 
the learning curve decelerates, but it keeps improving the solutions. Learning continues until the end of the run.
Therefore, an optimal set of controllers has not been found yet after 200 generations.

The epigenetic mechanism helps GP to find better solutions. Figure \ref{bologna_fitnessF} presents a similar behaviour for GP and GP with the epigenetic 
mechanism during the first 30 generations. However, the experiments with the epigenetic mechanism are able to generate solutions that
perform better in the scenario in later stages of the evolutionary process. 

It is worth noting that the difference between the final solution generated by GP with the epigenetic mechanism and the time-gap actuated 
controller is $71.46\%$ bigger than the difference between the time-gap actuated controller and the fixed time schedule. 

A non-parametric statistical Friedman test \cite{daniel1990applied} was performed over 5 independent executions of each of the four methods to test 
the statistical significance of the results presented. For standard GP and GP with the epigenetic mechanism only the results for the last generation are considered
in the test because they represent the best solution reached by each method in the static environment.   

\begin{table}
\centering \small
\caption[Friedman test ranks for Andrea Costa Scenario]{Friedman test ranks for Andrea Costa scenario.}
\begin{tabular}{|l|c|c|c|c|c|c|}\hline
\textbf{Method}&\textbf{Rank}&\textbf{Min}&\textbf{25th}&\textbf{Median}&\textbf{75th}&\textbf{Max}\\ \hline
Pre-timed schedule & 4 & 86.4077 & 86.408 & 86.408 & 86.408 & 86.408\\ \hline
Time-gap actuated & 3 & 78.3233 & 78.323 & 78.323 & 78.323 & 78.323\\ \hline
Genetic Programming & 2 & 66.2792 & 69.623 & 71.259 & 72.704 & 72.704\\ \hline
EpiGenetic Programming & \textbf{1} & 62.4788 & 62.479 & \textbf{62.699} & 65.627 & 67.750\\ \hline
\end{tabular}
\label{bologna_friedman}
\end{table}

From the Friedman test, we can conclude that there was a statistically significant difference in average delay depending on the method used in solution to the Andrea 
Costa scenario with ${\chi}^2(3) = 15$ and $p = 0.0018166$. Average ranks obtained by each method in the Friedman test are presented in Table \ref{bologna_friedman}. 
They indicate that Genetic Programming with the epigenetic mechanism has better performance than the other three methods tested.

Post hoc analysis of pairwise comparisons with $\alpha = 0.05$ were conducted through Dunn's tests \cite{dunn1964multiple}. 
The results of the Dunn's tests are presented in Table \ref{bologna_posthoc}. 
There were no significant differences between the pre-timed schedule and time-gap actuated controller ($z = 1.224745$, $p = 0.220671$),
between time-gap actuated controller and Genetic Programming ($z = 1.224745$, $p = 0.220671$), and between Genetic Programming and EpiGenetic Programming ($z = 1.224745$, $p = 0.220671$). 
However, there were statistically significant reductions in average delay for the Genetic Programming vs pre-timed schedule ($z = 2.44949$, $p = 0.014306$), 
EpiGenetic Programming vs pre-timed schedule ($z =  3.674235$, $p = 0.000239$), and EpiGenetic Programming vs time-gap actuated controller ($z =  2.44949$, $p = 0.014306$).

\begin{table}
\centering
\caption[Post hoc comparisons for Andrea Costa scenario]{Post hoc Dunn's test pairwise comparisons with $\alpha = 0.05$ for Andrea Costa scenario.}
\begin{tabular}{|p{6cm}|c|c|}
\hline
\textbf{Methods}&$z=(R_0 - R_i)/SE$&$p$\\
\hline
Pre-timed vs EpiGP& 3.674235 & \textbf{0.000239}\\ \hline
Pre-timed vs GP& 2.44949& \textbf{0.014306}\\ \hline
Time-gap actuated vs EpiGP& 2.44949& \textbf{0.014306}\\ \hline
Pre-timed vs Time-gap actuated& 1.224745& 0.220671\\ \hline
Time-gap actuated vs GP& 1.224745& 0.220671\\ \hline
GP vs EpiGP& 1.224745& 0.220671\\ \hline
\end{tabular}
\label{bologna_posthoc}
\end{table}

\subsubsection{Different traffic densities}
\label{subsec:bolognaDensities}

Any machine learning method runs the risk of over-training solutions. In other words,
they can generate solutions only suitable for the data frame used during the training phase. An over-trained solution lacks the 
capacity to perform well if the data frame is even only slightly modified. GP is not an exception to this phenomenon.

In our experiment, over-trained signal traffic controllers would not be able to adapt to traffic conditions different from 
the hour used as training set. This would mean that different controllers would be required throughout the day. However,
the idea of evolving controllers is precisely to eliminate the requirement of scheduled modifications. 
Therefore, to consider the experiment a success, the solutions generated by GP with the epigenetic mechanism should perform better 
than the fixed time schedule and the actuated controller under different traffic densities.  

An additional experiment was realized to validate that the solution generated by the evolutionary methods in the previous experiment were not 
over-trained with the traffic conditions used in the Andrea Costa scenario. 
In other words, we want to validate the adaptability of the controllers evolved with GP and GP with the epigenetic mechanism 
to different traffic densities.

\begin{figure}
\centering
\includegraphics[scale=0.6]{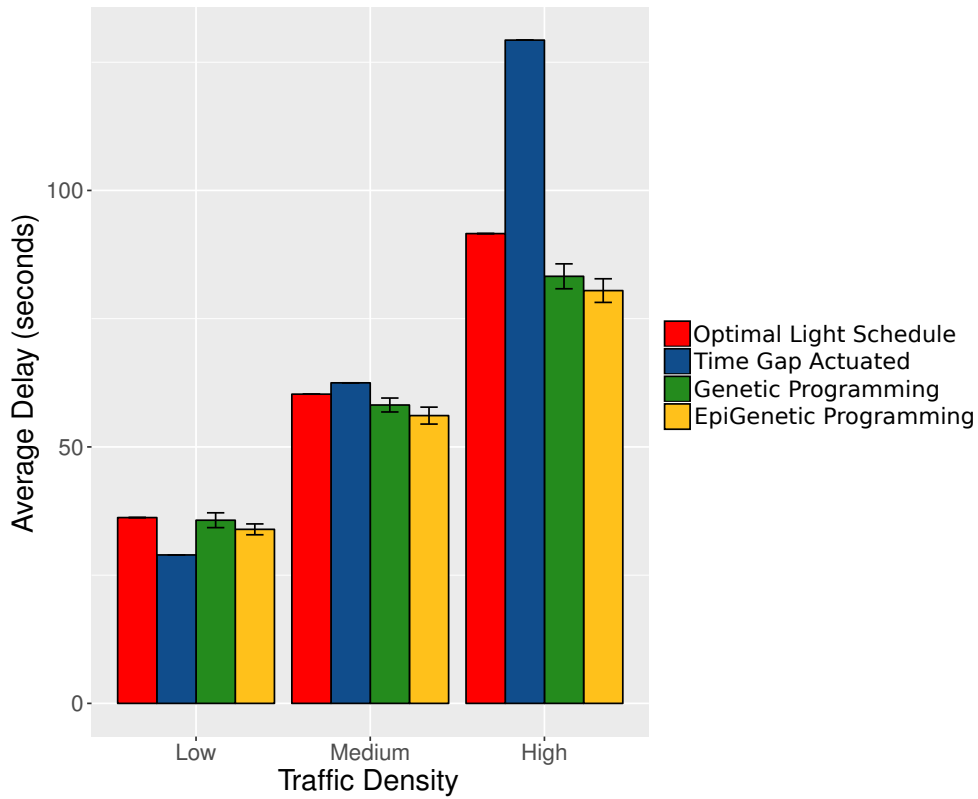}
\caption[Comparison of experiments with different densities for the Andrea Costa scenario]{Comparison of experiments with different densities for the Andrea Costa scenario.}
\label{bologna_densityCompF}
\end{figure}

Ideally, the controllers would be tested with real data of different hours for the same network.
However, the Andrea Costa scenario only provides traffic data for a single hour. To overcome this issue, three new scenarios 
with different traffic densities and random routes where generated for the Andrea Costa network using the tool called
{\tt randomTrips.py} provided with SUMO suite \cite{sumoWiki}. Each scenario has a duration of 60 minutes.

Figure \ref{bologna_densityCompF} presents the comparison of the four different methods for the three different density scenarios.
The average and standard error of the five independent experiments is presented for the standard GP method and for 
GP with the epigenetic mechanism. 

Table \ref{tab:bolognaDiff} presents the percentage of improvement of the actuated controller, GP and GP with the epigenetic mechanism over the 
optimized pre-timed schedule for the four different experiments realized with the Andrea Costa scenario. Negative values indicate that the optimized pre-timed schedule 
was a better solution than the given method. In three of the four scenario configurations, GP with the epigenetic mechanism outperformed 
the other methods.

For the three scenarios with different traffic densities, GP with the epigenetic mechanism improved the results of the optimized pre-timed schedule. 
However, the level of improvement is less than the improvement observed for the 
experiment with real data, as it can be seen in the second, third and fourth columns of Table \ref{tab:bolognaDiff}.
  
\begin{table}[!t]
    \renewcommand{\arraystretch}{1}
	\caption[Relative differences for Andrea Costa scenario with different densities]{Average delay relative differences with optimized pre-timed schedule for Andrea Costa scenario with different densities.}
	\label{tab:bolognaDiff}
	\centering
		\begin{tabular}{|l|c|c|c|c|}
		\hline 
		\textbf{Method} 
      & \multicolumn{1}{|p{0.4in}|}{\textbf{Training}} 
      & \multicolumn{1}{|p{0.7in}|}{\textbf{Low Density}} 
      & \multicolumn{1}{|p{0.8in}|}{\textbf{Medium Density}} 
      & \multicolumn{1}{|p{0.8in}|}{\textbf{High Density}} \\    \hline
		Time-gap Actuated & 9.36 \% & \textbf{20.1 \%} & -15.17 \% & -41.18 \%  \\  \hline
		Genetic Programming & 18.70 \% & 1.42 \% & 3.50 \% & 9.09 \% \\  \hline
		EpiGenetic Programming & \textbf{23.23 \%} & 6.33 \% & \textbf{6.93 \%} &\textbf{12.13 \%}\\  \hline
	\end{tabular}
\end{table} 

The behaviour of the time-gap actuated method in these scenarios presents some particularities. It obtained the best performance in the scenario
with low density. However, it behaved poorly in the scenarios with medium and high densities. The time-gap actuated method
works by prolonging traffic phases whenever a continuous stream of traffic is detected. However, especially in the scenario with high
traffic density, several of the intersections are constantly saturated. Therefore, the actuated method is not able to optimize 
the duration of the phases when traffic is over-saturated in every direction.  

To test the statistical significance of the difference between Genetic Programming and GP with the epigenetic mechanism in the scenarios with
different traffic densities, three different null hypothesis tests, one for each of the individual scenarios, were performed. The results of these tests are presented in table
\ref{tab:bolognaDensitiesDiff}. 

\begin{table}[!t]
    \renewcommand{\arraystretch}{1}
	\caption[Statistical analysis of Andrea Costa scenario with different traffic densities]{Null hypothesis tests for Genetic Programming and EpiGenetic Programming in Andrea Costa scenario with different traffic densities.}
	\label{tab:bolognaDensitiesDiff}
	\centering
		\begin{tabular}{|l|c|c|c|c|}
		\hline 
		\textbf{Method} & \textbf{Variability} & \textbf{Freedom degrees} & \textbf{$t$ score} & \textbf{$p$ value} \\  \hline
		Low Density &  0.513355 & 8 & 3.465355 & \textbf{0.0085} \\  \hline
		Medium Density & 0.956051 & 8 & 2.159742 &   0.0628 \\  \hline
		High Density & 1.497586 & 8 &  1.858957 & 0.1001  \\  \hline
	\end{tabular}
\end{table} 

From the results of the null hypothesis tests presented in Table \ref{tab:bolognaDensitiesDiff}, we can conclude that although GP with 
the epigenetic mechanism performs better than GP in all the experiments, the differences for the scenarios with medium density and high density 
are not statistically significant. These results mean that even if GP with the epigenetic mechanism is able to generate better solutions, 
we cannot conclude the controllers created by both methods are statistically different in their capacity to adapt to changes in the traffic density 
given the sample space of the experiments performed (five independent executions) in the scenarios with different traffic densities.

However, the evidence presented demonstrates that the signal traffic controllers generated by both evolutionary methods are not over-trained 
for the hour of traffic peak used during the evolutionary process. 
Consequently, the controllers can be used throughout the day and reduce the average delay of the
vehicles circulating in this specific section of the city.   

\subsection{Results}
\label{subsec:bolognaResults}

The experiments determined that Genetic Programming is able to generate traffic signal controllers adaptable to traffic variation for a
real scenario of small size. The method was compared with an optimized pre-timed schedule and an actuated control based on time gaps. 
The results display an improvement of $23.23\%$ over the optimized pre-timed schedule, and 
an improvement of $17.7\%$ over an actuated control included in SUMO suite based on time gaps.
These differences are statistically significant as demonstrated in Section \ref{subsec:bolognaExperiments}.
The generated solutions were tested using scenarios with different traffic densities. 

Genetic Programming evolved an independent controller for each intersection. 
This approach differs from other heuristics where the goal is a general control mechanism. 
Hence, the adaptive local controllers generated by GP are able to adapt to unexpected traffic density changes.   

The adaptive controllers generated do not require communication between intersections, and modifications to the traffic 
schedules are based on local information. Standard GP and GP with the epigenetic mechanism only use the information provided by entry-exit detectors
to generate reactive signal controllers. This approach reduces the cost of implementation in the real-world.
However, the model can be easily extended to consider communication between intersections or between vehicles and intersections. 

The experiments presented in Section \ref{subsec:bolognaScenario} used the traffic schedule generated by UTOPIA. 
Additional experiments are required to analyze the performance of GP with the epigenetic mechanism in the Andrea Costa scenario when 
an optimized traffic schedule is not available.

\section{Discussion}
\label{sec:sumoDisc}

This chapter presented a comparison of different signal control methods for two additional traffic scenarios executed with SUMO. 
SUMO is an open-source, free to use and multi-platform microscopic traffic simulation suite widely used by the research community that 
was designed to handle large road networks and allows online signal manipulation through standard Python. 

The results displayed an average improvement for GP with the epigenetic mechanism of 21.88\% over the standard pre-timed schedule  
in the single intersection scenario and an improvement of 23.23\% over optimized pre-timed schedule for the Andrea Costa scenario 
(this difference is statistically significant as demonstrated in Section \ref{subsec:bolognaExperiments}). 
Both scenarios use a single hour of traffic in the evolution of
adaptive actuated controllers. This is a different approach to the scenarios presented in Chapter \ref{chap:synthScenarios}, where a 
time interval window configuration was used to change the traffic conditions along more than 16 hours. 

This modification implies that the scenarios discussed in this chapter present less variability than those of the previous chapter. 
However, the epigenetic mechanism still provides an average improvement over standard Genetic Programming of 4.24\% in the single intersection scenario  
and an improvement over standard Genetic Programming of 8.23\% for the Andrea Costa scenario (this difference is statistically significant as demonstrated in Section \ref{subsec:bolognaExperiments}). 
A possible explanation for this behaviour is that traffic is a dynamic system and GP takes advantage of the epigenetic mechanism to evolve more adaptable controllers
using the intrinsic variability of the traffic signal control problem.

GP with the epigenetic mechanism can be compared with a broader set of methods used to optimize traffic signal control. As part of this research, an attempt
to compare it with the self-organizing method described in \cite{zubillaga2014measuring} was done. However, it was not possible to execute an implementation
in SUMO suite of the mentioned method with a standard traffic scenario.  

Larger scenarios should be evaluated to analyze the behaviour of the method under the influence of different circumstances. 
Codeca, et al. \cite{codeca2015luxembourg} 
presented a full city scenario with 24 hours of real traffic data. However, testing GP with the epigenetic mechanism in a scenario of such size requires 
modifications to the framework proposed in Appendix \ref{apdx:sumo_diff}. The main limitation in the evolution of traffic controllers for larger
scenarios with GP is the computational time required by the simulation of the scenario. An alternative solution is to use parallel computing to reduce 
simulation clock time. However, the current implementation of TraCI and SUMO cannot be executed using parallel computation.   
 % External Simulator

\chapter{Concluding Remarks}
\label{chap:conclusions}

  \graphicspath{{figures/PNG/}{figures/PDF/}{figures/EPS/}{figures/}}
  \lhead{\emph{Chpt 7: Concluding Remarks}}  % Set the left side page header to "Concluding Remarks"

\section{Summary}

This thesis introduced a novel epigenetic mechanism that considers environmental changes to trigger epigenetic mutations.
The proposed mechanism is inspired by DNA methylation process. It silences sections of the chromosome using
additional markers. Those markers are modified based on changes to environmental conditions during the lifespan of the
individual. The markers are transferred to an individual's offspring in a similar manner as DNA methylation markers are transferred 
during cellular division.

Genetic Programming with the epigenetic mechanism was tested in the traffic signal control problem. Urban traffic signal control is a complex 
combinatorial problem characterized by high variability and high uncertainty that affects the daily life of many citizens. The proposed mechanism improves 
the performance of Genetic Programming in the evolution of actuated controllers for traffic signals because it facilitates the adaptation of the individuals 
to dynamic changes of the traffic conditions. 

The evolved controllers are not pre-calculated signal schedules but adaptive rules evolved for online modification of the duration of signal phases. 
This approach adapts to dynamic changes in traffic density and requires less monitoring and less human interaction because it dynamically 
adjusts the signal behaviour depending on the local traffic conditions of each optimized intersection of the traffic network.

From a computer science perspective, it is worth noting that there is a lack of standard test suites and benchmarks for the traffic signal control 
problem. We can group the papers of the literature review into two different sets: those that use a basic scenario consisting of a few intersections 
and compare their method with fixed time schedules 
\cite{abu2003design, haijema2017dynamic, zhang2009multi, li2016traffic, padmasiri2014genetic, yuan2016optimal},
and those that use a scenario based on a real world network and compare their method with commercial solutions 
\cite{robertson1969transyt, sanchez2010traffic, sims1980sydney, mauro1990utopia, braun2011evolutionary, covell2015micro, yang2016actuated}.  

This state of affairs has been reported as a problem in the field \cite{balaji2010urban, gokulan2010distributed, li2016traffic} because it 
complicates the objective comparison of different methods. However, in recent years, different scenarios based on real cities have been released to 
be used by the research community \cite{codeca2015luxembourg, bieker2015traffic}. These scenarios contain an accurate representation of real 
networks and include real-world traffic data. The research community working on traffic signal optimization should promote the use of standard test cases
to facilitate the objective comparison of different solutions with the aim of defining verified benchmarks. For this reason, the Andrea
Costa scenario (see Chapter \ref{sec:bologna}) was included in the set of scenarios tested in this research project.

Our epigenetic mechanism was evaluated in four traffic scenarios with different properties and traffic conditions, including the scenario 
presented in \cite{bieker2015traffic}, using two different microscopic simulators. Genetic Programming was able to generate competitive actuated controllers 
for all the scenarios tested. The incorporation of the epigenetic mechanism improved the performance of GP in all the scenarios tested. 
The level of improvement depends on the conditions of each scenario. 
The single intersection scenario presented the lowest improvement, 4.24\%, for the epigenetic mechanism over standard GP.
The highway scenario presented the highest improvement, 10.34\%, for the epigenetic mechanism over standard GP 
(this difference is statistically significant as demonstrated in Chapter \ref{subsec:hgwResults}). 
The results of the highway scenario and the Andrea Costa scenario were analyzed using statistical tests and the
differences in the performance of some methods proved to be statistically significant.

A microscopic simulator was created from the ground up as part of the work for this thesis. MiniTraSim is able to simulate changing dynamic 
traffic conditions of complex traffic networks with several roads of multiple lanes using an approach with low computational cost. 
The vehicular model of MiniTraSim is based on the Traffic Cellular Automata described in \cite{nagel1992cellular} and \cite{sanchez2010traffic}, but incorporates 
concepts borrowed from multi-agent systems and object oriented programming to reduce the computational cost.

A generic framework to evaluate traffic controllers written in Python using SUMO was designed and developed for this thesis. The framework is open-source 
and is available in a sourceforge repository \cite{ricalde2017sorceforge}. The released configuration files were created to work with the scenario presented in 
Chapter \ref{sec:bologna}, but they can be modified to work with any traffic network in SUMO. This is the first open-source configurable 
framework to test machine learning methods on the traffic signal control problem using SUMO. 

\section{Future Research}

All the experiments presented in this thesis indicate that the epigenetic mechanism provides an improvement in evolution of traffic signal controllers when
it is used with Genetic Programming. Future work could be directed towards tweaking the method to be efficient 
under different conditions.
The value returned by the evolved traffic controllers can be used as multiplication factor of 
the duration of signal phases instead of addition or subtraction. 
Based on the approach presented in \cite{gershenson2012adaptive} and \cite{zubillaga2014measuring}, all the incoming vehicles within a predefined 
distance $d$ from the signal can be included in the traffic queue variables, whether they are stationary or moving, instead of only considering stopped vehicles. 
By taking into consideration the moving vehicles, the algorithm may become less reactive and more predictive in terms of traffic congestion.   

The epigenetic mechanism can be tested in the optimization of larger realistic scenarios. 
For example, Luxembourg full city scenario with 24 hours of real traffic data presented by Codeca, et al. \cite{codeca2015luxembourg}.
However, the computational time required for execution of GP with the epigenetic mechanism in a scenario of such size requires modifications to the 
framework proposed by this research and presented in Appendix \ref{apdx:sumo_diff}. An alternative solution is to use parallel computing to reduce simulation 
clock time. However, a different traffic simulator needs to be used for such a task because TraCI and SUMO cannot be executed using parallel computation.

This method can be compared with a broader set of methods used to optimize traffic signal controllers. Examples are 
the self-organizing method described in \cite{zubillaga2014measuring} and the green wave method described in \cite{yang2016actuated}.
As part of this research, an attempt to compare EpiGenetic Programming with the self-organizing method described in \cite{zubillaga2014measuring} 
was done. However, due to implementation problems, the results obtained were not consistent with \cite{zubillaga2014measuring} and the experiment 
was terminated. 

Moreover, the framework presented in this thesis can be used to test different machine learning methods, such as Learning Classifier 
Systems \cite{lanzi2003learning}, Support Vector Machines \cite{steinwart2008support} and  Deep Neural Networks \cite{collobert2008unified}, 
in a real traffic scenario without major code modifications. However, the exhaustive computational time required by traffic simulation should be
considered in the design of any further projects.

The field of traffic control is in a transition towards the inclusion of information obtained through vehicle to vehicle communication 
and vehicle to signal communication \cite{feng2015real, tuohy2015intra, goel2017self, salman2018fuzzy}. Experiments including this type of additional 
information in the variable set of Genetic Programming with the epigenetic mechanism can be performed and compared with the approach presented in this 
thesis and other methods.  

After the proposed method is tested in more scenarios and tweaked to be efficient under different conditions, it may be ready to be implemented in a real
pilot test. Such test should be composed by six different stages: (1) create a traffic simulation scenario of the network to optimize;
(2) evolve traffic controllers for the scenario created; (3) test traffic controllers with validation cases of the scenario not included in the training;
(4) measure the level of improvement provided by the evolved controller for the traffic scenario; in case the improvement of the previous step was significant, 
(5) deploy the evolved controllers in the real traffic network to be tested during uncritical time schedules; and in case the performance of the evolved 
controllers in the real world is similar to the behaviour observed in the simulator, (6) gradually increase the use of the controllers during standard and 
critical time schedules. Local government and society should be involved in such pilot test and their feedback at every stage should be  listened and 
considered as part of the design for the next stage and future implementations.

The epigenetic mechanism can be used to solve other problems. Problems where some elements of the domain vary with the 
progression of time (dynamic environments) can benefit from the epigenetic mechanism proposed in this thesis, 
even if minor modifications to the method would need to be considered. 

Finally, new modifications to EAs inspired by different epigenetic mechanisms, such as RNA silencing and prions, can be designed and tested. 
Epigenetic mechanisms play an important role in biological processes such as phenotype differentiation, memory consolidation within generations 
and environmentally induced epigenetic modification of behaviour. Novel epigenetic mechanisms may improve the performance of Evolutionary Algorithms in 
the solution to many different problems where current implementations fall short, such as the time-varying knapsack problem \cite{he2016algorithms},
the job shop scheduling problem \cite{applegate1991jobshop} and weather forecasting \cite{la2016automatic}.
 % Concluding Remarks

%% ----------------------------------------------------------------
% Now begin the Appendices, including them as separate files

\addtocontents{toc}{\vspace{2em}} % Add a gap in the Contents, for aesthetics

\appendix % Cue to tell LaTeX that the following 'chapters' are Appendices

\chapter{Building a Traffic Simulator}
\label{apdx:our_sim}

  \graphicspath{{figures/PNG/}{figures/PDF/}{figures/EPS/}{figures/}}
  \lhead{\emph{Appx A: Building a Traffic Simulator}}  % Set the left side page header to "Building a Traffic Simulator"

Although several commercial and open source simulators are available, a microscopic model simulator was created from the ground up as part of this 
research project. This decision was motivated by multiple reasons. The goals were to have full control of the environment and to simulate changing dynamic 
traffic conditions of complex traffic networks with several roads of multiple lanes using an approach with low computational cost. Our simulator is 
called \textit{Minimum Traffic Simulator} (MiniTraSim).

We decided to use the microscopic approach for traffic simulation because it better models the complex properties of urban traffic behaviour than the
other two approaches (see Chapter \ref{subsec:microscopic}). MiniTraSim was built in C\# because we wanted to use an object oriented programming language 
and the convenience of a standard graphic library. Only standard libraries and standard components were used in MiniTraSim implementation to allow portability 
to other platforms using Mono \cite{xamarin2018mono}.

MiniTraSim can be executed in graphic mode and console mode. The graphic user interface for the graphic mode was presented in Figure \ref{mySim_gui}.
The internal logic of MiniTraSim works in a similar way to the cellular automaton described in \cite{nagel1992cellular} and \cite{sanchez2010traffic}, 
but incorporates concepts borrowed from multi-agent systems and object oriented programming to allow the simulation of multiple roads. The time step used 
for the temporal discretization is of one second.

MiniTraSim operates in a two-dimensional matrix. Every position in the matrix has three possible types of values: road, land (no-road) or the identifier 
of a vehicle in that position. The representation used allows the modelling of multiple roads with multiple lanes and multiple directions. The toroidally 
closed environment was replaced by an open environment with entry nodes for different roads. The probability of entry to the network is 
modelled using a Poisson distribution. 

Additional layers were incorporated to integrate information of simulated components (roads, intersections, traffic signals and vehicles). For example, an object 
stores the current position, current speed, current direction, current road and final destination of each individual vehicle. Using this information, 
the model only requires to update the matrix positions identified as potential changes, instead of updating the complete matrix at each time step, 
thus reducing the computational costs of the simulation.

New rules were incorporated to those defined by Nagel and Schreckenberg \cite{nagel1992cellular} to address inter-vehicular distance on intersections and react
to lane changing behaviour. The remainder of this appendix provides a detailed description of MiniTraSim properties and how they were implemented. 

\section{Road Representation}
\label{sec:raod_rep}

During simulation time, the traffic network is stored as an integer two-dimensional matrix called \textit{network matrix}. Every cell of the network matrix represents 
a $3\times3$ $m^2$ piece of land. The integer value stored in the cell indicates the type of object contained in that space at the given time step: the value 0 
is used for any type of land different from road or vehicle and is ignored by MiniTraSim. A road is represented using the value 1, any positive integer greater 
than 1 indicates that a vehicle (identified with the given number) is located on that cell position at the given time step.
      
\begin{figure}
	\centering
	\includegraphics[scale=1.2]{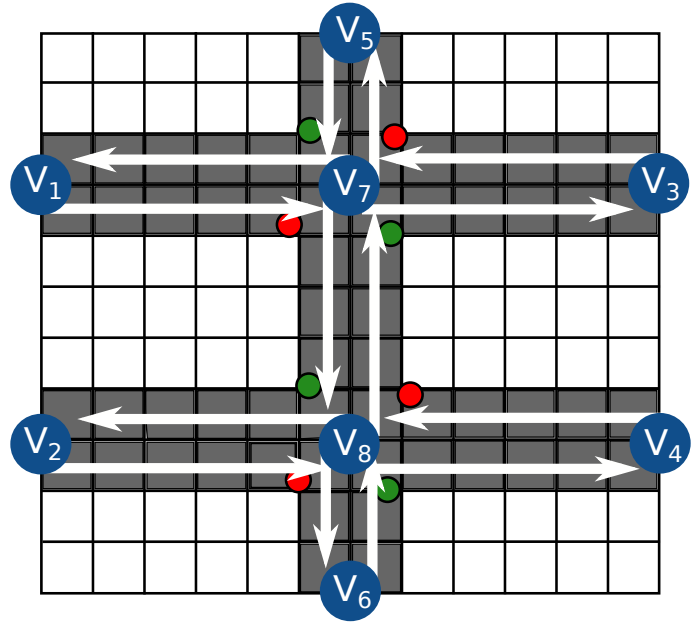}
	\caption[MiniTraSim layer for intersections and roads]{Representation of simulation step including information of MiniTraSim layer for intersections and roads.}
	\label{sim_internalRep_roads}
\end{figure}

In addition to the  matrix representation, the traffic network is stored in an independent layer as a simple directed 
graph $G = (V, A)$ where $V$ represent the entry/exit nodes and intersections and $A$ (arrows) represent all the connecting roads. In MiniTraSim, 
the graph is represented as a dynamic data structure. For each node, the system stores its position in the matrix, an indicator in case it is an entry/exit
node and a list of the nodes connecting \textit{from} and \textit{to} that node. For each arrow, the system stores the connecting nodes, an indicator in 
case the road is bidirectional and the number of lanes in that direction.

In our model, a node $V_i$ cannot have arrows that connect to itself. This implies that dead end streets cannot be modelled
in MiniTraSim. Figure \ref{sim_internalRep_roads} presents an example of the matrix representing a small network with the additional 
information stored for intersections and roads.

The use of a cellular automaton to model the behaviour of the system implies additional limitations \cite{maerivoet2005cellular}. Only roads 
with no curves can be modelled. The roads can only be oriented in the main four cardinal directions: north, west, south and east. It is clear 
that the number of real world scenarios that can be modelled in MiniTraSim given those limitations is reduced. 

More complex approaches, such as the stochastic car-following model \cite{krauss1998microscopic}, allow to work with more realistic conditions. However,
given the time constraints of this research project and the extensive number of simulations that were required by each experiment, 
it was decided to use cellular automata to represent vehicle dynamics.

\begin{figure}
	\centering
	\includegraphics[scale=0.6]{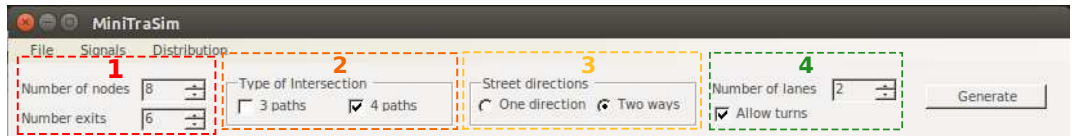}
	\caption[MiniTraSim traffic network generation menu]{MiniTraSim menu with parameters for the generation of the traffic network.}
	\label{mySim_header}
\end{figure}

A traffic network can be entered to MiniTraSim in two ways: configuring a new traffic network using the tool menu presented in 
Figure \ref{mySim_header} or importing a manually generated text file using a custom format. MiniTraSim includes a basic tool to generate a 
random traffic network of fixed size and variable complexity. The user can select different parameters marked by numbers in 
Figure \ref{mySim_header}: (1) number of intersections and entry/exit nodes, (2) type of road connections, (3) type of roads: single direction 
or bidirectional, and (4) number of lanes and indicator to allow left turn in the intersections. 

The tool only provides basic functionality. Every node in the network generated shares the same properties. However, the properties 
of individual intersections can be modified once the network is defined.

After the network is generated, a sequence of tasks is automatically executed as a required post-processing procedure. The system validates 
that all the nodes have access to an exit. Any node without a connection is removed from the network. All the required traffic signals are generated 
for the intersections. Finally, a copy of the graph representing the traffic network is transformed into the two-dimensional matrix used in MiniTraSim.   

\section{Traffic Signals}
\label{sec:signal_control} 

\begin{figure}
	\centering
	\includegraphics[scale=1.2]{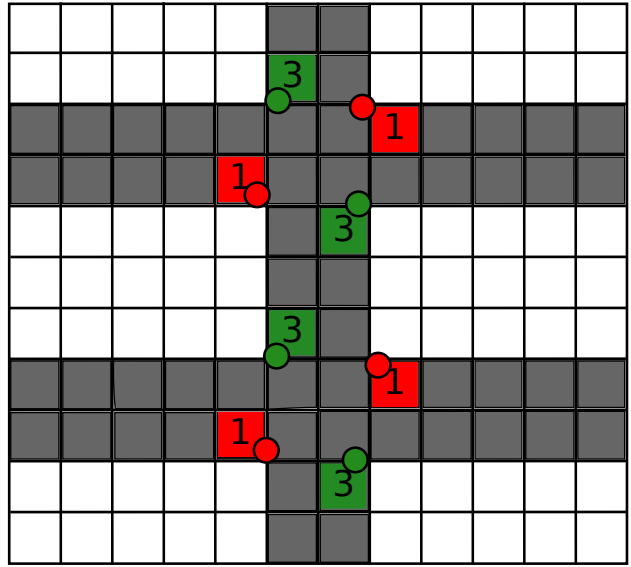}
	\caption[MiniTraSim layer for signal status]{Representation of simulation step including information of MiniTraSim layer for signal position and status.}
	\label{internalRepresentation_lights}
\end{figure}

In the simulation environment, the traffic signals are represented as an additional integer matrix called \textit{signal matrix}. The signal 
matrix has the same size as the network matrix. Each signal covers all the positions in the signal matrix corresponding to road positions in 
the network matrix that are controlled by the signal. All the positions not associated with a signal have value 0. Value 1 indicates a signal 
with stop status (red light). Value 2 indicates a signal with status of pass with caution (yellow light). Value 3 indicates a signal with status 
of pass (green light). Value 4 indicates a signal with status of turn left (green arrow to the left). Figure \ref{internalRepresentation_lights} 
presents an example of the positions in the signal matrix different from 0 of a small traffic network.   

 \begin{table}
    \caption[MiniTraSim default traffic signal schedule with left turn]{MiniTraSim default traffic signal schedule with left turn.}
      \label{lightLapses_turn} 
      \centering 
      \begin{tabular}{|c|c|} \hline
       \textbf{Light} & \textbf{Duration} \\ \hline
		Red &  35 seconds  \\ \hline
		Green &  25 seconds  \\ \hline
		Green arrow to the left &  5 seconds  \\ \hline
		Yellow &  5 seconds  \\ \hline
      \end{tabular}
 \end{table}

\begin{figure}
	\centering
	\includegraphics[scale=1]{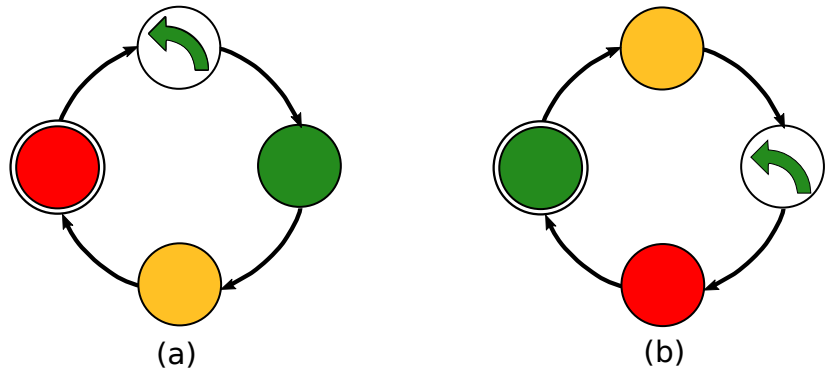}
	\caption[MiniTraSim traffic signal plan considering left turn]{MiniTraSim traffic signal plan considering left turn. The plan (a) is used for all the 
	signals controlling traffic in the vertical direction. The plan (b) is used for all the signals controlling traffic in the horizontal direction.}
	\label{lighSched_turn}
\end{figure}

\begin{figure}
	\centering
	\includegraphics[scale=1]{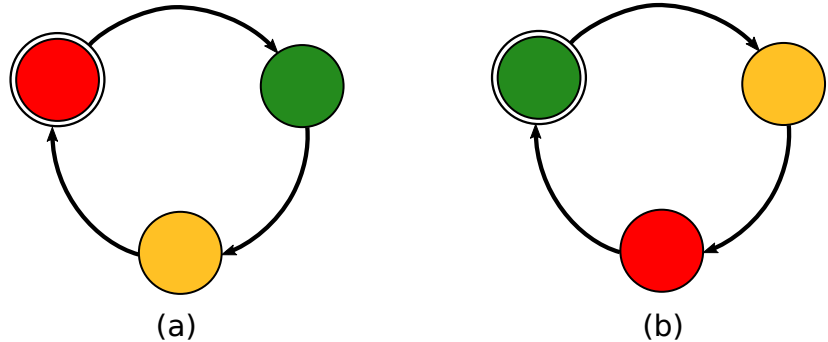}
	\caption[MiniTraSim traffic signal without left turn]{MiniTraSim traffic signal plan without left turn. The plan (a) is used for all the signals controlling traffic 
	in the vertical direction. The plan (b) is used for all the signals controlling traffic in the horizontal direction.}
	\label{lighSched_noTurn}
\end{figure}

All traffic signals in the network are synchronized and remain like that unless they are modified. To achieve this synchronization, all intersections share the 
same default signal schedule. If the network allows to turn left, the signals follow the plan presented in Figure \ref{lighSched_turn}. The duration for 
each of the phases is presented in Table \ref{lightLapses_turn}. If the network does not allow to turn left, the signals follow the plan presented 
in Figure \ref{lighSched_noTurn} with the signal phase schedule presented in Table \ref{lightLapses_noTurn}.

Additional functionality was implemented to keep count of the number of vehicles stopped in every direction for all the intersections. A set of counters 
(one for each traffic signal) are update every second and the information can be used for local decision by the traffic controllers (see Chapter \ref{chap:synthScenarios}). 

A list with additional information for all the traffic signals is used to speed up control and update operations of the signal matrix. For each traffic signal, 
the list contains its position in the matrix, its current status, the signal plan, duration of each phase, references 
to the associated intersection and associated road, internal operational clock, and the counter for the number of vehicles stopped in the controlled 
road. A change in the status of a signal in the list triggers an update of the associated cell positions of the signal matrix.

 \begin{table}
     \caption[MiniTraSim default traffic signal schedule without left turn]{MiniTraSim default traffic signal schedule without left turn.}
      \label{lightLapses_noTurn} 
      \centering 
      \begin{tabular}{|c|c|} \hline
       \textbf{Light} & \textbf{Duration} \\ \hline
		Red &  33 seconds  \\ \hline
		Green &  27 seconds  \\ \hline
		Yellow &  6 seconds  \\ \hline
      \end{tabular}
 \end{table}

\section{Entry Probability}
\label{sec:entry_prob}

The Poisson distribution correctly models arrival of vehicles, on one or multiple lanes \cite{mauro2013update}. The 
flexibility of the Poisson distribution allows to simulate changing traffic densities.

The probability that a vehicle arrives at a specific lane for a given entry node $P'(t)$ is calculated using Equation \ref{eq:entry}, where
$t$ represents the time step of the simulation, $P(t)$ corresponds to the Poisson probability mass function, $c$ is a modifier constant between $[0,1]$ 
used as the floor of the probability and $p_{max}$ is the maximum possible value for $P'(t)$ defined between $[0,1]$
\begin{equation} \label{eq:entry}
P'(t) = min\{P(t)+c,p_{max}\}.
\end{equation}  
The Poisson probability mass function $P(t)$ is calculated using Equation \ref{eq:poissonPmf}, where $t$ represents the time step of the simulation and
$\mu$ represents the average number of vehicles that arrive at the entry node
\begin{equation} \label{eq:poissonPmf}
P(t) =  \frac{\mu^t}{t!} e^{-\mu}.
\end{equation}  
The average number of vehicles that arrive at the entry node $\mu$ is calculated using Equation \ref{eq:averageVehic}, where $V$ represents the total volume of
vehicles to arrive for that entry node and $T$ represents the total amount of time to be modelled using the distribution 
\begin{equation} \label{eq:averageVehic}
\mu = \frac{V}{T}.
\end{equation}  
Traffic waves are a really common phenomenon in urban traffic networks. To be able to represent traffic waves in MiniTraSim, each entry node in the network is associated with 
a set of arrival probabilities $P'_i(t)$. Each arrival probability distribution has non-overlapping start and end time steps and different parameters $c$, 
$p_{max}$, $V$ and $T$. During simulation time, a vehicle is inserted in an entry node if a random number between $[0,1]$ is in the range of the 
corresponding arrival probability $P'_i(t)$ and the entry position in the network matrix is not currently occupied by other vehicle. 

\section{Routing}
\label{sec:routing}

Vehicle routing is a complex problem \cite{ritzinger2016survey}. In most of the existing microscopic simulators, the calculation of the routes is 
not done during simulation time, but as part of the pre-processing for the traffic scenario \cite[pp. 130]{SUMO2012}. The reason for this is the high 
computational cost for the calculation of shortest routes. 

Once the origin node and destination node are defined and given the constraints of only considering the network topology, the definition of 
vehicle routes can be reduced to the problem of finding the shortest path \cite{ahuja1990faster}. However, additional variables such as dynamic 
traffic density, number of lanes and road conditions need to be considered to be able to propose efficient routes. Furthermore, in a real scenario, 
not all drivers follow the shortest path to arrive at their destination. Some drivers prefer to drive through roads they know better. Other drivers 
get lost and randomly wander around in the city trying to find their destination. 

To consider some of these aspects, we decided to consider two different vehicle behaviours in MiniTraSim: random walk and fixed destination. The 
behaviour of a specific vehicle is defined when the vehicle is inserted into the network. For the fixed destination behaviour, the final destination
is also defined when the vehicle is inserted in the network. Unlike SUMO and other simulators, MiniTraSim defines the vehicle routes during simulation time.

Random walk is the simpler of both behaviours. If a vehicle approaches an intersection, the driver will randomly select the next road from a list 
of all possible connections of the intersection. The possible connections are defined based on the current lane of the vehicle. To avoid U turns, 
the road which the vehicle transits is removed from the list. This procedure is repeated for every intersection until an exit node is reached. 

For vehicles following fixed destination behaviour, calculating the shortest path would require the program to explore the full network when the 
vehicles are inserted and to store in memory the defined routes. To reduce computational cost and memory consumption, it was decided to define a 
basic local heuristic. 
If a vehicle approaches an intersection, a list of candidates is defined including all the nodes connected to the intersection ahead. 
To avoid U turns, the origin node of the road through which the vehicle transits is removed from the list of candidates. If the network does not 
allow left turns, the node located in the left direction is also removed.

The Euclidean distance $d(f,n) = \sqrt{(f_x-n_x)^2+(f_y-n_y)^2}$ between final destination $f$ and every node remaining in the candidate list $n$ is calculated.  
The driver will select the road that leads to the node with the minimum Euclidean distance. If two or more candidate nodes have the same Euclidean distance to 
the final destination, one of them is randomly selected. Depending on the current lane of the vehicle, an attempt to change of lane may be triggered 
(see Section \ref{sec:vehic_mod}).  

It is clear that the heuristic will not provide the shortest route in all the cases. Depending on the topology of the network, the heuristic may not even
lead each vehicle to the selected final destination. However, it provides a valid next road for every case, avoids dead locks, and has a low 
computational cost. Moreover, in our opinion, the heuristic describes well the behaviour of an average driver. Everybody has ended up lost on an unknown 
road when heading towards a defined destination at least once.

\begin{figure}
	\centering
	\includegraphics[scale=0.6]{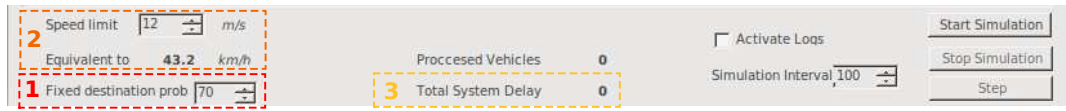}
	\caption[MiniTraSim simulation parameters menu]{Menu with simulation parameters in MiniTraSim.}
	\label{mySim_footer}
\end{figure}

By default, 70\% of the vehicles inserted into the network follow the fixed destination behaviour and 30\% follow random walks. However, this ratio can be modified
in the interface before a simulation is started by a change in the parameter marked with number (1) in Figure \ref{mySim_footer}.

\section{Vehicles Model}
\label{sec:vehic_mod}

\begin{figure}
	\centering
	\includegraphics[scale=1.2]{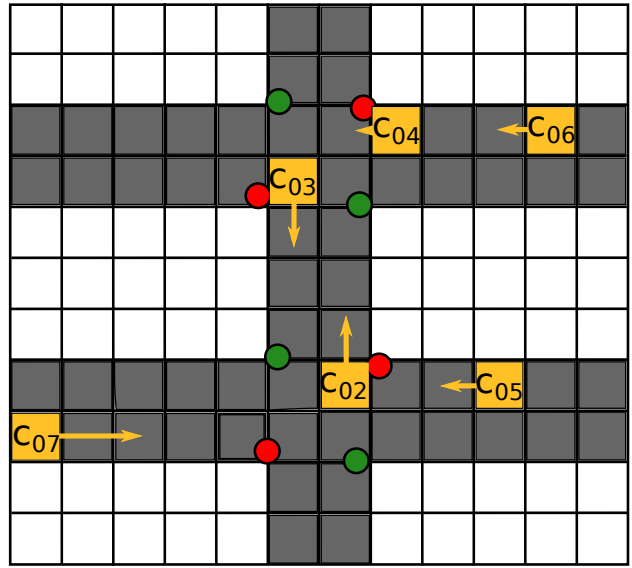}
	\caption[MiniTraSim layer for vehicles]{Representation of simulation step including information of MiniTraSim layer for vehicles.}
	\label{sim_internalRep_cars}
\end{figure}

Vehicles are represented as integer numbers greater than 1 in the network matrix. The number in the matrix corresponds to an unique identifier.
A list with additional information for all active vehicles is used to quickly access and update the network matrix. For each vehicle, the list stores
its identifier, its position, its direction, its current speed, its current road, its current lane, a counter of the amount of time it has been stopped and its 
final destination in case the vehicle is following the fixed destination behaviour. Figure \ref{sim_internalRep_cars} presents an example of the network 
matrix representing a small scenario with some of the additional information stored for vehicles.  

Vehicles follow a cellular automata model similar to the one proposed by Nagel and Schreckenberg in \cite{nagel1992cellular}. For an arbitrary configuration of
the network matrix and the corresponding active vehicle list, one update of the system consists of the following seven consecutive steps: 
\begin{enumerate}
  \item \textbf{Acceleration}: if the velocity $v$ of a vehicle is lower than $v_{max}$ and if the distance to the next vehicle and the 
  distance to the next traffic signal are larger than $v+1$, then the speed is increased by one: $v \leftarrow v + 1$.
  \item \textbf{Break}: if a vehicle at site $i$ sees the next vehicle, or a traffic signal indicating stop, at site $i+j$ (with $j \leq v$), 
  it reduces its speed to $j-1$: $v \leftarrow j-1$.
  \item \textbf{Change of lane}: if the vehicle requires a change from its current lane $l_{now}$ to a different lane $l_{new}$ of the same 
  road and the destination lane is empty in the previous $v$ positions and the next $v+1$ positions, the vehicle changes of lane: 
  $l_{now} \leftarrow l_{new}$.
  \item \textbf{Emergency break}: if a vehicle at site $i$ sees a vehicle changing to its lane at site $i+k$ (with $k \leq v/2$), it does an 
  emergency break procedure and tries a full stop: $v \leftarrow 0$.
  \item \textbf{Change of road}: if the vehicle reaches an intersection and needs to change of road, it updates its direction $d$ following 
  the direction of the new road $A_d$: $d \leftarrow A_d$. 
  \item \textbf{Randomization}: with probability $p$, the velocity of each vehicle (if greater than zero) is decreased by one: $v \leftarrow v-1$. 
  \item \textbf{Car motion}: each vehicle is advanced $v$ positions in its current direction $d$ and current lane $l_{now}$.   
\end{enumerate}

All the roads in the network share the same speed limit $v_{max}$ of 4, equivalent to 12 $m/s$ and 43.2 $km/h$. However, this value can be modified
in the interface by a change in the parameter marked with number (2) in Figure \ref{mySim_footer}. 
A value of $0.1$ was selected for $p$ in the randomization rule in order to reduce the number of phantom jams \cite{maerivoet2005cellular} 
introduced by the cellular automaton. Besides the red light, a vehicle will also consider a green arrow to the left as a stop if it is not required 
to do a left turn or it is required to do a left turn but is not located in the appropriate lane of the road to do it. 

In roads with multiple lanes, vehicles required to turn left or turn right (see section \ref{sec:routing}) try to move to the appropriate lane to 
do it as soon the turn decision is made. The appropriate lane for a left turn is the lane further to the left of the road. The appropriate lane for 
a right turn is the lane further to the right of the road. If the traffic conditions do not allow the vehicle to change to the appropriate lane 
before arriving to the intersection, the vehicle turn attempt is considered a failure and the next road is recalculated.

When a vehicle is inserted into the network (see Section \ref{sec:entry_prob}), an object with the following initial values is appended to the active 
vehicle list. The vehicle identifier is defined based on a global vehicle counter of the simulator. The current position of the vehicle is set to its 
entry node and the selected lane. Its direction is defined based on the road connecting the entry node to the next intersection. The initial speed 
will always be defined as $v_{max}$ because we assume the vehicle is arriving from an unsaturated side road. In case the vehicle follows the fixed 
destination behaviour, an exit node strictly different from its entry node, is randomly selected as final destination from a list with all the exit 
nodes of the network. The stopped time counter is initialized with 0 seconds. 

The main differences between our approach and \cite{nagel1992cellular} are the addition of rules for lane change, road change and emergency break, 
as well as the inclusion of considerations for the traffic signals, the lane and the direction of the vehicles. These modifications were required 
to be able to work in a bi-dimensional environment with traffic signals and multiple lanes. It is clear that the changes increase the computational 
cost of the cellular automaton. However, the cost is still lower than the cost of more realistic models. Basic empirical evidence of this is provided 
in Chapter \ref{sec:singleIntersec}.

\section{Performance Measures}
\label{sec:performMea}

Different traffic units have been used in the literature to measure traffic. Some examples are average travel speed, level of service, average 
intersection delay, average queue length, travel time and total system delay \cite{rao2012measuring}. For MiniTraSim and the majority of the 
experiments presented in Chapter \ref{chap:synthScenarios} and Chapter \ref{chap:sumo}, we selected the total system delay as performance measure.

Total system delay is defined as the sum of the stop time of all vehicles in the system for a defined duration. 
Total system delay displays the effect of congestion in terms of the amount of lost travel time. Because it is a continuous direct value, it
allows transportation professionals to estimate how the modifications to a transportation system affect a particular corridor or the entire system 
\cite{rao2012measuring}. The running value of the total system delay can be consulted in the interface in the label marked with number (3) in 
Figure \ref{mySim_footer}. The final value is also stored in MiniTraSim log.

Even though average speed is more popular in the literature, we selected total delay as the measure of performance because a specific property 
of the vehicular model. The randomization rule of the cellular automaton allows the program to capture natural speed fluctuations due to human behaviour 
or changing external conditions \cite{maerivoet2005cellular}. However, it introduces additional noise to the average speed. Even though the rule 
also affects the total delay (when the rule causes a full stop), its effect on this measure is less than the effect on the average speed. 

\section{Evaluation Order}
\label{sec:simOrder}

Several layers and different update procedures have been discussed in this appendix. Figure \ref{internalRepresentation} displays a 
schematic representation of these layers. At every time step, MiniTraSim follows a specific order of execution for the updates to allow the 
correct operation of the simulator:

\begin{figure}
	\centering
	\includegraphics[scale=0.8]{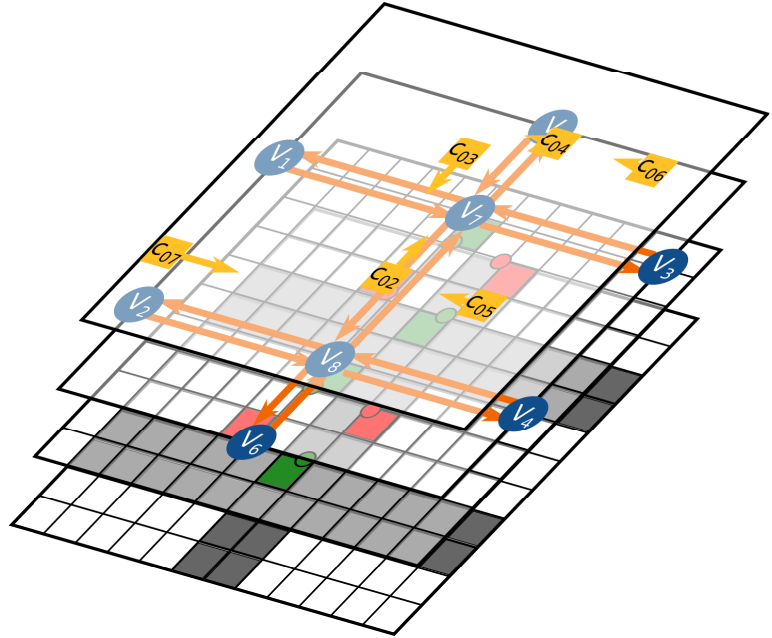}
	\caption[Layers of MiniTraSim]{Representation of the multiple layers in MiniTraSim simulation step.}
	\label{internalRepresentation}
\end{figure}

\begin{enumerate}
 \item \textbf{Traffic Signals}: All the traffic signals in the system are updated following the procedure described in Section \ref{sec:signal_control}, 
 including the required updates to the signal matrix.
 \item \textbf{Vehicle Arrivals}: The procedure described in Section \ref{sec:entry_prob} is executed to validate new entries to the network. 
 Any new vehicle is added to the list of active vehicles.
 \item \textbf{Vehicle Updates}: All the active vehicles are updated with the procedure described in Section \ref{sec:vehic_mod}.
 \item \textbf{Data collection}: When a vehicle reaches an exit node, it is removed from the active vehicle list and the value of its stopped time counter is 
 added to the value of the total system delay counter (see Section \ref{sec:performMea}).
\end{enumerate}
	%CBuilding a Traffic Simulator 

\chapter{SUMO Modifications for Adaptable Signal Control}
\label{apdx:sumo_diff}

  \graphicspath{{figures/PNG/}{figures/PDF/}{figures/EPS/}{figures/}}
  \lhead{\emph{Appx B: SUMO Modifications for Adaptable Signal Control}}  % Set the left side page header to "Building a Traffic Simulator"

A framework to evaluate traffic controllers using SUMO was designed and developed for the experiments presented in Chapter \ref{chap:sumo}.
The framework requires the controllers to be defined as a set of Python classes. Each class should contain a method called {\tt trafficRule}
with two integer arguments: {\tt vQueue} and {\tt hQueue}, and return an integer number. The returned value is used by the framework to
perform online modifications to the time schedule of the traffic signal associated with each controller.

Two main modifications were required to implement this framework:
additional configuration files are required for an existing SUMO traffic scenario and additional layers are required for the interaction 
between Genetic Programming and SUMO. This appendix contains a detailed description, with configuration examples, of all the 
modifications implemented. 

All the code presented in this appendix is open-source and is available in a sourceforge repository \cite{ricalde2017sorceforge}.
It is important to emphasize that the proposed framework is modular and minor modifications are required in order to test different
machine learning methods in solution to traffic scenarios. However, the exhaustive computational time required by traffic simulation 
should be considered in the design of any project related to traffic signal optimization.

\section{Detectors}
\label{sec:sumo_detect}

To operate in the same way as MiniTraSim (see Appendix \ref{sec:signal_control}), the controllers require access to the number of vehicles 
in full stop on the inbound roads of the controlled intersection. Fortunately, SUMO entry-exit detectors allow the identification of individual 
vehicles in full stop condition for a given section of road. A entry-exit detector is defined by the entry position and exit position of a 
specific lane from a specific road.

In SUMO, entry-exit detectors are defined within an additional configuration file which has to be loaded by the simulator. The configuration file 
is defined using the following format:

{\footnotesize
\singlespacing
\begin{verbatim}
<additional>
   <entryExitDetector id="<ID>" freq="<AGGREGATION_TIME>" file="<OUTPUT_XMLFILE>" 
   timeThreshold="<FLOAT>" speedThreshold="<FLOAT>">
      <detEntry lane="<LANE_ID1>" pos="<POSITION_ON_LANE>"/>
      <detExit lane="<LANE_ID1>" pos="<POSITION_ON_LANE>"/>
   </entryExitDetector>
   ... further detectors ...   
</additional>
\end{verbatim}
}

\begin{figure}
\centering
\includegraphics[scale=0.7]{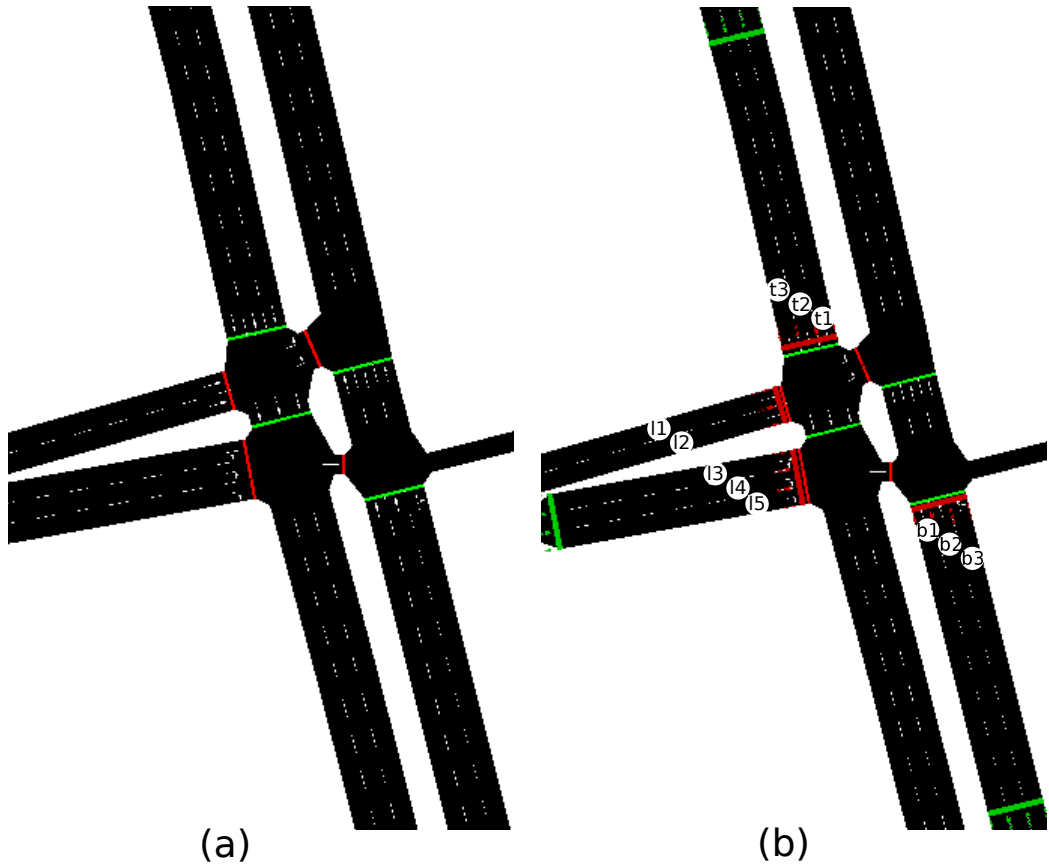}
\caption[Example of an intersection in SUMO]{Example of an intersection in SUMO. (a) does not include detectors. (b) includes entry-exit 
detectors with unique labels. The entry points of the detectors are marked as wide green lines. The exit points of the detectors are marked as wide red lines.}
\label{sumo_intersection}
\end{figure}

An independent entry-exit-detector should be defined for each lane of each inbound road of each intersection affected by a controller. 
For example, Figure \ref{sumo_intersection}(a) presents an intersection without detectors. Figure \ref{sumo_intersection}(b) presents 
the same intersection with all the entry-exit detectors required to operate with an actuated controller. 

The detectors provide
the inputs required by the {\tt trafficRule}. An entry-exit detector allows the collection of different data from the vehicles located inside the road 
section assigned to it. The framework only uses a single value from the entry-exit detectors: last step halting number. Last step halting number is 
defined as the number of stopped vehicles during the last time step in the zone covered by the detector.

\section{Intersection Configuration}
\label{sec:sumo_ligths}

The framework requires to configure three different components associated with each intersection: (1) entry-exit detectors should be 
associated with one of the four directions considered in the controller logic: top, bottom, left or right; (2) the intersection should 
be associated with an initial time schedule; and (3) specific phases in the time schedule should be associated with the controller events: 
vertical effect, horizontal effect and decision step.

The first step is to associate the detectors with the directions used in the controller logic. An additional configuration file is required to 
define these associations. The configuration file is defined using the following format: 

{\footnotesize
\singlespacing
\begin{verbatim}
<configuration>
   <intersection id="<INTERSECTION_ID>">
      <option dir="bottom">
         <detector id="<DETECTOR_ID>" multiplier="<PRIORITY>"/>
         ... further detectors ...
      </option>	
      <option dir="top">
         <detector id="<DETECTOR_ID>" multiplier="<PRIORITY>"/>
         ... further detectors ...
      </option>	
      <option dir="left">
         <detector id="<DETECTOR_ID>" multiplier="<PRIORITY>"/>
         ... further detectors ...
      </option>	
      <option dir="right">
         <detector id="<DETECTOR_ID>" multiplier="<PRIORITY>"/>
         ... further detectors ...
      </option>	
   </intersection>
   ... further intersections ...
</configuration>
\end{verbatim} 
} 

The parameter identified as \texttt{multiplier} is an integer number and can be used to increase the priority of a specific lane. This priority can be used for 
lanes dedicated for special vehicles. For example, bus-exclusive lanes or lanes for emergency vehicles. For standard lanes, it is 
recommended to use the default value of 1. 

Once the detectors are associated
with a direction, the value of {\tt vQueue} will be equal to the sum of all the top and bottom detectors 
(modified by the corresponding multipliers), and the value of {\tt hQueue} will be equal to the sum of all 
the left and right detectors (modified by the corresponding multipliers).

For example, for the intersection presented in Figure \ref{sumo_intersection}(b) the association with entry-exit detectors is simple: 
Detectors t1, t2 and t3 are associated with \textbf{top} direction; b1, b2 and b3 are associated with \textbf{bottom} 
direction; and l1, l2, l3, l4 and l5 are associated with \textbf{left} direction. The configuration for this intersection is as follows:

{\footnotesize
\singlespacing
\begin{verbatim}   
<intersection id="221">
   <option dir="bottom">
      <detector id="221_b1" multiplier="1"/>
      <detector id="221_b2" multiplier="1"/>
      <detector id="221_b3" multiplier="1"/>
   </option>		
   <option dir="top">
      <detector id="221_t1" multiplier="1"/>
      <detector id="221_t2" multiplier="1"/>
      <detector id="221_t3" multiplier="1"/>
   </option>
   <option dir="left">
      <detector id="221_l1" multiplier="1"/>
      <detector id="221_l2" multiplier="1"/>
      <detector id="221_l3" multiplier="1"/>
      <detector id="221_l4" multiplier="1"/>
      <detector id="221_l5" multiplier="1"/>
   </option>
</intersection>	 
\end{verbatim}
} 

In SUMO, the traffic signals are usually configured with the definition of traffic signal schedules. Signal schedules are defined within an 
additional configuration file which has to be loaded by the simulator. The configuration file is defined using the following format:

{\footnotesize
\singlespacing
\begin{verbatim}
<additional>
   <tlLogic id="<INTERSECTION_ID>" type="<SIGNAL_TYPE>" 
    programID="<SIGNAL_PLAN_NAME>">   
        <phase duration="<DURATION_0>" state="<STATUS_OF_SIGNALS>" 
         minDur="<MINIMUM_DURATION_0>" maxDur="<MAXIMUM_DURATION_0>"/>
        <phase duration="<DURATION_1>" state="<STATUS_OF_SIGNALS>" 
         minDur="<MINIMUM_DURATION_1>" maxDur="<MAXIMUM_DURATION_1>"/>
        ... further phases ...   
   </tlLogic>   
   ... further intersections ...   
</additional>
\end{verbatim}
} 

\begin{figure}
\centering
\includegraphics[scale=0.4]{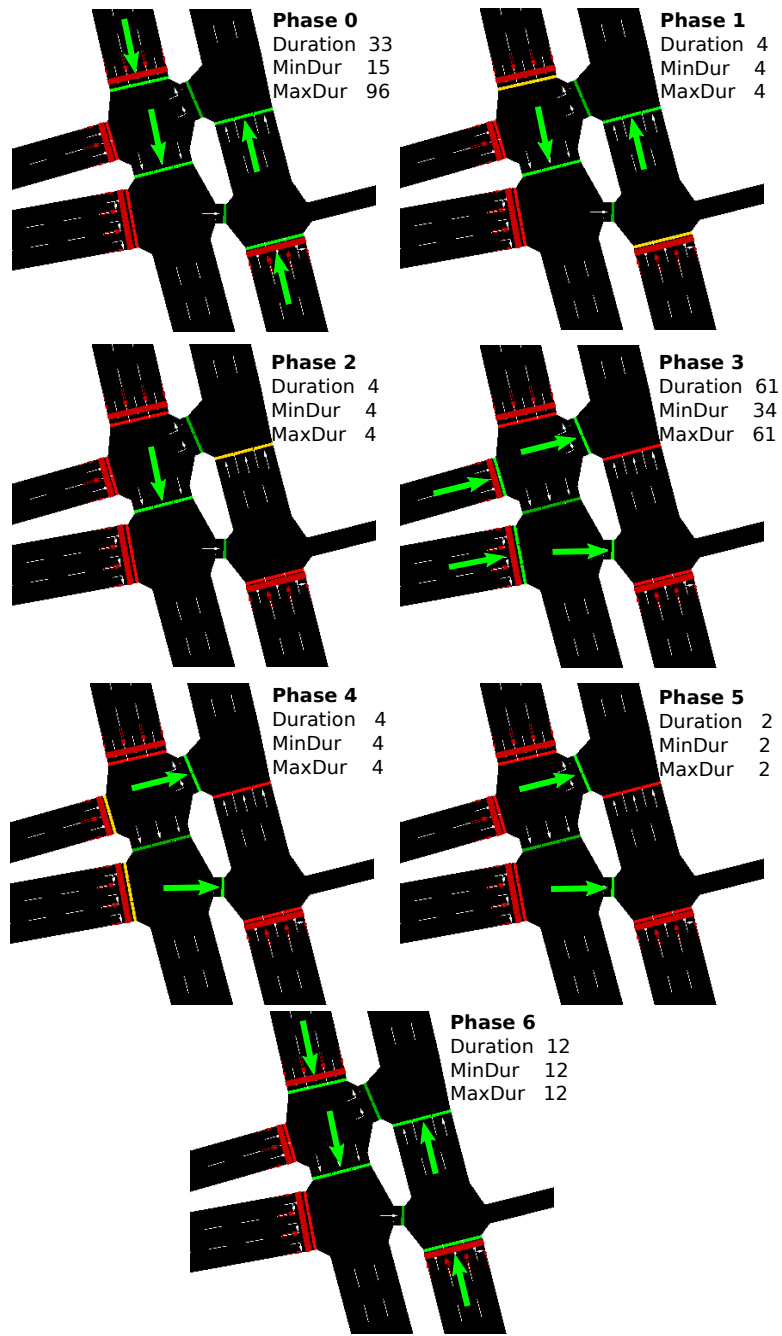}
\caption[Graphic representation of time schedule]{Graphic representation of time schedule.}
\label{bologna_timeSchedule}
\end{figure}

The state of each phase is a string of characters containing the light status for each lane of each road in the intersection. 
For example, a valid traffic signal schedule for the intersection of Figure \ref{sumo_intersection}(b) is presented in Figure \ref{bologna_timeSchedule} 
and defined with the following configuration:

{\footnotesize
\singlespacing
\begin{verbatim}
<tlLogic id="221" type="static" programID="utopia" offset="0">   
    <phase duration="33" state="GGGggrrGGGGGGGgrrrrGGG" minDur="15" maxDur="96"/>     
    <phase duration="4" state="GGGggrryyyyyyygrrrrGGG" minDur="4" maxDur="4"/>        
    <phase duration="4" state="yyyggrrrrrrrrrGrrrrGGG" minDur="4" maxDur="4"/>        
    <phase duration="61" state="rrrGGGGrrrrrrrGGGGGggg" minDur="34" maxDur="61"/>     
    <phase duration="4" state="rrrGGyyrrrrrrrGyyyyggg" minDur="4" maxDur="4"/>        
    <phase duration="2" state="rrrGGrrrrrrrrrGrrrrGGG" minDur="2" maxDur="2"/>        
    <phase duration="12" state="GGGggrrGGGGGGGgrrrrGGG" minDur="12" maxDur="12"/>     
</tlLogic> 
\end{verbatim}
} 

A configuration file with the time schedule of the traffic signals is usually provided for every traffic scenario in SUMO. 
However, the parameters {\tt minDur} and {\tt maxDur} are optional. These parameters are required by our framework to define the allowed range
of durations for each phase in the controllers. In case a traffic scenario already existing in SUMO is being configured to be used with the framework, 
the user will need to provide {\tt minDur} and {\tt maxDur} values for all the phases in the previously defined time schedule configuration file. 

The last configuration file required for the intersection is used to define the association of specific phases with controller events. The controller has three
different events: vertical influence, horizontal influence and decision step. When a phase is associated with vertical influence events or horizontal influence events, 
the duration of the phase will be modified by the value returned by the controller. Meanwhile, the controller will be executed at 
the beginning of a phase marked with the decision step event.   

Zero, one or more phases can be associated with each event. Event associations are defined within an additional configuration file
which has to be loaded by the experiment runner. The configuration file is defined using the following format:

\pagebreak
{\footnotesize
\singlespacing
\begin{verbatim}
<phases>
   <intersection id="<INTERSECTION_ID>">
      <horizontal phase="<PHASE_INDEX>"/>
      <vertical phase="<PHASE_INDEX>"/>
      <decision phase="<PHASE_INDEX>"/>
      ... further phase events ... 
   </intersection>
   ... further intersections ...      
</phases>
\end{verbatim}
} 

Because the phases do not have an unique identifier in the time schedule configuration, their positional index is used for the event configuration file. For example, 
the intersection presented in Figure \ref{sumo_intersection}(b) with the time schedule presented in Figure \ref{bologna_timeSchedule} 
has the following event configuration:

{\footnotesize
\singlespacing
\begin{verbatim}
<intersection id="221">
   <vertical phase="0"/>
   <decision phase="2"/>
   <horizontal phase="3"/>
   <decision phase="5"/>		
</intersection>
\end{verbatim}
} 

As part of the proposed framework, a Python program with the name {\tt light.py} was written to manage all configurations related to the intersections.
The program loads three configuration files for intersections: association of detectors and directions, initial time schedules, and phases 
associated with controller events. The program also maintains a reference to the controller associated with each intersection and contains a method to
calculate the normalized adaptive factor. The code is available in the sourceforge repository of the framework \cite{ricalde2017sorceforge}.
   
\section{Experiment Runner}
\label{sec:sumo_runner}

The main component of the framework is called \textit{Experiment runner}. The experiment runner layer is contained in a single Python program with the name 
{\tt runnerExternal.py}. The program is a modification to a component included in the SUMO release. The original component was a 
tutorial for traffic light control via the TraCI interface. The layer performs the following tasks:
\begin{enumerate}
  \item Manages the communication with SUMO through TraCI, 
  \item Controls the simulation time step,
  \item Loads information of the intersections to be optimize consuming the classes contained in {\tt light.py}, 
  \item Access the intersection traffic queues of the detectors,
  \item Executes the traffic controller associated with each actuated intersection,
  \item Updates the time schedule of the intersections based on the controller decision, 
  \item Modifies the controllers when an epigenetic mutation is triggered.
\end{enumerate}

It is important to emphasize that the proposed framework is modular and minor modifications are required in order to test other
machine learning methods in solution to traffic scenarios. However, the exhaustive computational time required by traffic simulation should be
considered in the design of any project related to traffic signal optimization.
The code for the framework is available in a sourceforge repository \cite{ricalde2017sorceforge}.
	%Evolving SUMO Signal Controllers 

\addtocontents{toc}{\vspace{2em}}  % Add a gap in the Contents, for aesthetics
\backmatter

%% ----------------------------------------------------------------
\label{Bibliography}
\lhead{\emph{Bibliography}}  % Change the left side page header to "Bibliography"
\bibliographystyle{ieeetr}

\bibliography{ref}  % The references (bibliography) information are stored in the file named "ref.bib"

\addtocontents{toc}{\vspace{2em}} % Add a gap in the Contents, for aesthetics

\setstretch{1}  % It is better to have smaller font and larger line spacing than the other way round

\chapter{About the Author}
\label{apdx:about}
  \graphicspath{{figures/PNG/}{figures/PDF/}{figures/EPS/}{figures/}}
  \lhead{\emph{Appx C: About the Author}}  % Set the left side page header to "Building a Traffic Simulator"

%--------------------TITLE-------------
\par{\centering
		{\Large Esteban Ricalde Gonz\'alez
	}\bigskip\par}

%--------------------SECTIONS-----------------------------------
%Section: Education
\section*{Education}
\begin{tabular}{rp{11cm}}	
 {\small \emph{Current}} & \textbf{PhD} in \textsc{Computer Science}, \textbf{MUN}, Canada\\
& \textsc{Thesis}: \textit{A Genetic Programming System with an Epigenetic Mechanism for Traffic Signal Control} \\
& \textsc{Supervisor}: Prof. Wolfgang Banzhaf \\
& \textsc{Gpa}: 90/100\\&\\&\\
{\small \textsc{Sept} 2008}& \textbf{MSc} in \textsc{Computer Science}, \textbf{UNAM}, Mexico\\
& \textsc{Thesis}: \textit{Analysis of neutrality on GP} \\
& \textsc{Supervisor}: Prof. Katya Rodr\'iguez-V\'azquez \\
& \textsc{Gpa}: 90/100\\&\\&\\
{\small \textsc{July} 2005} & \textbf{BEng} in \textsc{Computer Systems}, \textbf{ITSX}, Mexico\\
& \textsc{Gpa}: 92/100
\end{tabular}

\section*{Courses and Certifications}
\begin{tabular}{rp{13cm}}
 {\small 2016} & \textbf{Introduction to Advanced Computing}\\
 & ACENET, ComputeCanada regional partner, Memorial University of Newfoundland, St. John's, CA\\&\\
 {\small 2013}& \textbf{Teaching Skills Enhancement Program}\\
 & Distance Education, Learning and Teaching Support, Memorial University of Newfoundland, St. John's, CA \\&\\
 {\small 2005} & Course \textbf{Computational Models Inspired in Biology}\\
 & IIMAS, UNAM, Mexico City, MEX. \\&\\
 {\small 2004} & \textbf{CCNA Cisco Certification} (3.1 version)\\
 & School of Mechanical and Electrical Engineering, UV, Xalapa, MEX\\&\\
 {\small 2004} & Course \textbf{Teaching Basic Skills}\\
 & Educators Association of Veracruz State, Xalapa, MEX
\end{tabular}

\section*{Academic Experience}
\subsection*{Teaching}
\begin{tabular}{rp{11cm}}
 {\small \textsc{Fall} 2016} & \textbf{COMP-1510}, Introduction to Programming for Scientific Computing\\
 & Computer Science Department, Memorial University of Newfoundland, St. John's, NL, CA\\&\\
 {\small \textsc{Fall} 2015} & \textbf{COMP-1510}, Introduction to Programming for Scientific Computing\\
 & Computer Science Department, Memorial University of Newfoundland, St. John's, NL, CA\\&\\
 {\small 2004-2005} & \textbf{Computing} Teacher\\
 & Xalitic Junior High School, Xalapa, MEX
\end{tabular}

\subsection*{Teaching Assistantships}
\begin{tabular}{rp{11.5cm}}
 {\small \textsc{Winter} 2016} & \textbf{COMP-6925}, Advanced Operating Systems. \textit{Marking}\\
 {\small \textsc{Winter} 2015} & \textbf{COMP-2000}, Collaborative and Emergent Behaviour. \textit{Lab}\\
 {\small \textsc{Fall} 2014} & \textbf{COMP-3716}, Software Methodology. \textit{Marking}\\
 {\small \textsc{Winter} 2014} & \textbf{COMP-1510}, Introduction to Programming for Scientific Computing. \textit{Marking {\&} Lab}\\
 {\small \textsc{Winter} 2013} & \textbf{COMP-3754}, Introduction to Information and Intelligent Systems. \textit{Marking}\\
 {\small \textsc{Fall} 2013} & \textbf{COMP-1700}, Introduction to Computer Science. \textit{Lab}\\
 {\small \textsc{Fall} 2012} & \textbf{COMP-3724}, Computer Organization. \textit{Marking}\\
 {\small \textsc{Fall} 2012} & \textbf{COMP-4721}, Operative Systems. \textit{Marking}\\
 & Computer Science Department, Memorial University of Newfoundland, St. John's, NL, CA\\&\\
 {\small \textsc{Fall} 2012} & Spanish Conversation Class. \textit{Monitor}\\
 {\small \textsc{Winter} 2012} & Spanish Conversation Class. \textit{Monitor}\\
& Digital Learning Centre, Memorial University of Newfoundland, St. John's, NL, CA\\&\\
 {\small 2003-2005} & Network Administrator and Technical Support. \textit{Lab}\\
& CCNA Cisco Laboratory, School of Mechanical and Electrical Engineering, UV, Xalapa, MEX
\end{tabular}

\section*{Scholarships}
\begin{tabular}{rl}
 {\small 2012-2016} & \textbf{MUN} scholarship for PhD student \\
 {\small \textsc{Summer} 2007} & \textbf{PAPIIT Scholarship} awarded by DGAPA, UNAM\\
 {\small 2005-2007} & \textbf{CONACYT} scholarship for graduated studies number 222262
\end{tabular}

\section*{Languages}
\begin{tabular}{rl}
\textsc{Spanish:}&Native language\\
\textsc{English:}& 597 points TOEFL PBT on August 2011
\end{tabular}

\section*{Publications}
\begin{tabular}{rp{12cm}}
 {\small \textsc{Dec} 2017} &  \textbf{Esteban Ricalde} and Wolfgang Banzhaf. A Genetic Programming Approach for the Traffic Signal Control Problem with Epigenetic Modifications. 
 \textit{European Conference on Genetic Programming}. Springer International Publishing, 2016, pp. 133-148. \\&\\
 {\small \textsc{Mar} 2016} &  \textbf{Esteban Ricalde} and Wolfgang Banzhaf. Evolving Adaptive Traffic Signal Controllers for a Real Scenario Using Genetic Programming with an Epigenetic Mechanism. 
 \textit{16th IEEE International Conference on Machine Learning and Applications (ICMLA)}. IEEE, 2017, pp. 897-902.\\&\\
 {\small \textsc{July} 2007} &  \textbf{Esteban Ricalde} and Katya Rodr{\'i}guez-V{\'a}zquez. A GP neutral function for the Artificial Ant Problem. 
 \textit{GECCO 2007 Proceedings and Companion Material CD-ROM}. ACM, 2017, pp. 2565. (Late Breaking Paper)\\&\\
 {\small \textsc{Nov} 2006} & Christopher Stephens, Edgar Arenas, Jorge Cervantes, Beatriz Peralta, \textbf{Esteban Ricalde} and Cristian Segura. When are Building Blocks Useful?. 
 \textit{Fifth International Conference on Artificial Intelligence 2007}. IEEE, Mexico, 2006, pp. 217-226.
\end{tabular}

\section*{Conference Attendance}

\begin{tabular}{rp{12cm}}
 {\small \textsc{May} 2015} &  Atlantic Provinces Transportation Forum 2015 \\
 & Talk \textbf{Intelligent Solutions to the Traffic Signal Control Problem} \\
 & Harris Centre, Memorial University of Newfoundland, St. John's, NL, CA\\&\\
 {\small \textsc{Oct} 2010} &  XVII National Week of Science and Technology \\
 & Talk \textbf{Artificial Intelligence: How far have we got?} \\
 & Instituto Tecnol\'ogico de Estudios Superiores de Apan, Apan, MEX\\&\\
 {\small \textsc{Jan} 2007} &  Foundations of Genetic Algorithms IX. \textit{Volunteer} \\
 & Nuclear Sciences Institute, UNAM, Mexico City, MEX\\&\\
 {\small \textsc{Oct} 2006} &  Visual Computing: Fundamentals and Applications \\
 & Poster \textbf{Ant Colony Simulation with Augmented Reality Visualization} \\
 & IIMAS, UNAM, Mexico City, MEX
\end{tabular}

\begin{tabular}{rp{12cm}}
 {\small \textsc{Oct} 2006} &  XXXIX National Congress of the Mexican Mathematical Society  \\
 & Talk \textbf{Ant Colony Simulation: Multi-agent System Control Heuristic (part 1)} \\
 & Villahermosa, MEX\\&\\
 {\small \textsc{Apr} 2006} &  SF-Conacyt 2006 Workshop: Approximate Dynamic Programming (ADP). \textit{Attendant} \\
 & Cocoyoc, MEX\\&\\
 {\small \textsc{May} 2005} & Talk \textbf{Genetic Algorithms and Evolutionary Computation} \\
 & Veracruz, MEX\\&\\
 {\small \textsc{Nov} 2004} & Free Software GNU / Linux International Conference. \textit{Attendant} \\
 & Universidad de Xalapa, Xalapa, MEX\\&\\ 
 {\small \textsc{Nov} 2002} &  III Informatics and Computer Systems Congress  \\
 & Talk \textbf{Children Psychomotor Troubleshooting Multimedia System} \\
 & Instituto Tecnol\'ogico Superior de Misantla, Misantla, MEX\\&\\
 \textsc{Nov} 2001 & VIII Information Technology Symposium. \textit{Attendant} \\
 & Universidad Veracruzana, Xalapa, MEX
\end{tabular}

\section*{Awards and Honours}
\begin{tabular}{rp{11.5cm}}
 {\small \textsc{Mar} 2015} & \textbf{Fellow of the School of Graduate Studies}\\
 & SGS, Memorial University of Newfoundland, St. John's, NL, CA\\&\\
 {\small 2013-2015} & Computer Science Graduate Society\\ 
 & \textbf{Founder member and president}\\
 & Memorial University of Newfoundland, St. John's, NL, CA\\&\\
 {\small \textsc{Mar} 2004} & Computer Contest (state phase) for ninth grade. Cultural Technologic and Academic Evaluation, SEST Meeting\\
 & \textbf{Jury member}, Xalapa, MEX\\&\\
 {\small \textsc{June} 2002} & ITSX First Programming Contest\\
 & \textbf{Third place}, Xalapa, MEX\\&\\
 {\small \textsc{Winter} 2000} & Instituto Tecnol\'ogico Superior de Xalapa\\
 & \textbf{Highest GPA}, Xalapa, MEX
\end{tabular}

%Section: Work Experience at the top
\section*{Work Experience}
\begin{tabular}{rl}
 {\small 2017-2019} & \textbf{Sofware Developer} \\
 & Radient360 Solutions Inc. \\
 & St John's, Newfoundland, CAN\\&\\
 {\small 2015-2016} & \textbf{Freelance Data Analyst} \\
 & Hey Orca.  \\
 & St John's, Newfoundland, CAN\\&\\
 {\small 2011-2012} & \textbf{Senior Developer} \\
 & RGOV - Microsoft Innovation Center. \\
 & Mexico City, MEX\\&\\
 {\small 2009-2011} & \textbf{Research and Development Engineer} \\
 & SeguriData. \\
 & Mexico City, MEX\\&\\
 {\small 2008-2009} & \textbf{Support Engineer} \\
 & Citibank. \\
 & Mexico City, MEX\\&\\
 {\small 2007-2008} & \textbf{C Programmer} \\
 & Bufete Consultor Mexicano. \\
 & Mexico City, MEX\\&\\
 {\small 2002-2003} & \textbf{Web Developer} \\
 & Biological Research Institute, University of Veracruz. \\
 & Xalapa, MEX
\end{tabular}

\section*{Computer Skills}
\begin{tabular}{rp{9cm}}
Operating System:& Linux (RedHat, Mandrake, Debian, Fedora, Ubuntu, NLD, Suse), Unix (Irix, Solaris, OSS), Windows (95, 98, XP, Vista, Server, 7, 10)\\
Office Packages:& Microsoft Office, Libre Office, \LaTeX, InkScape, GIMP \\
Data Bases:& SQL Server, Oracle, mySQL, neo4j\\
Modelling Languages:& Matlab, SUMO, R, PRISM \\
Basic Knowledge:& Delphi, LISP, Fortran\\
Intermediate Knowledge:& Java, PHP, Prolog, CSHTML, Python\\
Advanced Knowledge:& C/C++, C{\#}, HTML, JavaScript
\end{tabular}

\section*{Interests and Activities}
\begin{itemize}
\item \textbf{Tangled Threads / Fils Entrem\^el\'es}, Traveling exhibition in Newfoundland 	\& Labrador, Canada. \textit{Illustrator} (2017) 
\item \textbf{Culture-to-Community Educational Outreach Program}. \textit{Cultural presentation volunteer} (Winter 2016) 
\item Aikido. Black belt with 18 years of practice
\item Reading. Fantasy and SciFi books 
\item Drawing and Painting 
\end{itemize}

	%CAME About the author 

\end{document}